\documentclass{article}

\usepackage[preprint]{neurips_2020}

\usepackage[utf8]{inputenc} % allow utf-8 input
\usepackage[T1]{fontenc}    % use 8-bit T1 fonts
\usepackage{hyperref}       % hyperlinks
\usepackage{url}            % simple URL typesetting
\usepackage{booktabs}       % professional-quality tables
\usepackage{amsfonts}       % blackboard math symbols
\usepackage{nicefrac}       % compact symbols for 1/2, etc.
\usepackage{microtype}      % microtypography

\usepackage{algorithm}
\usepackage{algpseudocode}
\usepackage{amsmath}
\usepackage{graphicx}
\usepackage{caption}
\usepackage[font=small]{subcaption}
\usepackage{wrapfig, booktabs}

\usepackage{xcolor}

\DeclareMathOperator*{\argmax}{\arg\!max}

\title{Adversarial defenses via a mixture of generators}

\author{%
  Maciej {\.Z}elaszczyk\\
  Faculty of Mathematics and Information Science\\
  Warsaw University of Technology\\
  00-662, Warsaw \\
  \texttt{m.zelaszczyk@mini.pw.edu.pl} \\
  \And
  Jacek Ma{\'n}dziuk\\
  Faculty of Mathematics and Information Science\\
  Warsaw University of Technology\\
  00-662, Warsaw \\
  \texttt{mandziuk@mini.pw.edu.pl} \\
}

\begin{document}

\maketitle

\begin{abstract}
  In spite of the enormous success of neural networks, adversarial examples remain a relatively weakly understood feature of deep learning systems. There is a considerable effort in both building more powerful adversarial attacks and designing methods to counter the effects of adversarial examples. We propose a method to transform the adversarial input data through a mixture of generators in order to recover the correct class obfuscated by the adversarial attack. A canonical set of images is used to generate adversarial examples through potentially multiple attacks. Such transformed images are processed by a set of generators, which are trained adversarially as a whole to compete in inverting the initial transformations. To our knowledge, this is the first use of a mixture-based adversarially trained system as a defense mechanism. We show that it is possible to train such a system without supervision, simultaneously on multiple adversarial attacks. Our system is able to recover class information for previously-unseen examples with neither attack nor data labels on the MNIST dataset. The results demonstrate that this multi-attack approach is competitive with adversarial defenses tested in single-attack settings.
\end{abstract}

\section{Introduction}
\label{introduction}

The discovery of adversarial examples in the image domain has fueled research to better understand the nature of such examples and the extent to which neural networks can be fooled. Work on new adversarial methods has produced various approaches to the generation of adversarial images. These \emph{adversarial attacks} can be either geared toward misclassification in general or toward inducing the classifier assign the adversarial example to a specific class. Conversely, \emph{adversarial defenses} are approaches designed with the goal of nullifying the effect of adversarial effects on a classifier.

In this work, we aim to show the possibility of defending against adversarial examples by utilizing a set of neural networks competing to reverse the effects of adversarial attacks.

\textbf{Our contributions:}
\begin{itemize}
    \item We show that it is possible to use a GAN-inspired architecture to transform adversarial images to recover correct class labels.
    \item We show that such an architecture can be extended to handle multiple adversarial attacks at once by making the generators compete to recover the original images on which the adversarial examples are based.
    \item We demonstrate that such a system can be trained with no supervision, i.e. with no knowledge of:  1) the type of the adversarial attacks used, 2) the original images, 3) the class labels of the original images.
    \item A system trained in this way is able to recover correct class labels on previously-unseen adversarial examples for a number of adversarial attacks.
    \item The accuracy of a pretrained classifier on the images transformed by the system is meaningfully higher than that of a random classifier. For a number of our setups, including multi-attack ones, it is on par with the results of modern adversarial defenses, which are typically tested in single-attack settings.
    \item For a range of adversarial attacks, the system is able to invert the adversarial attack and produce images close to the clean images that were attacked.
    \item We highlight the fact that the components of such an unsupervised system specialize to some extent, revealing the distinction between attacks which produce images interpretable to humans and attacks which severely distort the visual aspects of the original images.
\end{itemize}

\begin{wrapfigure}{r}{0.5\textwidth}
  \centering
  \includegraphics[width=0.42\textwidth]{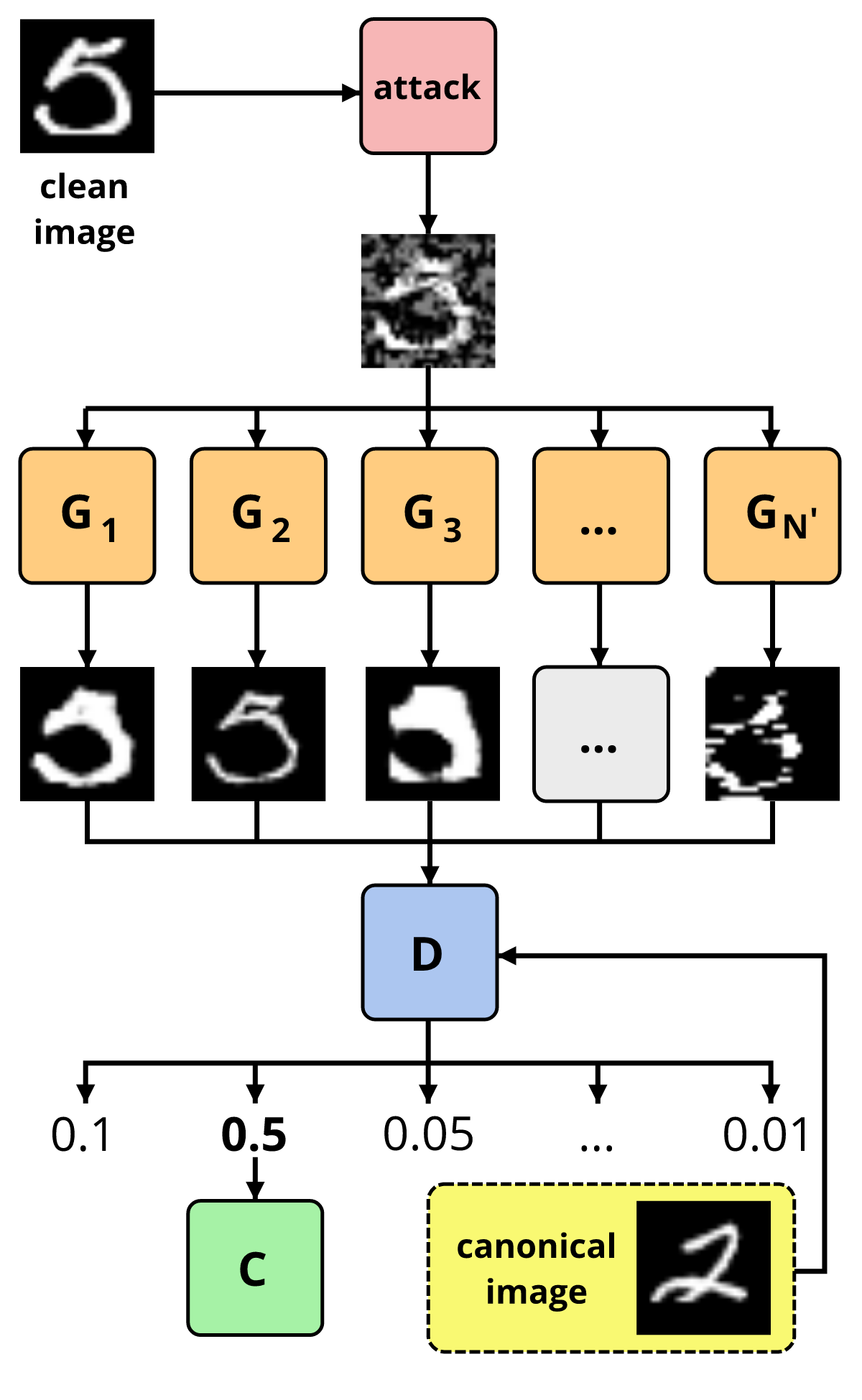}
  \caption{Overview of the system. A clean image is attacked and fed to the generators. The output of the generators is scored by the discriminator and the best-performing generator ($G_{2}$) is updated. The discriminator is trained on the output of all the generators and on canonical images. At test time, the output of the best-performing generator is passed to the pretrained classifier ($C$).}
\end{wrapfigure}

\section{Related work}
\label{section:related-work}

This work is related to a number of lines of research linked to adversarial examples. \cite{Szegedy2014} show that it is possible to construct adversarial images by adding perturbations to original images with the goal of the perturbation causing misclassification. This is done via the L-BFGS method. The resulting changes in the image may be imperceptible to the human observer but they still successfully cause the classifier to assign an incorrect class to the image relative to the unperturbed one. \cite{Goodfellow2015} show that the linear behavior of neural networks in high-dimensional spaces suffices to produce adversarial examples. They introduce the \emph{fast gradient sign method} (FGSM) - a method of obtaining adversarial examples computationally less intensive than \cite{Szegedy2014}. \cite{Ilyas2019} tie the existence of adversarial examples to the presence of statistical, but not human-interpretable, features in the data. They show that such features are pervasive in popular datasets and while predictive of the class, they are not necessarily robust.

The L-BFGS and FGSM methods were followed by an array of diverse adversarial attack methods. \cite{Papernot2016a} present the \emph{Jacobian-based  Saliency  MapAttack}  (JSMA). \cite{Madry2018} suggest that \emph{projected gradient descent} (PGD) can be used to generate adversarial examples and that PGD is an universal adversarial method. \cite{Moosavi-Dezfooli2016} propose the DeepFool method to obtain adversarial examples and compare the robustness of classifiers to such examples. \cite{Carlini2017} show that \emph{defensive distillation} (one of the adversarial defense methods) is still susceptible to adversarial attacks. \cite{Brendel2018} introduce \emph{Boundary Attack} - a decision-based adversarial attack gradually decreasing the size of the perturbation while ensuring that it, in fact, stays adversarial.

While initial research on adversarial examples focused on intentionally perturbing the whole image, further works have shown that attacks may be aimed at smaller areas - \cite{Su2019} show that one-pixel attacks can produce successful adversarial examples. Additionally, it has been shown that the adversarial attack do not necessarily need to be tied to a specific image. It is possible to generate general patches which can then be superimposed on original images to produce adversarial examples as demonstrated by \cite{Brown2017}. These patches do not need to be produced in a strictly digital way - it is possible to physically print them and attach to other images and still get adversarial examples.

The emergence of adversarial examples has led to research on the possibility to defend from the adversarial attacks, a process known as an \emph{adversarial defense}. \cite{Papernot2016b} introduce the \emph{distillation defense} as a mechanism to counter adversarial attacks. \cite{Guo2018} focus on image transformations as adversarial defenses. In particular, they show that \emph{total variance minimization} and \emph{image quilting} can be effective adversarial defenses. \cite{Buckman2018} present the possibility of using image discretization as an adversarial defense. They show that a specific form of discretization, \emph{thermometer encoding} increases the robustness of neural networks to adversarial attacks. \cite{Mustafa2019} propose to disentangle the intermediate learned features and show that this increases the robustness of the classifiers without deteriorating the performance on unperturbed images. \cite{Pang2019} posit that the robustness of neural networks to adversarial attacks can be improved by ensuring more diversity in ensemble models.

\emph{Generative adversarial networks} (GANs) \cite{Goodfellow2014} are a network architecture directly stemming from the existence of adversarial examples. They are linked to our work not only via adversarial examples but also by the fact that we use a GAN-inspired architecture to handle image transformations. A more direct inspiration comes from the idea that it may be possible to train ensembles of conditional generators to identify and reverse transformations in the image domain - a concept proposed by \cite{Parascandolo2018}.

\section{Recovering relevant features through a set of generators}
\label{section:mixture}

We approach the problem of finding adversarial defenses at the data preprocessing step. We aim to link ensemble methods with GAN-based training to invert the adversarial transformations.

\subsection{Adversarial attacks as mechanisms}

Following \cite{Parascandolo2018}, let us consider a canonical distribution $P$ on $\mathbb{R}^d$ and $N$ \emph{mechanisms} $A_1, \dots, A_N$, which are functions. These mechanisms give rise to $N$ distributions $Q_1, \dots, Q_N$, where $Q_j = A_j(P)$. We specifically consider a given mechanism $A_j$ to be an adversarial attack built with respect to a classifier $C$ on a given dataset. An important assumption is that at training time we receive two datasets: (1) a canonical, unperturbed dataset $\mathcal{D}_{P}$ drawn from the canonical distribution $P$ and (2) a transformed dataset $\mathcal{D}_Q = (x_i)_{i = 1}^{n}$ sampled independently from a mixture of $Q_1, \dots, Q_N$.

The aim is to learn, if possible, approximate inverse mappings from the transformed examples from $\mathcal{D}_Q$ to the base unperturbed images from $P$. In general, the task of inverting adversarial images may be a difficult one as adversarial attacks can be complex transformations, possibly severely distorting the canonical images they are based upon. Keeping this in mind, the learned mappings may not necessarily preserve the visual features but instead preserve features necessary for a neural network classifier to recover the correct label unseen by the inverse mappings.

An important point is that $\mathcal{D}_Q$ is constructed by randomly applying one of the attacks $A_1, \dots, A_N$ to images from $P$. $\mathcal{D}_P$ contains images from $P$, other than the ones used to construct $\mathcal{D}_Q$.

\subsection{Competitive generators}
\label{subsection:competitive-generators}

We consider a set of $N^{\prime}$ functions $G_{1}, \dots, G_{N^{\prime}}$, which will be referred to as \emph{generators}, parameterized by $\theta_{1}, \dots, \theta_{N^{\prime}}$. In principle, we do not require $N = N^{\prime}$. We additionally consider a function $D: \mathbb{R}^{d} \to \left[0, 1\right]$, a \emph{discriminator}, parameterized by $\theta_{D}$ which is required to take values close to $0$ for input outside of the support of the canonical distribution and values close to $1$ for input from within this support.

The training is an adversarial procedure inspired by the training process of GANs \citep{Goodfellow2014}. Algorithm \ref{alg} outlines the training procedure. For each input example $x^{\prime} \in \mathcal{D}_{Q}$, each generator $G_j$ is conditioned on this input. Based on the evaluation of the discriminator $D(G_{j}(x^{\prime}))$, the generator with the highest score receives an update to its parameters $\theta_{j^{*}}$, where $j^{*} = \argmax{D(G_{j}(x^{\prime}))}$. Following that, $D$ receives an update to its parameters $\theta_{D}$ based on the output of all the generators and on the samples from the canonical distribution. The optimization problem can be stated as:

\begin{equation} \label{eq:1}
    \theta^{*}_{1}, \dots, \theta^{*}_{N^{\prime}} = \argmax_{\theta_{1}, \dots, \theta_{N^{\prime}}}{\mathbb{E}_{x^{\prime} \sim Q} \left( \max_{j \in \{ 1, \dots, N^{\prime} \}} D(G_{\theta_{j}}(x^{\prime})) \right)}.
\end{equation}

The discriminator is trained to maximize the objective:

\begin{equation} \label{eq:2}
    \max_{\theta_{D}} \left( \mathbb{E}_{x \sim P} \log{(D_{\theta_{D}})} + \frac{1}{N^{\prime}} \sum_{j=1}^{N^{\prime}} \mathbb{E}_{x^{\prime} \sim Q} \left( \log{(1 - D_{\theta_{D}}(G_{\theta_{j}}(x^{\prime}) ) )} \right) \right)
\end{equation}

\begin{algorithm}
\caption{Mixture of generators adversarial defense training algorithm}\label{alg}
Choose $A_1, \dots, A_N$ as adversarial attacks.
\begin{algorithmic}
\For{fixed number of initialization epochs}
\State Train $G_{1}, \dots, G_{N^{\prime}}$ to approximate the identity transformation on adversarial images.
\EndFor
\For{fixed number of training epochs}
\State Select batch of canonical images from train set.
\State Use attack $A_i$ selected at random to generate adversarial images from the batch of canonical images.
\State Feed the adversarial images to $G_{1}, \dots, G_{N^{\prime}}$ to produce new images.
\State Provide the produced images to $D$ to produce scores for each image.
\State For the highest-scored generator $G_j$, update its weights against $D$.
\State Update the weights of $D$ based on the images produced by all the generators $G_{1}, \dots, G_{N^{\prime}}$ as well as on canonical images.
\EndFor
\end{algorithmic}
\end{algorithm}

\section{Experiments}
\label{experiments}

\begin{wraptable}[19]{r}{8cm}
  \caption{Types of adversarial attacks. $\epsilon$ denotes a hyperparameter controlling the strength of the attack. An attack is considered successful if the LeNet5 classification of the attack image is incorrect}
  \label{tab:attacks}
  \centering
  \begin{tabular}{l*2{c}}
    \toprule
    Attack & $\epsilon$ & Success \\
    \midrule
    fast gradient sign method (FGSM) & $0.5$ & $89.8\%$ \\
    projected gradient descent (PGD) & $0.5$ & $100.0\%$ \\
    DeepFool (DF) & $0.5$ & $100.0\%$ \\
    additive uniform noise (AUN) & $3.5$ & $90.6\%$ \\
    basic iterative method (BIM) & $0.2$ & $90.6\%$ \\
    additive Gaussian noise (AGN) & $100$ & $90.6\%$ \\
    repeated AGN (RAGN) & $15$ & $94.5\%$ \\
    salt and pepper noise (SAPN) & $10$ & $90.6\%$ \\
    sparse L1-descent (SLIDE) & $25$ & $88.3\%$ \\
    \bottomrule
  \end{tabular}
\end{wraptable}

We evaluate the potential of the described system to recover the features necessary for correctly classifying adversarial images from the MNIST dataset. This is done by choosing the generators and the discriminator to be deep neural networks. The generators are fully-convolutional networks \citep{Long2015} designed to preserve the size of the input data. The discriminator is a convolutional neural network with an increasing number of filters. Both network types use ELU activations \citep{Clevert2016} and batch normalization \citep{Ioffe2015}. The architectural details of the networks are presented in the supplementary material.

Raw $28 \times 28$ pixel MNIST images are scaled to the $(0, 1)$ range. A separate LeNet5-inspired classifier  \citep{Lecun1998} is pretrained on the MNIST train set for $100$ epochs with Adam, with an initial learning rate of $10^{-3}$. This architecture is chosen for its simplicity and relatively high accuracy. The network achieves a $98.7\%$ accuracy on the MNIST test set. The MNIST train set is divided equally into two disjoint train subsets. The first one serves as the canonical train set, which is not transformed further. The images from the other one are subject to adversarial attacks and form the transformed train set. It is worth stressing that there is no overlap between the canonical and transformed datasets, i.e. no images from the canonical train set are the basis for the images in the transformed dataset. Additionally, the system is unsupervised in that it does not use labels at train or test time.

\begin{figure}
     \centering
     \begin{subfigure}[b]{0.1\textwidth}
         \centering
         \includegraphics[width=\textwidth]{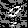}
         \caption{FGSM}
         \label{fig:fgsm}
     \end{subfigure}
     \hfill
     \begin{subfigure}[b]{0.1\textwidth}
         \centering
         \includegraphics[width=\textwidth]{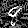}
         \caption{PGD}
         \label{fig:pgd}
     \end{subfigure}
     \hfill
     \begin{subfigure}[b]{0.1\textwidth}
         \centering
         \includegraphics[width=\textwidth]{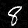}
         \caption{DF}
         \label{fig:deep-fool}
     \end{subfigure}
     \hfill
     \begin{subfigure}[b]{0.1\textwidth}
         \centering
         \includegraphics[width=\textwidth]{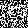}
         \caption{AUN}
         \label{fig:additive-uniform-noise-attack}
     \end{subfigure}
     \hfill
     \begin{subfigure}[b]{0.1\textwidth}
         \centering
         \includegraphics[width=\textwidth]{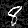}
         \caption{BIM}
         \label{fig:basic-iterative-method}
     \end{subfigure}
     \hfill
     \begin{subfigure}[b]{0.1\textwidth}
         \centering
         \includegraphics[width=\textwidth]{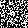}
         \caption{AGN}
         \label{fig:additive-gaussian-noise-attack}
     \end{subfigure}
     \hfill
     \begin{subfigure}[b]{0.1\textwidth}
         \centering
         \includegraphics[width=\textwidth]{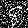}
         \caption{RAGN}
         \label{fig:repeated-additive-gaussian-noise-attack}
     \end{subfigure}
     \hfill
     \begin{subfigure}[b]{0.1\textwidth}
         \centering
         \includegraphics[width=\textwidth]{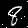}
         \caption{SAPN}
         \label{fig:salt-and-pepper-noise-attack}
     \end{subfigure}
     \hfill
     \begin{subfigure}[b]{0.1\textwidth}
         \centering
         \includegraphics[width=\textwidth]{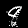}
         \caption{SLIDE}
         \label{fig:sparse-l1-descent-attack}
     \end{subfigure}
        \caption{Sample attack images}
        \label{fig:sample-attack-images}
\end{figure}

The full list of potential attacks is given in Table~\ref{tab:attacks} along with $\epsilon$ hyperparameters and the success rate of an attack on a random batch of $128$ clean images from the MNIST train set. The attacks employed result in distinct visual distortions to the original images, while retaining relatively high success rates. Figure~\ref{fig:sample-attack-images} shows the effects of applying the attacks to a random image from the MNIST train set. While most of the images are still recognizable to a human observer, the additive uniform noise and additive Gaussian noise attacks are not. Based on this, it can be expected that any system trying to recover original images from adversarial ones would have a hard time on these two attacks.

\subsection{Training}
We train our system by first initializing the generators over $10$ epochs on the transformed images to approximate an identity transformation. This \emph{identity initialization} procedure helps to stabilize training and partially mitigates the problem of one generator dominating the other ones \citep{Parascandolo2018}. Once identity initialization is complete, the system is trained for $100$ epochs following the procedure outlined in Section \ref{subsection:competitive-generators}. Adversarial images are fed into all the generators, which produce new images based on the input. The output of all the generators is then scored by the discriminator. The generator which achieves the highest score is then updated accordingly, as would be the case in a standard GAN training procedure. The rest of the generators remain unchanged (\ref{eq:1}). The discriminator additionally receives canonical images and scores them. It is then updated against the output of all the generators and based on the scores of the canonical images (\ref{eq:2}). We use Adam with an initial learning rate of $10^{-3}$ for all networks in both identity initialization and actual training.

\subsection{Single-attack settings}
In the first experiment, we train the system in single-attack settings with one generator. We repeat this for each attack separately. This means that the generator only receives images transformed by one attack in each setting. We evaluate the system on the previously-unseen MNIST test set by feeding transformed images to the generators, scoring the output of the generators by the discriminator and passing the highest-scored output on to the pretrained LeNet5 classifier. We measure the accuracy of the predictions in terms of the percentage of the correctly classified images. The results are presented in Table~\ref{tab:generators-single}. For most attacks, the accuracy of the classification after the defense is above $90\%$, which highlights the possibility to use this approach as a successful adversarial defense mechanism. We also see that the success of the method seems to be related to the degree of visual distortions in the transformed images. The attacks for which our method generated less satisfactory results are also the ones which generate the most distorted images (the additive uniform noise and additive Gaussian noise attacks). Interestingly, for the DeepFool attack our method was able to marginally improve the original accuracy of the pretrained LeNet5 classifier on the test set. This suggests that for this particular attack our method comes close to completely removing the impact of the adversarial attack on classification.

\begin{table}
  \caption{LeNet5 accuracy after defense in single attack/generator setting}
  \label{tab:generators-single}
  \centering
  \begin{tabular}{l*9{c}}
    \toprule
    Attack & FGSM & PGD & DF & AUN & BIM & AGN & RAGN & SAPN & SLIDE \\
    \midrule
    Accuracy & $97.5\%$ & $93.5\%$ & $98.9\%$ & $18.9\%$ & $97.5\%$ & $11.4\%$ & $92.9\%$ & $98.1\%$ & $95.3\%$ \\
    \bottomrule
  \end{tabular}
\end{table}

\begin{wraptable}{r}{7cm}
  \caption{LeNet5 accuracy after defense with multiple attacks/generators. Additional results are presented in the supplementary material}
  \label{tab:generators-multi}
  \centering
  \begin{tabular}{l*3{c}}
    \toprule
    Number of attacks & $3$ & $5$ & $9$ \\
    \midrule
    Trained separately & $94.9\%$ & $73.9\%$ & $73.9\%$ \\
    Trained jointly & $84.4\%$ & $64.2\%$ & $63.3\%$ \\
    Faster initialization & $81.2\%$ & $69.3\%$ & $62.5\%$ \\
    \bottomrule
  \end{tabular}
\end{wraptable}

\subsection{Multi-attack settings}
Given the results in the single-attack setting, we evaluate our system in settings with multiple attacks and generators. In the following experiments, we have chosen the number of attacks to be equal to the number of generators. This is not a hard and fast rule, and we could conceivably include a larger number of generators. In preliminary tests, we observed that limiting the number of generators can have adverse effects on the performance of the system but using a number of generators slightly lower than the number of attacks generally does not lead to a breakdown in performance. For simplicity, we proceed with a number of generators equal to the number of attacks.

We train and evaluate our system in three multi-attack settings: \textbf{(a)} $3$ generators, $3$ attacks: FGSM, PGD, DeepFool, \textbf{(b)} $5$ generators, $5$ attacks: FGSM, PGD, DeepFool, AUN, BIM, \textbf{(c)} $9$ generators, $9$ attacks: all the attacks listed in Table~\ref{tab:attacks}.

For each of these settings, it is possible to adopt two approaches to training the architecture. In the first one, the system is trained jointly, end-to-end. This means that at train time neither of the generators has access to the information about the type of an attack applied to the images. In the other approach, we simply reuse the generators trained in single-attack settings and the discriminator trained in the first multi-attack approach. The main difference is that in the latter setting the generators do have information about the particular attack at train time. Such an adversarial defense is inherently an easier task - a reality reflected in the results presented in Table~\ref{tab:generators-multi}. A system trained separately (top row) consistently outperforms the system trained jointly (middle row) to the tune of a $10\%$ difference in accuracy after the defense. However, our main interest is in the jointly-trained system, which does not have access to the type of attack at train time and can be trained in a fully unsupervised manner.

The results for the $3$-attack version of the jointly-trained system are comparable to other adversarial defense methods~\citep{Schott2019} evaluated in single-attack settings. The advantage of our method is that it can be applied to multiple attacks at once with no supervision. Additionally, it can be used in a modular manner with pretrained classifiers. For the $5$- and $9$-attack versions of this system, the accuracy drops as the defense becomes increasingly difficult, but the negative impact of increasing the number of attacks from $5$ to $9$ is visibly less pronounced than in the case of moving from $3$ to $5$ attacks.

\subsection{Faster initialization of the generators}
The modular nature of the set of generators allows to potentially limit the computation time spent in the identity initialization phase. Since the goal of this phase is to bring all generators to an equal footing, it is possible to only initialize one generator, perturb its weights randomly and treat such random perturbations as new generators in the actual training procedure. We train such variants of the system where all the generators at the start of the actual training are copies of one identity-initialized generator with $5\%$ of their weights perturbed by randomly setting them to $0$. The rest of the training procedure remains unchanged. The results of this approach are presented in Table~\ref{tab:generators-multi} (bottom row). The use of randomly-perturbed generators leads to qualitatively similar results, but allows to only initialize one generator instead of a full set of generators before the actual training commences.

\begin{figure}
     \centering
     \begin{subfigure}[b]{0.1\textwidth}
         \centering
         \includegraphics[width=\textwidth]{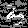}
     \end{subfigure}
     \hfill
     \begin{subfigure}[b]{0.1\textwidth}
         \centering
         \includegraphics[width=\textwidth]{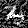}
     \end{subfigure}
     \hfill
     \begin{subfigure}[b]{0.1\textwidth}
         \centering
         \includegraphics[width=\textwidth]{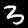}
     \end{subfigure}
     \hfill
     \begin{subfigure}[b]{0.1\textwidth}
         \centering
         \includegraphics[width=\textwidth]{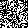}
     \end{subfigure}
     \hfill
     \begin{subfigure}[b]{0.1\textwidth}
         \centering
         \includegraphics[width=\textwidth]{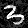}
     \end{subfigure}
     \hfill
     \begin{subfigure}[b]{0.1\textwidth}
         \centering
         \includegraphics[width=\textwidth]{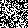}
     \end{subfigure}
     \hfill
     \begin{subfigure}[b]{0.1\textwidth}
         \centering
         \includegraphics[width=\textwidth]{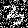}
     \end{subfigure}
     \hfill
     \begin{subfigure}[b]{0.1\textwidth}
         \centering
         \includegraphics[width=\textwidth]{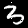}
     \end{subfigure}
     \hfill
     \begin{subfigure}[b]{0.1\textwidth}
         \centering
         \includegraphics[width=\textwidth]{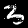}
     \end{subfigure}
     \begin{subfigure}[b]{0.1\textwidth}
         \centering
         \includegraphics[width=\textwidth]{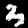}
     \end{subfigure}
     \hfill
     \begin{subfigure}[b]{0.1\textwidth}
         \centering
         \includegraphics[width=\textwidth]{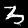}
     \end{subfigure}
     \hfill
     \begin{subfigure}[b]{0.1\textwidth}
         \centering
         \includegraphics[width=\textwidth]{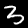}
     \end{subfigure}
     \hfill
     \begin{subfigure}[b]{0.1\textwidth}
         \centering
         \includegraphics[width=\textwidth]{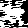}
     \end{subfigure}
     \hfill
     \begin{subfigure}[b]{0.1\textwidth}
         \centering
         \includegraphics[width=\textwidth]{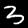}
     \end{subfigure}
     \hfill
     \begin{subfigure}[b]{0.1\textwidth}
         \centering
         \includegraphics[width=\textwidth]{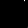}
     \end{subfigure}
     \hfill
     \begin{subfigure}[b]{0.1\textwidth}
         \centering
         \includegraphics[width=\textwidth]{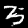}
     \end{subfigure}
     \hfill
     \begin{subfigure}[b]{0.1\textwidth}
         \centering
         \includegraphics[width=\textwidth]{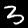}
     \end{subfigure}
     \hfill
     \begin{subfigure}[b]{0.1\textwidth}
         \centering
         \includegraphics[width=\textwidth]{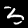}
     \end{subfigure}
     \rule{\textwidth}{0pt}
     \begin{subfigure}[b]{0.1\textwidth}
         \centering
         \includegraphics[width=\textwidth]{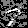}
     \end{subfigure}
     \hfill
     \begin{subfigure}[b]{0.1\textwidth}
         \centering
         \includegraphics[width=\textwidth]{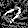}
     \end{subfigure}
     \hfill
     \begin{subfigure}[b]{0.1\textwidth}
         \centering
         \includegraphics[width=\textwidth]{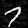}
     \end{subfigure}
     \hfill
     \begin{subfigure}[b]{0.1\textwidth}
         \centering
         \includegraphics[width=\textwidth]{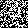}
     \end{subfigure}
     \hfill
     \begin{subfigure}[b]{0.1\textwidth}
         \centering
         \includegraphics[width=\textwidth]{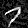}
     \end{subfigure}
     \hfill
     \begin{subfigure}[b]{0.1\textwidth}
         \centering
         \includegraphics[width=\textwidth]{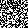}
     \end{subfigure}
     \hfill
     \begin{subfigure}[b]{0.1\textwidth}
         \centering
         \includegraphics[width=\textwidth]{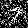}
     \end{subfigure}
     \hfill
     \begin{subfigure}[b]{0.1\textwidth}
         \centering
         \includegraphics[width=\textwidth]{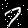}
     \end{subfigure}
     \hfill
     \begin{subfigure}[b]{0.1\textwidth}
         \centering
         \includegraphics[width=\textwidth]{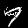}
     \end{subfigure}
     \begin{subfigure}[b]{0.1\textwidth}
         \centering
         \includegraphics[width=\textwidth]{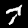}
         \caption{FGSM}
         \label{fig:fgsm-attack-defense}
     \end{subfigure}
     \hfill
     \begin{subfigure}[b]{0.1\textwidth}
         \centering
         \includegraphics[width=\textwidth]{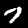}
         \caption{PGD}
         \label{fig:pgd-attack-defense}
     \end{subfigure}
     \hfill
     \begin{subfigure}[b]{0.1\textwidth}
         \centering
         \includegraphics[width=\textwidth]{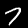}
         \caption{DF}
         \label{fig:deep-fool-attack-defense}
     \end{subfigure}
     \hfill
     \begin{subfigure}[b]{0.1\textwidth}
         \centering
         \includegraphics[width=\textwidth]{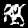}
         \caption{AUN}
         \label{fig:additive-uniform-noise-attack-defense}
     \end{subfigure}
     \hfill
     \begin{subfigure}[b]{0.1\textwidth}
         \centering
         \includegraphics[width=\textwidth]{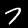}
         \caption{BIM}
         \label{fig:basic-iterative-method-attack-defense}
     \end{subfigure}
     \hfill
     \begin{subfigure}[b]{0.1\textwidth}
         \centering
         \includegraphics[width=\textwidth]{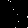}
         \caption{AGN}
         \label{fig:additive-gaussian-noise-attack-defense}
     \end{subfigure}
     \hfill
     \begin{subfigure}[b]{0.1\textwidth}
         \centering
         \includegraphics[width=\textwidth]{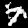}
         \caption{RAGN}
         \label{fig:repeated-additive-gaussian-noise-attack-defense}
     \end{subfigure}
     \hfill
     \begin{subfigure}[b]{0.1\textwidth}
         \centering
         \includegraphics[width=\textwidth]{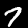}
         \caption{SAPN}
         \label{fig:salt-and-pepper-noise-attack-defense}
     \end{subfigure}
     \hfill
     \begin{subfigure}[b]{0.1\textwidth}
         \centering
         \includegraphics[width=\textwidth]{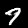}
         \caption{SLIDE}
         \label{fig:sparse-l1-descent-attack-defense}
     \end{subfigure}
        \caption{Sample attacks (first and third row) with the respective defenses (second and fourth row)}
        \label{fig:sample-attacks-defenses}
\end{figure}

The defenses generated by our system tend to reverse the effects of the adversarial attack, at least as far as a simple eyeball test is able to determine. The attack images based on one sample from the MNIST train set are presented in Figure~\ref{fig:sample-attacks-defenses} (first and third row) along with the images generated by our $9$-attack system trained jointly (second and fourth row). Each defense is the image generated by the best-scored generator. We see that for less distorted attack images our system is generally able to recover features corresponding to the clean image. In such cases, it does seem that the system is able to approximate the inversion of a particular attack. For more distorted images, for instance those generated by the AUN and AGN attacks, the system does not seem to approximate the inversion closely.

\subsection{Large generator approach}
One question that can be asked is how a mixture system fares against a single generator with similar capacity. Each of our generators has $28,609$ trainable weights. For a system consisting of $9$ generators, this gives $257,481$ trainable weights. We construct a system with one generator with $333,697$ trainable weights to match the capacity of $9$ standard generators. The details of the architecture are presented in the supplementary material.

We train a system with this large generator consistently with our standard training procedure for $9$ attacks. This slightly more powerful system achieves an accuracy of $66.7\%$ - similar to the accuracy achieved by a mixture of generators. However, the advantage of training a mixture vs. a single large net is that the elements of the system can be inspected more easily. On top of that, this allows for more flexibility in the setup of the defense and it opens the possibility to reuse elements of the system.

\begin{figure}[ht]
     \centering
     \begin{subfigure}[b]{0.49\textwidth}
         \centering
         \includegraphics[width=\textwidth]{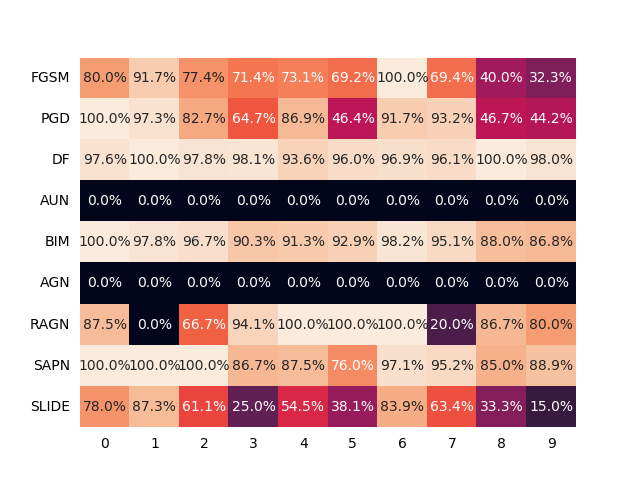}
     \end{subfigure}
     \hfill
     \begin{subfigure}[b]{0.49\textwidth}
         \centering
         \includegraphics[width=\textwidth]{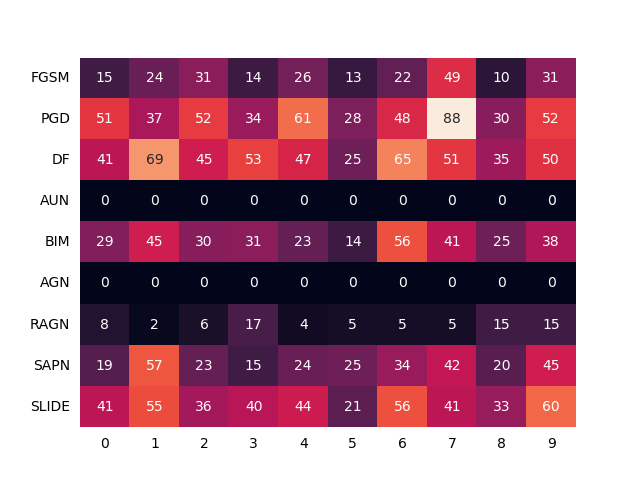}
     \end{subfigure}
     \hfill
     \begin{subfigure}[b]{0.49\textwidth}
         \centering
         \includegraphics[width=\textwidth]{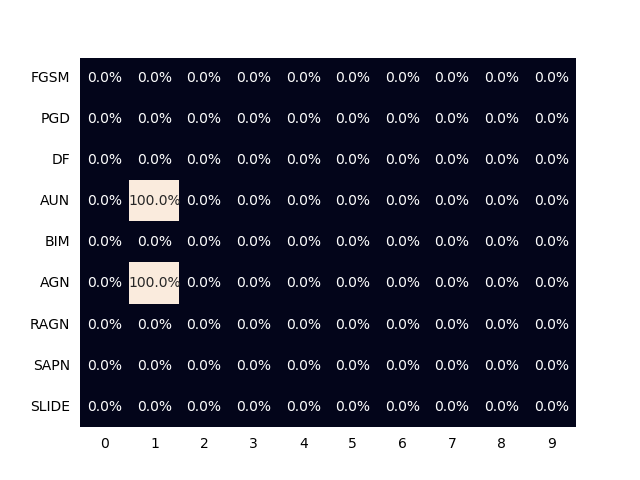}
     \end{subfigure}
     \hfill
     \begin{subfigure}[b]{0.49\textwidth}
         \centering
         \includegraphics[width=\textwidth]{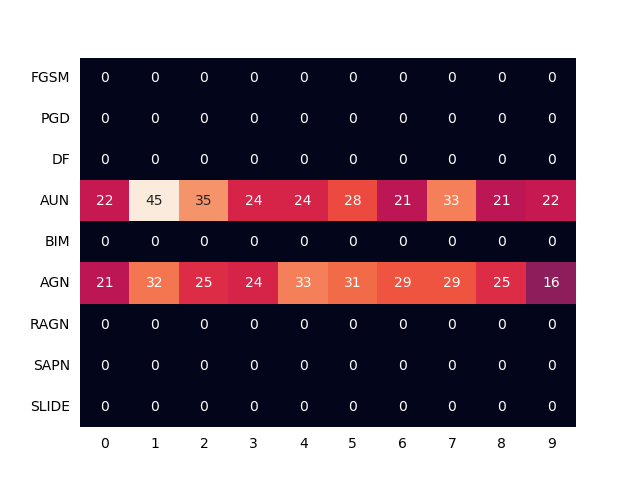}
     \end{subfigure}
     \hfill
     \begin{subfigure}[b]{0.49\textwidth}
         \centering
         \includegraphics[width=\textwidth]{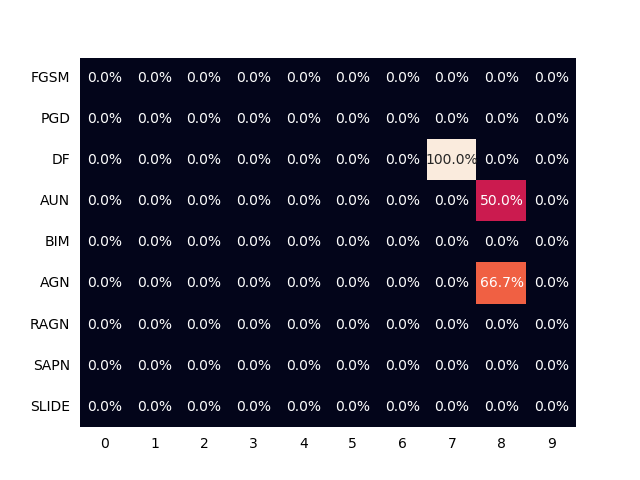}
     \end{subfigure}
     \hfill
     \begin{subfigure}[b]{0.49\textwidth}
         \centering
         \includegraphics[width=\textwidth]{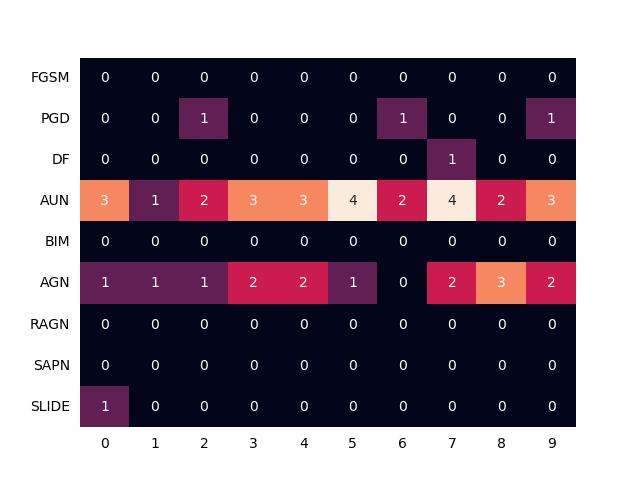}
     \end{subfigure}
        \caption{Examples of generator specialization: \emph{generalist} (top row), \emph{specialist} (middle row) and \emph{marginalist} (bottom row). The heatmaps present the accuracy for a given class/attack (left) and the number of samples on which a given generator won (right)}
        \label{fig:generator-specialization}
\end{figure}

\subsection{Generator \emph{expertise}}
An interesting aspect of the mixture of generators is the extent to which particular generators specialize in either specific attacks or image classes. We perform a breakdown of the accuracy measure by digit class and by attack for the $9$-attack jointly-trained system. In Figure~\ref{fig:generator-specialization}, we present the results for $3$ out of the $9$ generators to highlight the potential types of generators trained.

The first generator (top row) is a \emph{generalist} - it wins on a large number of sample images and achieves relatively high accuracy on the majority of the attacks. Notably, the two hard attacks we have identified so far (additive uniform noise and additive Gaussian noise attacks) are precisely the attacks this generator does not cover - it does not win on a single sample image from these attacks. \emph{Generalists} are the most ubiquitous among the $9$ trained generators.

The second generator only ever wins on samples from the two hard attacks. This generator is a \emph{specialist} in that it focuses on specific kinds of attacks at the expense of other ones. The accuracy of the \emph{specialist} is very high in only a handful of specific cases.

Finally, the third generator is a \emph{marginalist} - much like the \emph{specialist}, it focuses on two hard attacks but fails to win on a significant number of samples and has questionable accuracy levels.

\section{Conclusions}

In this work we show that it is possible to train a GAN-based mixture of generators to defend against adversarial examples as a data preprocessing step. The system is fully unsupervised in that it does not use attack or digit labels at all at train time. This system, trained adversarially with competing generators, is able to extract from adversarially transformed images features relevant for classification in both single-attack and multi-attacks settings. A system with $3$ generators is able to defend against $3$ adversarial attacks with results comparable with other modern single-attack adversarial defenses. At first, increasing the number of attacks to defend against, lowers the post-defense classification accuracy but the results suggest that extending the system further from $5$ to $9$ attacks/generators does not significantly decrease the accuracy.

The results show that for a majority of tested attacks, the system is not only able to recover statistical data features important for classification but that the generators also approximate the inverses of the adversarial transformations, at least on a visual level. The images recovered by our system retain internal visual coherence and are immediately interpretable by a human observer, except for those generated by two attacks which severely distort the original images (AUN, AGN).

The system is able to internally distinguish between these hard attacks and other attacks, which produce adversarial images that are still recognizable to humans. The distinction between these types of attacks is visible in the process of generator specialization. Various generators specialize in various parts of the data space and the training process is able to produce more general generators, which mainly operate on easier attacks, and more specialized generators, which in turn tend to focus on harder attacks with different levels of success.

We show that infusing the system with knowledge as to the type of the attack by training generators separately increases the post-defense accuracy by about $10\%$ in $3$-, $5$- and $9$-attack settings. It is also possible to limit the computational intensity of the system by copying and randomly perturbing one initialized generator without materially impacting the accuracy. The results for a more compartmentalized $9$-attack/generator system are on par with the results for a similar system designed with one large generator. Given the increased interpretability and modularity of the mixture approach, it does seem that a modular approach might be better suited to the problem of inverting adversarial attacks.

Our research links adversarial defenses with ensemble methods and GAN-based training. From a more general point of view, it provides a fully-unsupervised approach to defending against adversarial examples without modifying the underlying classifier. This could provide for a practically-relevant method to approach the problem of adversarial examples in a real-life setting, particularly when the type of attack is not known beforehand and it is not feasible to modify the classifier.

This work also highlights interesting areas for further research. In particular, it hints at the possibility of linking adversarial defenses with research on causality. It would be particularly interesting to see whether it is possible to further specialize the components of ensemble systems to reveal the underlying causal structure behind adversarial attacks.

\section*{Acknowledgment}
This project was funded by the POB Research Centre Cybersecurity and Data Science of the Warsaw University of Technology within the Excellence Initiative Program - Research University (ID-UB).

\bibliography{ms}
\bibliographystyle{apalike}

\clearpage

\section*{Supplementary material}

\appendix

\section{Architectural details}

For brevity, we omitted the details of the system architecture in the main paper. The architectures of the generators and the discriminator are presented in Table~\ref{tab:architecture}. The generators are fully-convolutional deep neural networks with exponential linear units (ELUs) and batch normalization. They take an image of a given size as input and produce an image of the same size as output. In principle, each generator follows the logic of a conditional GAN generator, where the network is conditioned on the input. In this concrete case, the generator is conditioned on an image to produce an image of the same size. The goal is to invert the adversarial attacks and generate clean images.

Additionally, in Table~\ref{tab:architecture-large} we present the architecture of the large generator discussed in Section 4.5 of the main paper. The large generator preserves the general characteristics of the individual generators by employing the same kernel size, padding, activation functions and batch normalization. The main difference between the large network and the individual generators is the number of layers and the fact that for the large generator the number of channels increases toward the middle layers and then decreases toward the output layer. This architecture is constructed so as to contain a number of parameters comparable to $9$ individual generators. The aim is to test whether separate generators are able to learn comparably well as one large network while offering more interpretability.

\begin{table}[h]
  \caption{Generator/discriminator architectures}
  \label{tab:architecture}
  \centering
  \begin{tabular}{ll}
    \toprule
    CNN Generators & CNN Discriminator \\
    \midrule
    Input 28x28 & Input 28x28 \\
    Conv 3x3, 32, padding=1, ELU & 3 x Conv 3x3, 16, padding=1, ELU \\
    BatchNorm & AvgPool 2x2 \\
    Conv 3x3, 32, padding=1, ELU & 2 x Conv 3x3, 32, padding=1, ELU \\
    BatchNorm & AvgPool 2x2 \\
    Conv 3x3, 32, padding=1, ELU  & 2 x Conv 3x3, 64, padding=1, ELU \\
    BatchNorm & AvgPool 2x2 \\
    Conv 3x3, 32, padding=1, ELU & FC 1024, ELU \\
    BatchNorm & FC 1, Sigmoid \\
    Conv 3x3, 1, padding=1, Sigmoid \\
    \bottomrule
  \end{tabular}
\end{table}

\begin{table}[h]
  \caption{Large generator architecture}
  \label{tab:architecture-large}
  \centering
  \begin{tabular}{l}
    \toprule
    CNN Generator \\
    \midrule
    Input 28x28 \\
    Conv 3x3, 32, padding=1, ELU \\
    BatchNorm \\
    Conv 3x3, 64, padding=1, ELU \\
    BatchNorm \\
    Conv 3x3, 128, padding=1, ELU \\
    BatchNorm \\
    Conv 3x3, 128, padding=1, ELU \\
    BatchNorm \\
    Conv 3x3, 64, padding=1, ELU \\
    BatchNorm \\
    Conv 3x3, 32, padding=1, ELU \\
    BatchNorm \\
    Conv 3x3, 1, padding=1, Sigmoid \\
    \bottomrule
  \end{tabular}
\end{table}

\subsection{Classification accuracy after defenses in repeated experiments}
In the interest of verifying the stability of the obtained results, we repeat the experiments for multi-attack settings. The results of these repeated experiments are presented in Table~\ref{tab:repeated}. In particular, we train the basic joint setup twice for $3$, $5$ and $9$ attacks. Given that the results adhere to what we obtained previously, we further focus on the \emph{faster initialization} setting. We repeat the experiments with faster initialization for $3$, $5$ and $9$ attacks $10$ times. Results for each individual experiment are presented along with means and standard errors. The outcomes show that the stability of the experiments increases with the number of generators. While we see some variability for the $3$-attack settings, it decreases for the $5$- and $9$-attack settings, which suggests that the results are relatively stable, in particular for the settings with a larger number of attacks.

\begin{table}[h]
  \caption{LeNet5 accuracy after defense, multiple attacks/generators. Means for the \emph{faster initialization} method are presented with standard errors.}
  \label{tab:repeated}
  \centering
  \begin{tabular}{l*3{c}}
    \toprule
    Number of attacks & $3$ & $5$ & $9$ \\
    \midrule
    Trained jointly & & & \\
    \midrule
    Experiment 1 & $74.1\%$ & $61.1\%$ & $65.1\%$ \\
    Experiment 2 & $82.7\%$ & $69.8\%$ & $57.7\%$ \\
    \midrule
    Trained jointly - faster initialization & & & \\
    \midrule
    Experiment 1 & $83.6\%$ & $63.6\%$ & $65.1\%$ \\
    Experiment 2 & $74.7\%$ & $68.0\%$ & $66.4\%$ \\
    Experiment 3 & $81.6\%$ & $66.3\%$ & $60.9\%$ \\
    Experiment 4 & $60.6\%$ & $66.4\%$ & $66.3\%$ \\
    Experiment 5 & $81.8\%$ & $69.0\%$ & $60.9\%$ \\
    Experiment 6 & $82.6\%$ & $69.0\%$ & $63.3\%$ \\
    Experiment 7 & $72.0\%$ & $71.2\%$ & $62.0\%$ \\
    Experiment 8 & $67.0\%$ & $65.7\%$ & $63.5\%$ \\
    Experiment 9 & $67.4\%$ & $67.1\%$ & $61.6\%$ \\
    Experiment 10 & $70.3\%$ & $70.2\%$ & $61.8\%$ \\
    \midrule
    Mean & $74.2 \pm 8.0\%$ & $67.7 \pm 2.3\%$ & $63.2 \pm 2.1\%$ \\
    \bottomrule
  \end{tabular}
\end{table}

\subsection{Generator analysis for single attacks}\label{sec:single}
\begin{figure}[ht]
     \centering
     \begin{subfigure}[b]{0.49\textwidth}
         \centering
         \includegraphics[width=\textwidth]{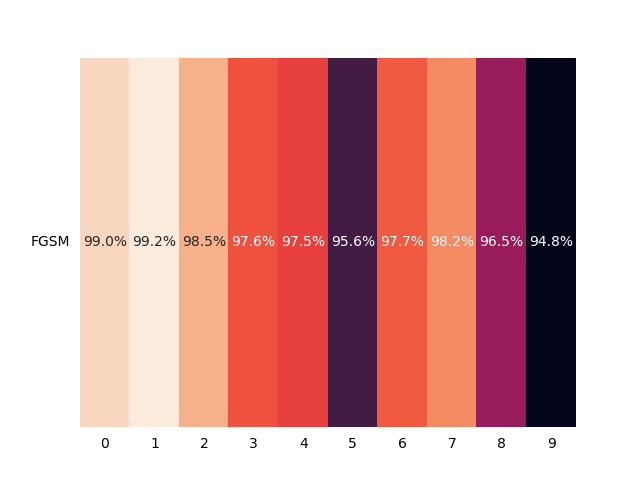}
     \end{subfigure}
     \hfill
     \begin{subfigure}[b]{0.49\textwidth}
         \centering
         \includegraphics[width=\textwidth]{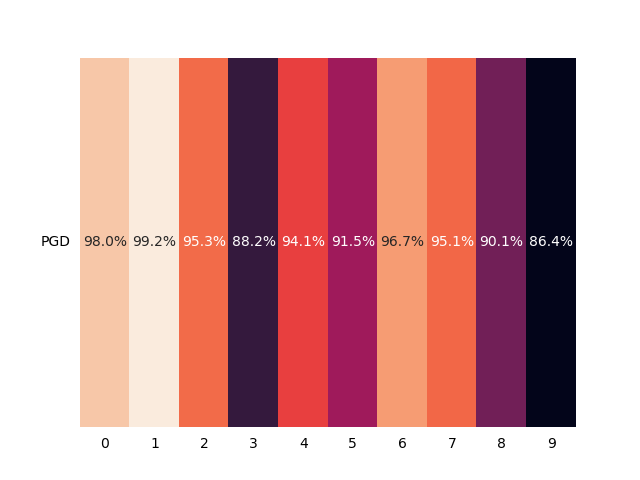}
     \end{subfigure}
     \hfill
     \begin{subfigure}[b]{0.49\textwidth}
         \centering
         \includegraphics[width=\textwidth]{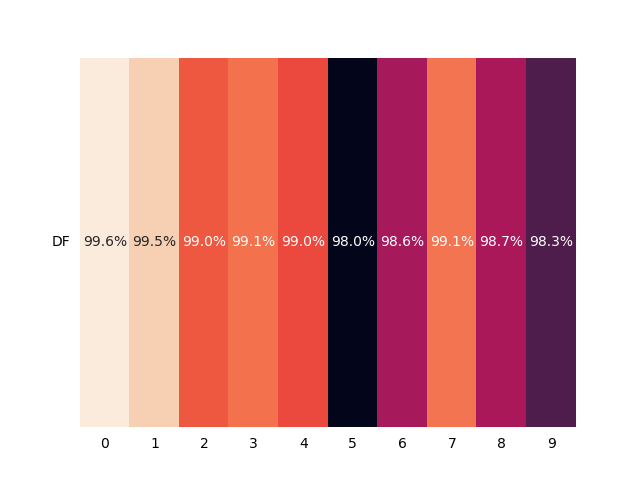}
     \end{subfigure}
     \hfill
     \begin{subfigure}[b]{0.49\textwidth}
         \centering
         \includegraphics[width=\textwidth]{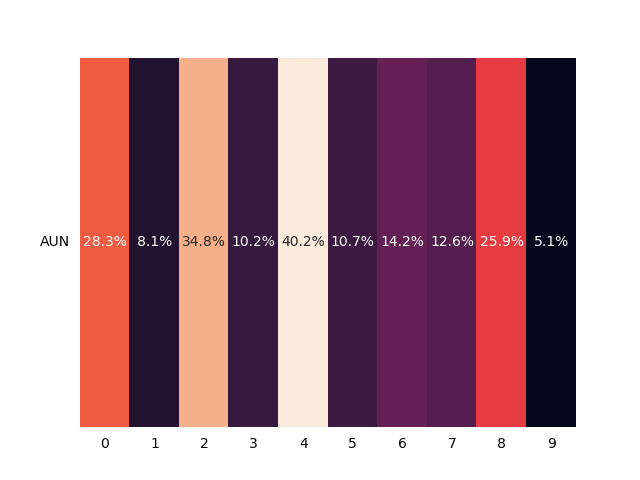}
     \end{subfigure}
     \hfill
     \begin{subfigure}[b]{0.49\textwidth}
         \centering
         \includegraphics[width=\textwidth]{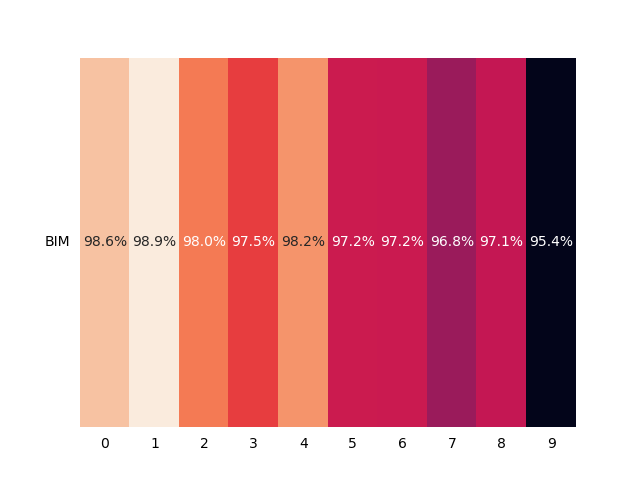}
     \end{subfigure}
        \caption{Class accuracy breakdown for single-attack settings, part 1 (generators $1$ to $5$). Each heatmap corresponds to one generator. The heatmaps present the accuracy for a given class.}
        \label{fig:attacks-single-p1}
\end{figure}

\begin{figure}[ht]
     \centering
     \begin{subfigure}[b]{0.49\textwidth}
         \centering
         \includegraphics[width=\textwidth]{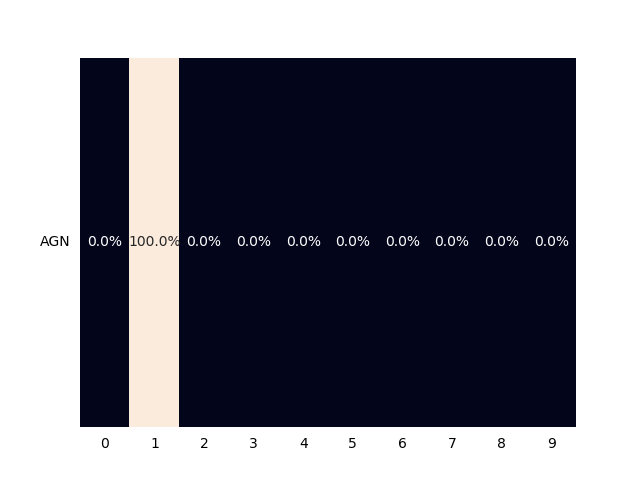}
     \end{subfigure}
     \hfill
     \begin{subfigure}[b]{0.49\textwidth}
         \centering
         \includegraphics[width=\textwidth]{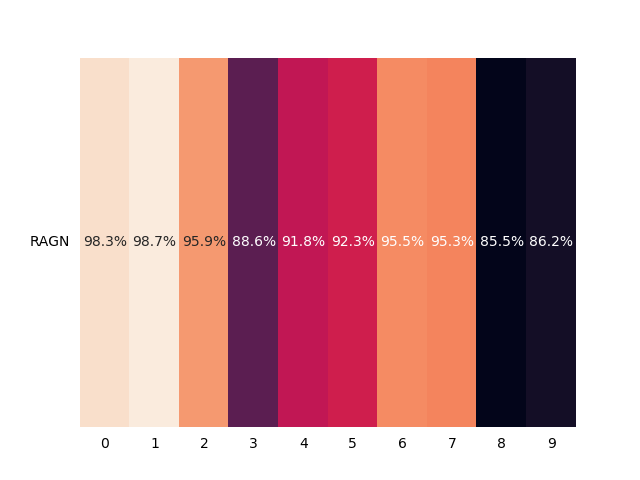}
     \end{subfigure}
     \hfill
     \begin{subfigure}[b]{0.49\textwidth}
         \centering
         \includegraphics[width=\textwidth]{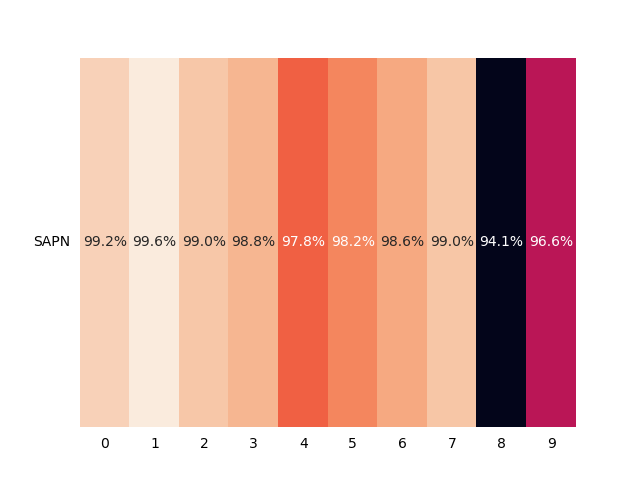}
     \end{subfigure}
     \hfill
     \begin{subfigure}[b]{0.49\textwidth}
         \centering
         \includegraphics[width=\textwidth]{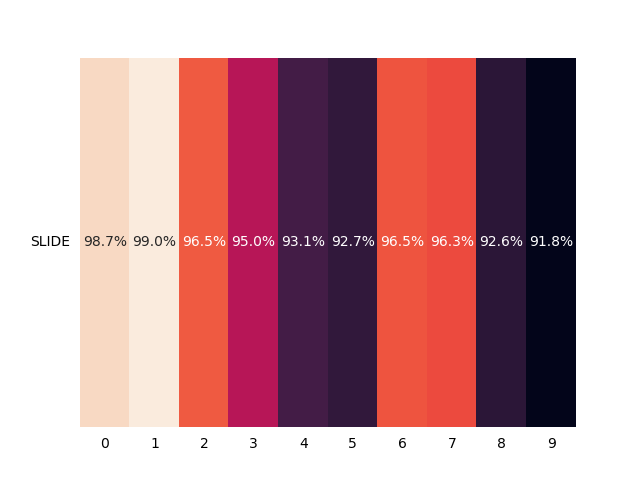}
     \end{subfigure}
        \caption{Class accuracy breakdown for single-attack settings, part 2 (generators $6$ to $9$). Each heatmap corresponds to one generator. The heatmaps present the accuracy for a given class.}
        \label{fig:attacks-single-p2}
\end{figure}

In reference to Table 2 of the main paper, Figures~\ref{fig:attacks-single-p1} and~\ref{fig:attacks-single-p2} provide an analysis of the LeNet5-based classification on images transformed by a system trained in single-attack settings. The results are reported for all $9$ attacks considered in the paper. The heatmaps show that in single-attack settings, the system is able to recover features relevant for classification for most attacked images. Two attacks, the additive uniform noise attack (AUN) and the additive Gaussian noise attack (AGN) stand out as particularly hard in terms of recovering the original labels - which confirms the visual inspection of the samples generated by these attacks (discussed in the main paper). Interestingly, we still see some specialization. For instance, the system trained on AGN allows the LeNet5 classifier to achieve a perfect classification accuracy post-defense on images labeled as ones.

\clearpage

%\clearpage

\subsection{Generator analysis for multiple attacks - trained jointly}\label{sec:multi}
\begin{figure}[ht]
     \centering
     \begin{subfigure}[b]{0.49\textwidth}
         \centering
         \includegraphics[width=\textwidth]{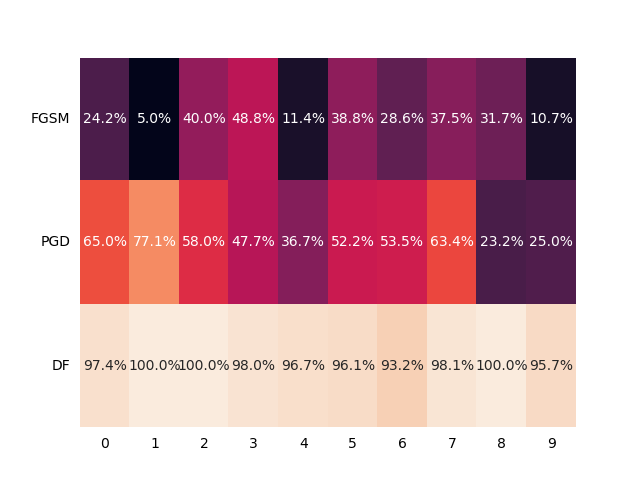}
     \end{subfigure}
     \hfill
     \begin{subfigure}[b]{0.49\textwidth}
         \centering
         \includegraphics[width=\textwidth]{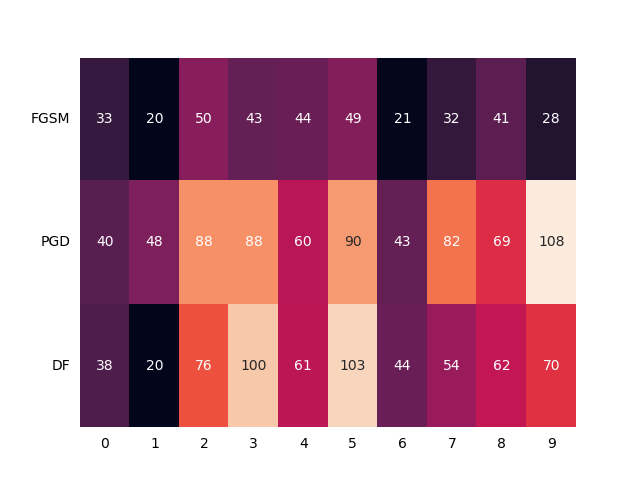}
     \end{subfigure}
     \hfill
     \begin{subfigure}[b]{0.49\textwidth}
         \centering
         \includegraphics[width=\textwidth]{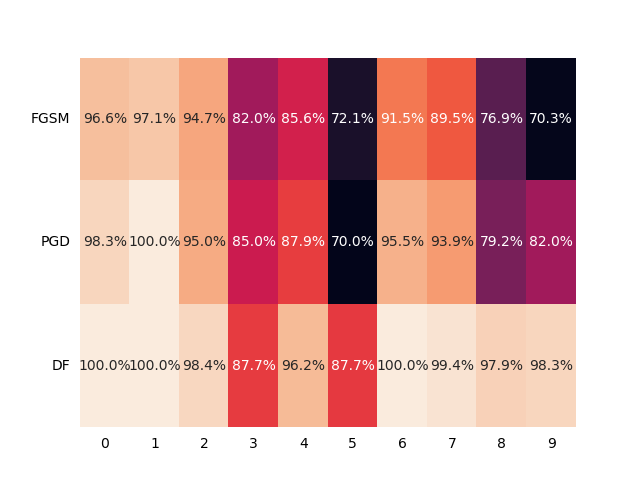}
     \end{subfigure}
     \hfill
     \begin{subfigure}[b]{0.49\textwidth}
         \centering
         \includegraphics[width=\textwidth]{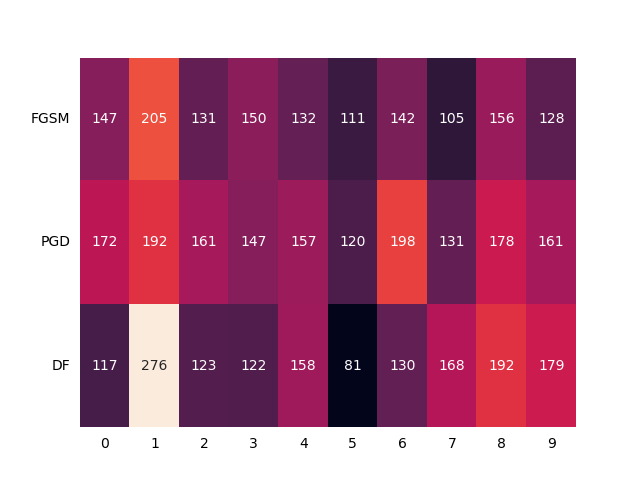}
     \end{subfigure}
     \hfill
     \begin{subfigure}[b]{0.49\textwidth}
         \centering
         \includegraphics[width=\textwidth]{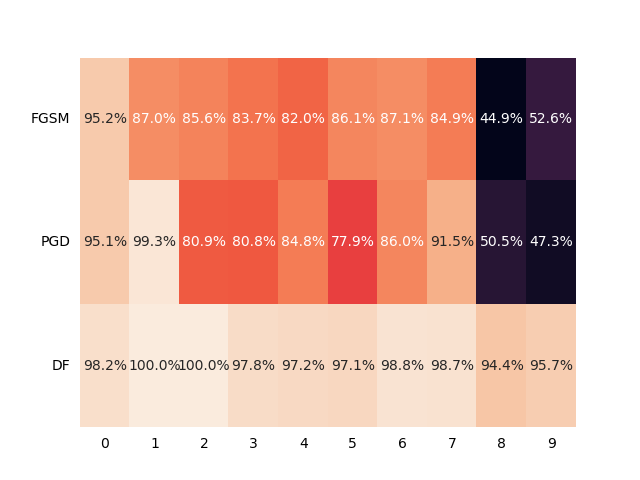}
     \end{subfigure}
     \hfill
     \begin{subfigure}[b]{0.49\textwidth}
         \centering
         \includegraphics[width=\textwidth]{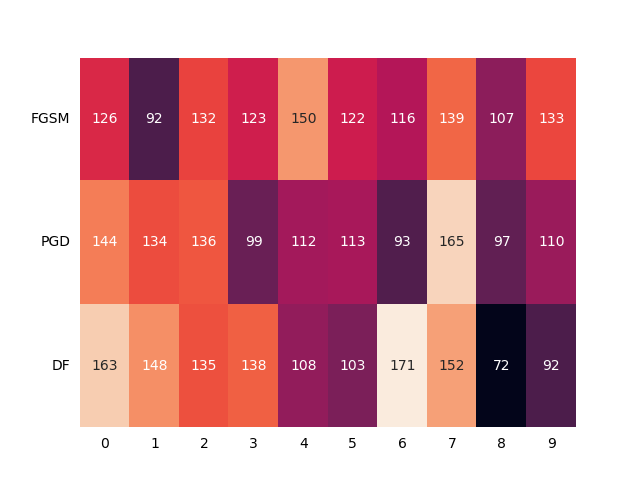}
     \end{subfigure}
        \caption{Class/attack accuracy breakdown for the $3$ attack/generator setting. Each row corresponds to one generator. The heatmaps present the accuracy for a given class/attack (left) and the number of samples on which a given generator won (right).}
        \label{fig:attacks-3}
\end{figure}

\begin{figure}[ht]
     \centering
     \begin{subfigure}[b]{0.49\textwidth}
         \centering
         \includegraphics[width=\textwidth]{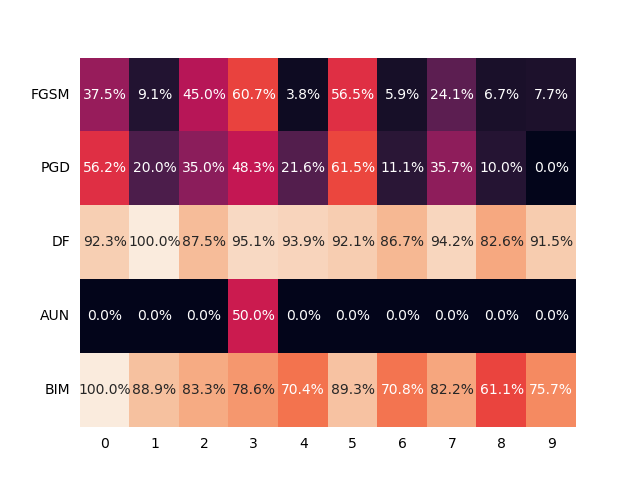}
     \end{subfigure}
     \hfill
     \begin{subfigure}[b]{0.49\textwidth}
         \centering
         \includegraphics[width=\textwidth]{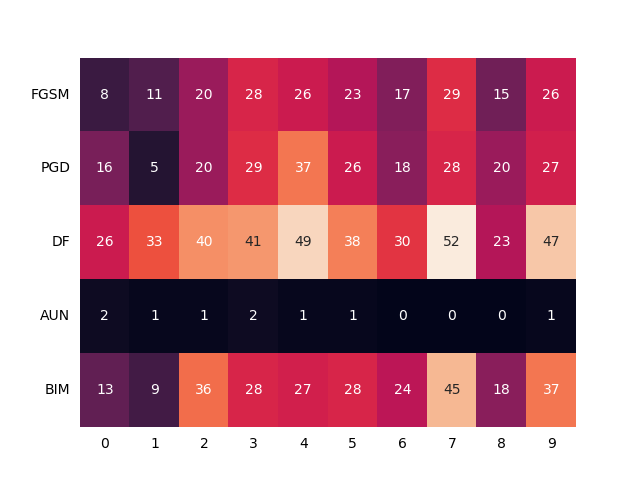}
     \end{subfigure}
     \hfill
     \begin{subfigure}[b]{0.49\textwidth}
         \centering
         \includegraphics[width=\textwidth]{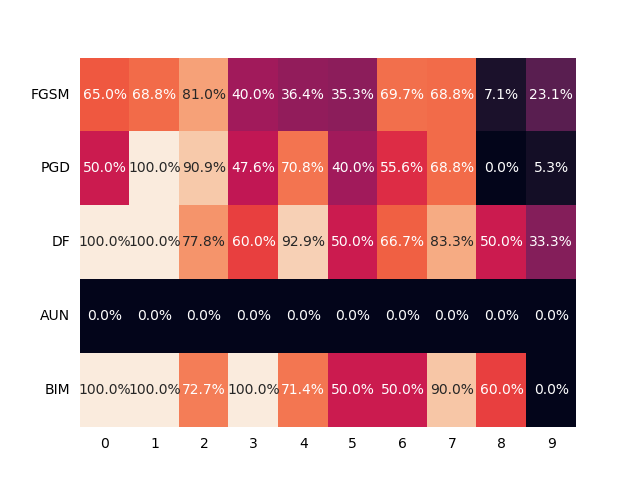}
     \end{subfigure}
     \hfill
     \begin{subfigure}[b]{0.49\textwidth}
         \centering
         \includegraphics[width=\textwidth]{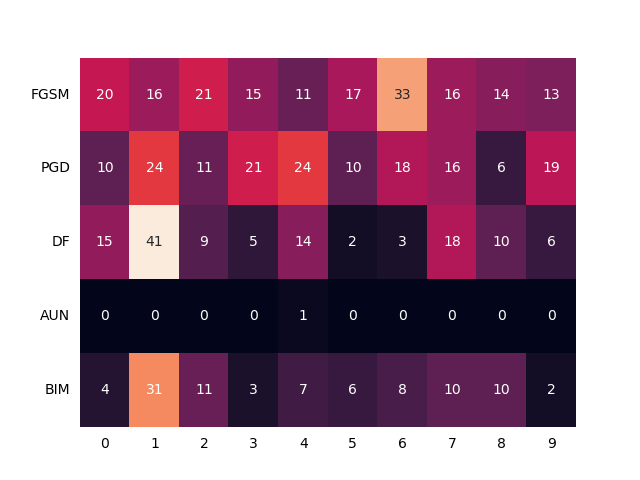}
     \end{subfigure}
     \hfill
     \begin{subfigure}[b]{0.49\textwidth}
         \centering
         \includegraphics[width=\textwidth]{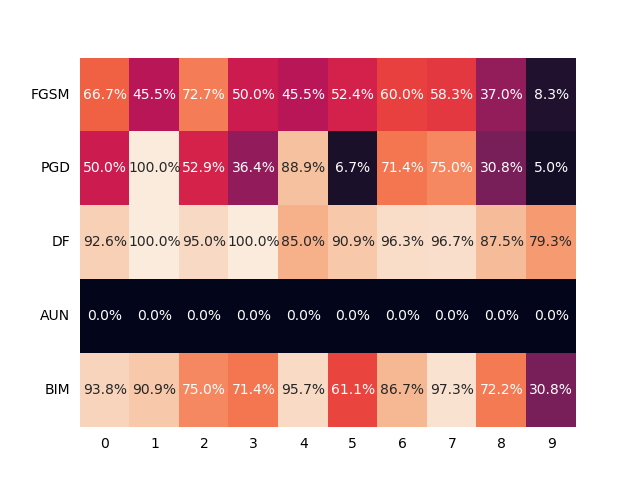}
     \end{subfigure}
     \hfill
     \begin{subfigure}[b]{0.49\textwidth}
         \centering
         \includegraphics[width=\textwidth]{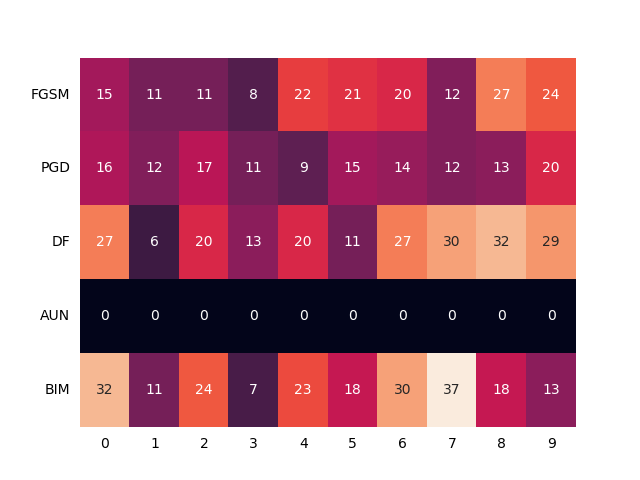}
     \end{subfigure}
        \caption{Class/attack accuracy breakdown for the $5$ attack/generator setting, part 1 (generators $1$ to $3$). Each row corresponds to one generator. The heatmaps present the accuracy for a given class/attack (left) and the number of samples on which a given generator won (right).}
        \label{fig:attacks-5-p1}
\end{figure}

\begin{figure}[ht]
     \centering
     \begin{subfigure}[b]{0.49\textwidth}
         \centering
         \includegraphics[width=\textwidth]{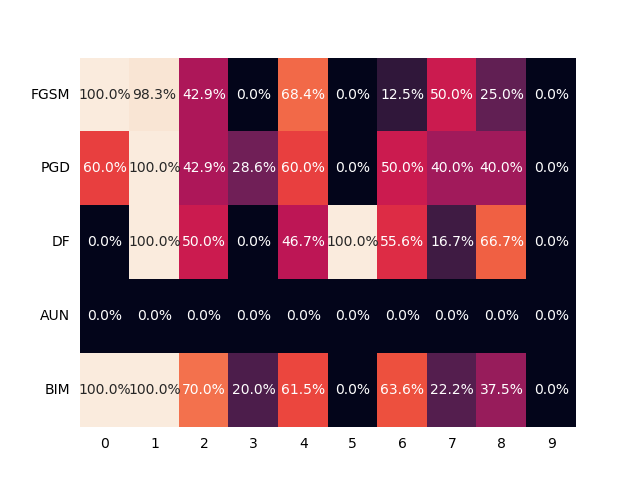}
     \end{subfigure}
     \hfill
     \begin{subfigure}[b]{0.49\textwidth}
         \centering
         \includegraphics[width=\textwidth]{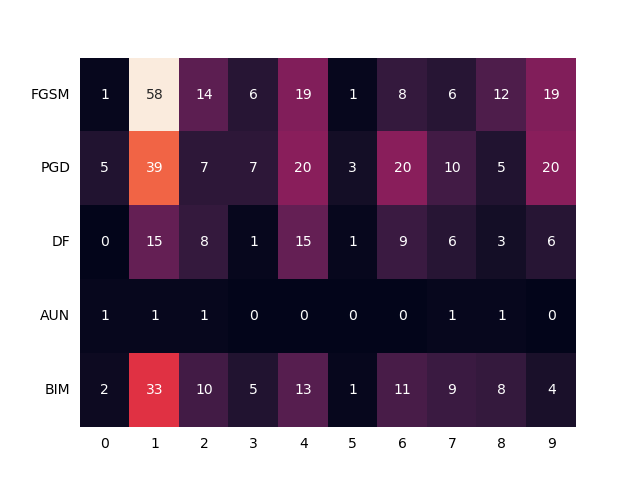}
     \end{subfigure}
     \hfill
     \begin{subfigure}[b]{0.49\textwidth}
         \centering
         \includegraphics[width=\textwidth]{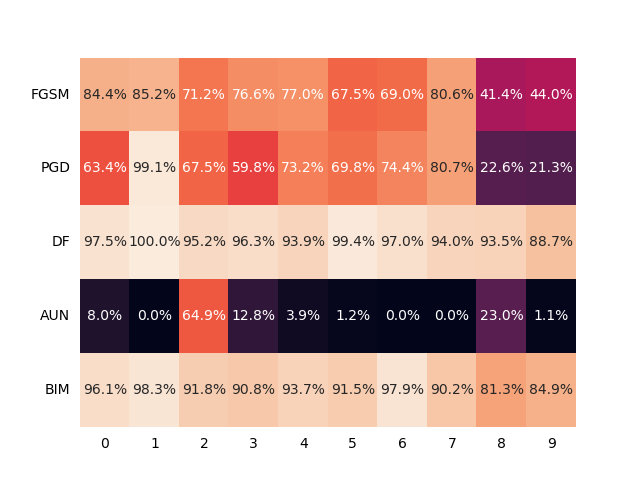}
     \end{subfigure}
     \hfill
     \begin{subfigure}[b]{0.49\textwidth}
         \centering
         \includegraphics[width=\textwidth]{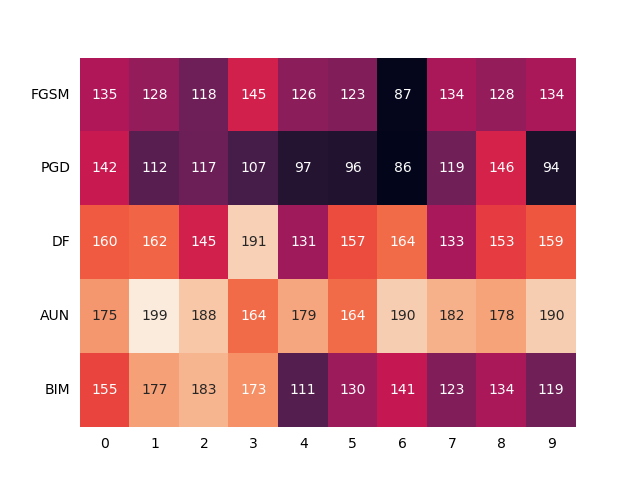}
     \end{subfigure}
        \caption{Class/attack accuracy breakdown for the $5$ attack/generator setting, part 2 (generators $4$ and $5$). Each row corresponds to one generator. The heatmaps present the accuracy for a given class/attack (left) and the number of samples on which a given generator won (right).}
        \label{fig:attacks-5-p2}
\end{figure}

\begin{figure}[ht]
     \centering
     \begin{subfigure}[b]{0.49\textwidth}
         \centering
         \includegraphics[width=\textwidth]{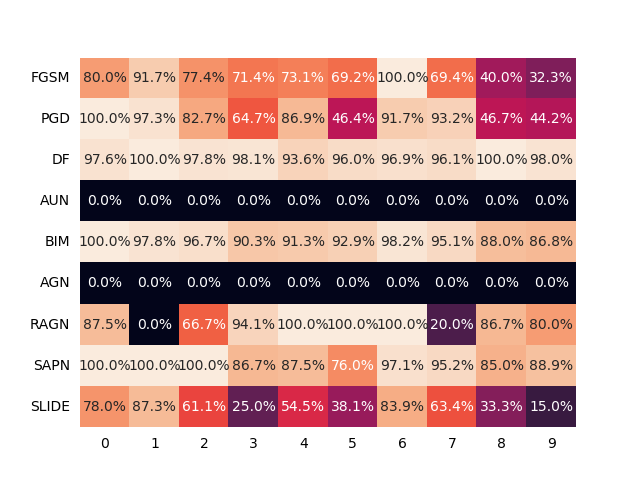}
     \end{subfigure}
     \hfill
     \begin{subfigure}[b]{0.49\textwidth}
         \centering
         \includegraphics[width=\textwidth]{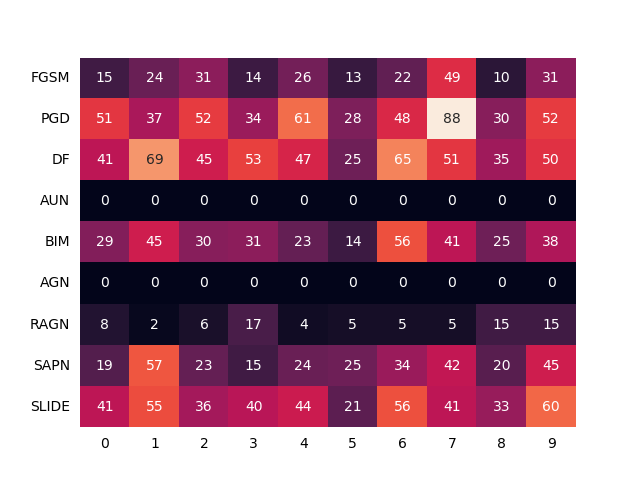}
     \end{subfigure}
     \hfill
     \begin{subfigure}[b]{0.49\textwidth}
         \centering
         \includegraphics[width=\textwidth]{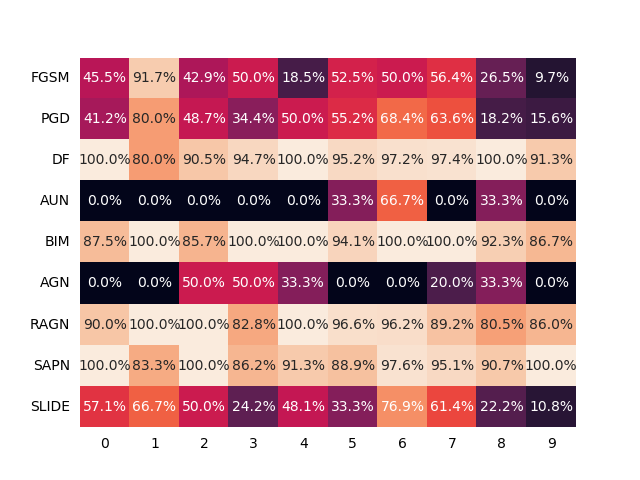}
     \end{subfigure}
     \hfill
     \begin{subfigure}[b]{0.49\textwidth}
         \centering
         \includegraphics[width=\textwidth]{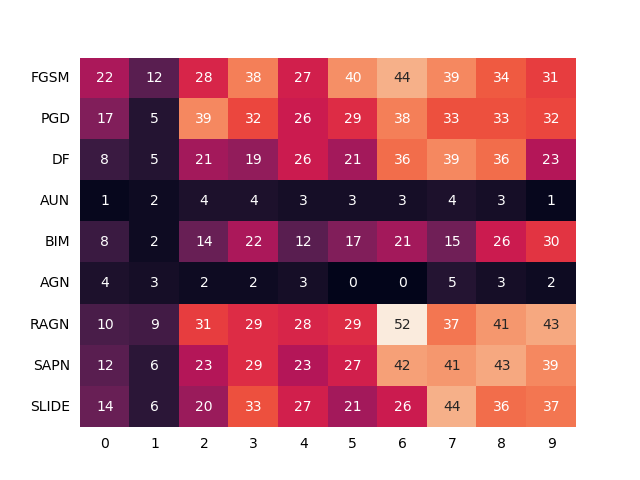}
     \end{subfigure}
     \hfill
     \begin{subfigure}[b]{0.49\textwidth}
         \centering
         \includegraphics[width=\textwidth]{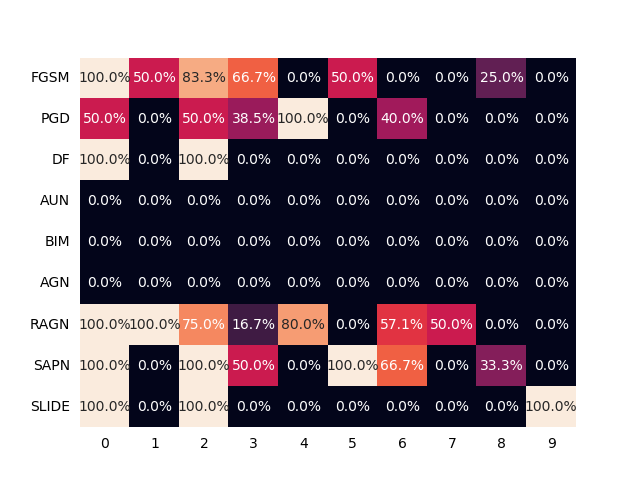}
     \end{subfigure}
     \hfill
     \begin{subfigure}[b]{0.49\textwidth}
         \centering
         \includegraphics[width=\textwidth]{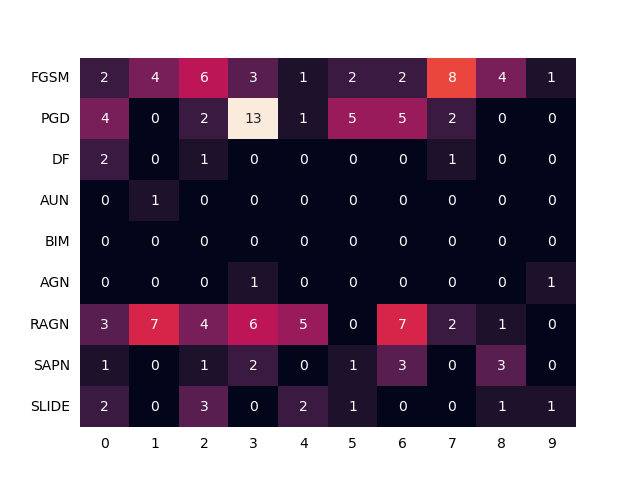}
     \end{subfigure}
        \caption{Class/attack accuracy breakdown for the $9$ attack/generator setting, part 1 (generators $1$ to $3$). Each row corresponds to one generator. The heatmaps present the accuracy for a given class/attack (left) and the number of samples on which a given generator won (right).}
        \label{fig:attacks-9-p1}
\end{figure}

\begin{figure}[ht]
     \centering
     \begin{subfigure}[b]{0.49\textwidth}
         \centering
         \includegraphics[width=\textwidth]{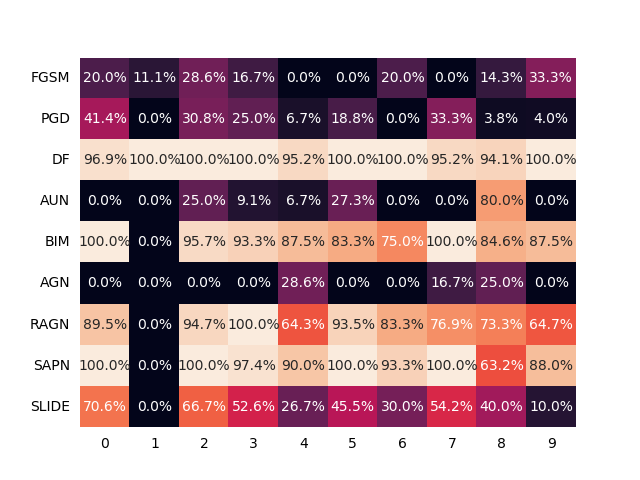}
     \end{subfigure}
     \hfill
     \begin{subfigure}[b]{0.49\textwidth}
         \centering
         \includegraphics[width=\textwidth]{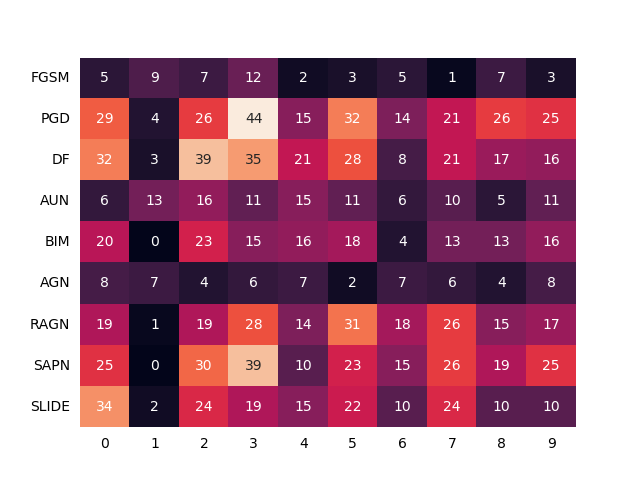}
     \end{subfigure}
     \hfill
     \begin{subfigure}[b]{0.49\textwidth}
         \centering
         \includegraphics[width=\textwidth]{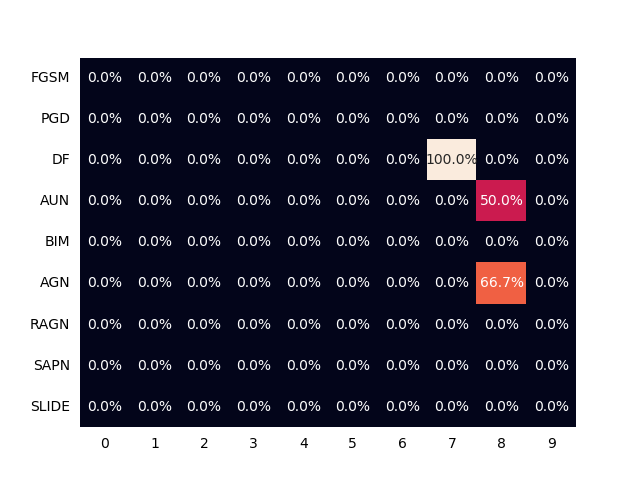}
     \end{subfigure}
     \hfill
     \begin{subfigure}[b]{0.49\textwidth}
         \centering
         \includegraphics[width=\textwidth]{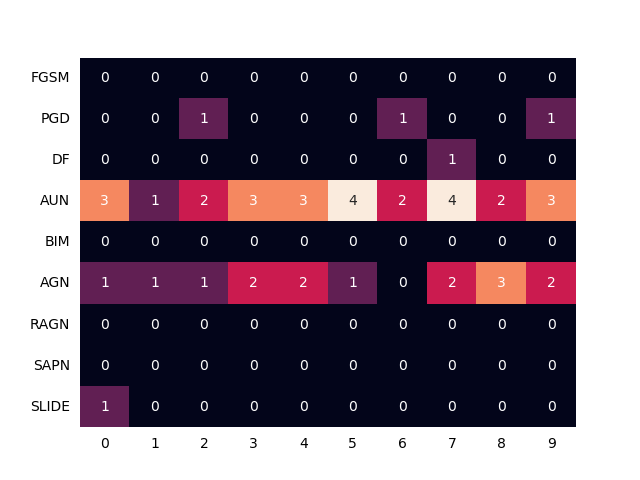}
     \end{subfigure}
     \hfill
     \begin{subfigure}[b]{0.49\textwidth}
         \centering
         \includegraphics[width=\textwidth]{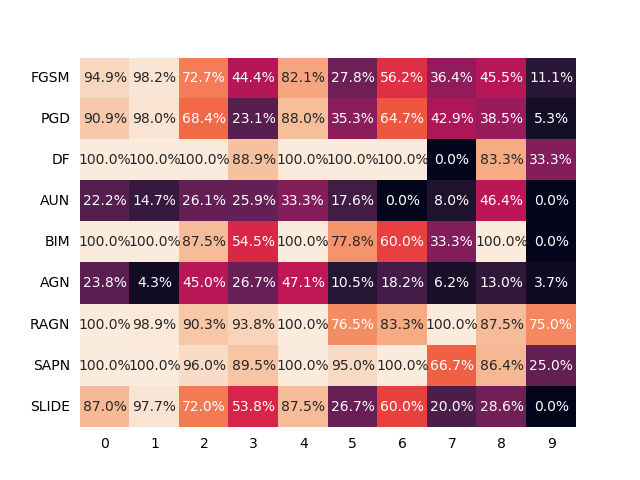}
     \end{subfigure}
     \hfill
     \begin{subfigure}[b]{0.49\textwidth}
         \centering
         \includegraphics[width=\textwidth]{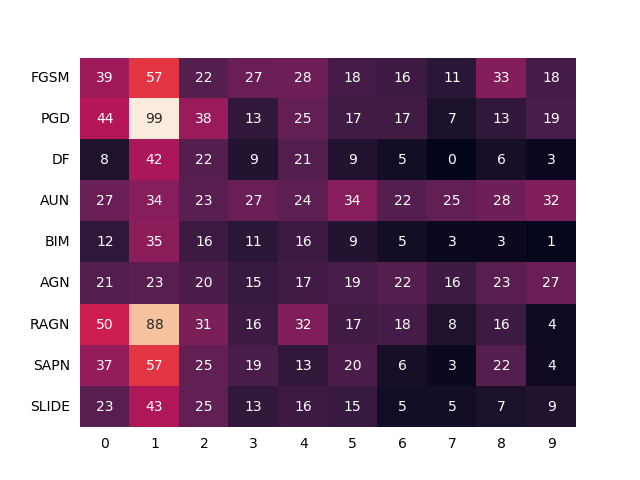}
     \end{subfigure}
        \caption{Class/attack accuracy breakdown for the $9$ attack/generator setting, part 2 (generators $4$ to $6$). Each row corresponds to one generator. The heatmaps present the accuracy for a given class/attack (left) and the number of samples on which a given generator won (right).}
        \label{fig:attacks-9-p2}
\end{figure}

\begin{figure}[ht]
     \centering
     \begin{subfigure}[b]{0.49\textwidth}
         \centering
         \includegraphics[width=\textwidth]{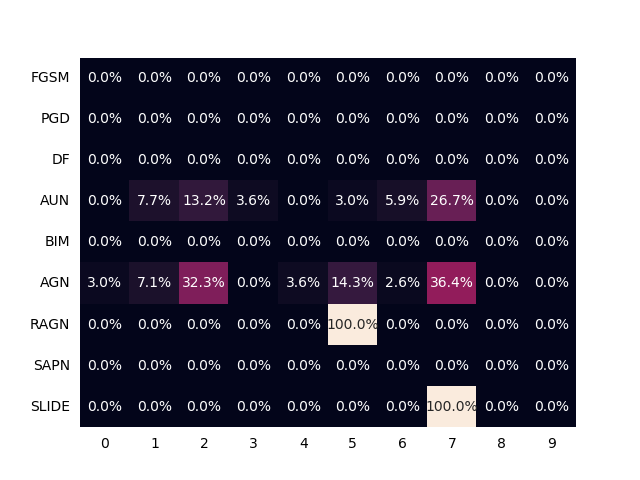}
     \end{subfigure}
     \hfill
     \begin{subfigure}[b]{0.49\textwidth}
         \centering
         \includegraphics[width=\textwidth]{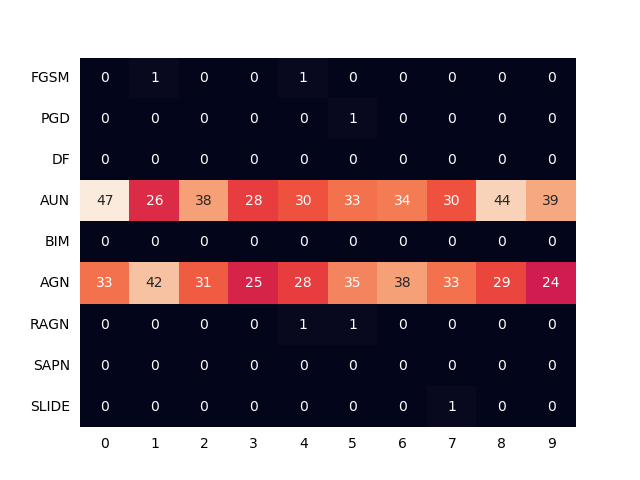}
     \end{subfigure}
     \hfill
     \begin{subfigure}[b]{0.49\textwidth}
         \centering
         \includegraphics[width=\textwidth]{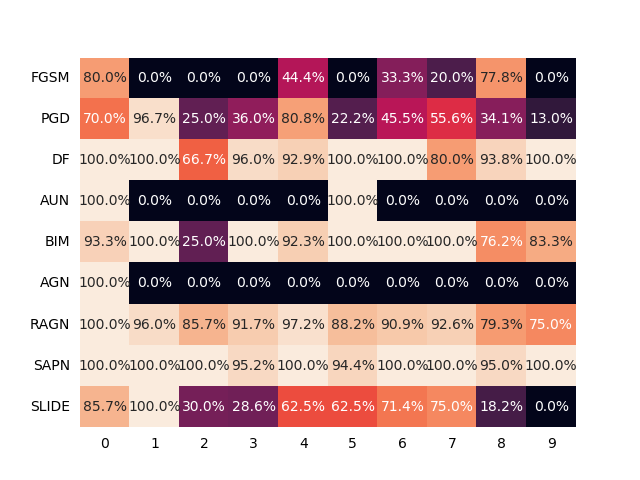}
     \end{subfigure}
     \hfill
     \begin{subfigure}[b]{0.49\textwidth}
         \centering
         \includegraphics[width=\textwidth]{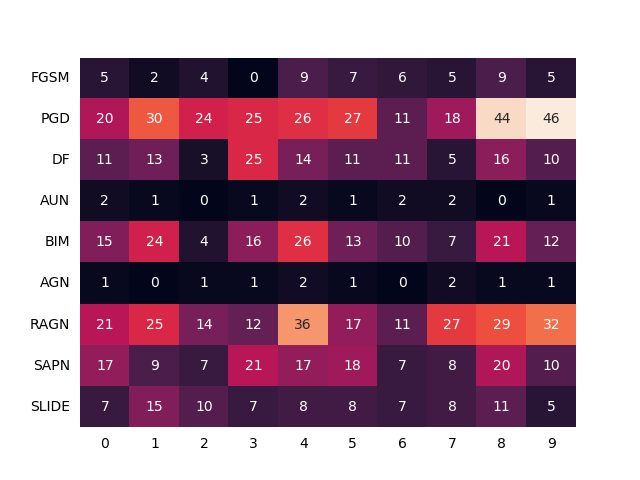}
     \end{subfigure}
     \hfill
     \begin{subfigure}[b]{0.49\textwidth}
         \centering
         \includegraphics[width=\textwidth]{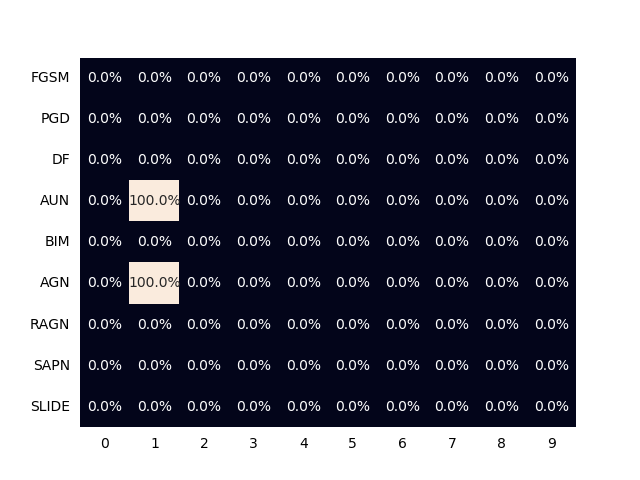}
     \end{subfigure}
     \hfill
     \begin{subfigure}[b]{0.49\textwidth}
         \centering
         \includegraphics[width=\textwidth]{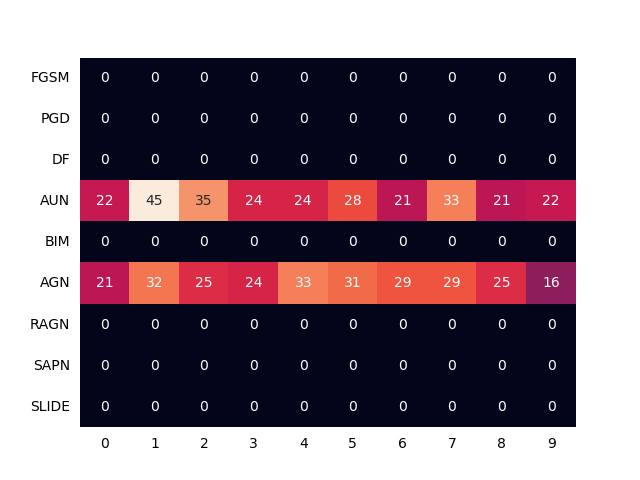}
     \end{subfigure}
        \caption{Class/attack accuracy breakdown for the $9$ attack/generator setting, part 3 (generators $7$ to $9$). Each row corresponds to one generator. The heatmaps present the accuracy for a given class/attack (left) and the number of samples on which a given generator won (right).}
        \label{fig:attacks-9-p3}
\end{figure}

Similarly to Section~\ref{sec:single}, we report the results for multi-attack/multi-generator settings discussed in the main paper. The setup with $3$ attacks/generators is presented in Figure~\ref{fig:attacks-3}. Analogous results for the setup with $5$ attacks/generators are presented in Figures~\ref{fig:attacks-5-p1} and~\ref{fig:attacks-5-p2}. The largest setup with $9$ attacks/generators is reported in Figures~\ref{fig:attacks-9-p1},~\ref{fig:attacks-9-p2} and~\ref{fig:attacks-9-p3}. The figures for the $9$ generator setup supplement the examples presented in Figure~4 in the main paper. Due to space limits we could not accommodate more heatmap examples in the body of the main paper.

The results confirm the earlier remarks that AUN and AGN, due to their high level of visual distortion of the images, are harder to defend against. We also see that there are roughly three types of generators: the ones covering a wide variety of attacks (\emph{generalists}), generators covering only a subset of attacks/classes and recovering the label data relatively well (\emph{specialists}), and generators covering a subset of attacks/classes without substantial success in recovering label data (\emph{marginalists}). This is particularly visible for the $9$-attack setting (Figures~\ref{fig:attacks-9-p1},~\ref{fig:attacks-9-p2} and~\ref{fig:attacks-9-p3}).

\clearpage

%\clearpage

%\clearpage

%\clearpage

%\clearpage

\clearpage

\subsection{Generator analysis in repeated experiments trained jointly with \emph{faster initialization}}
The results obtained using the faster initialization method do not materially differ (both in terms of post-defense accuracy and in terms of the qualitative assessment of the heatmaps) from the results of standard joint training without faster initialization shown in Section \ref{sec:multi}. Due to the comparable outcomes of the two approaches, we present further results for the faster initialization method as it affords us the possibility to use less computational resources and speed up training.

The heatmaps for $3$ sample sets of repeated multi-attack experiments with faster initialization are shown below in Figures \ref{fig:repeated-first}-\ref{fig:repeated-last} (the experiment numbers correspond to those presented in Table~\ref{tab:repeated}). Each set of experiments encompasses $3$ settings: $3$, $5$ and $9$ attacks/generators. The main observation is that the specialization of generators becomes more visible in settings with more attacks/generators. For the $3$-attack setting, the generators are of the \emph{generalist} type, with potential variation in the effectiveness of the defense on different attacks. For the $5$-attack setting, specialization in the additive uniform noise attack (AUN) becomes more visible, occasionally producing \emph{specialists}. Finally, for the $9$-attack setting, there is further specialization as some of the generators become \emph{specialists} or \emph{marginalists}.
\clearpage

\subsubsection{Experiment 3, trained jointly - faster initialization}

\begin{figure}[ht]
     \centering
     \begin{subfigure}[b]{0.49\textwidth}
         \centering
         \includegraphics[width=\textwidth]{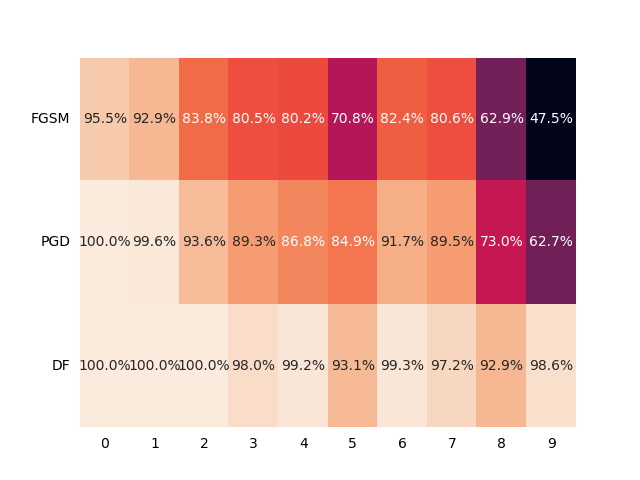}
     \end{subfigure}
     \hfill
     \begin{subfigure}[b]{0.49\textwidth}
         \centering
         \includegraphics[width=\textwidth]{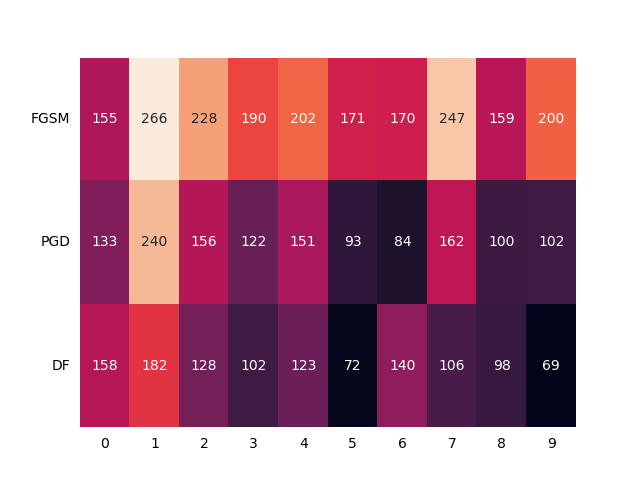}
     \end{subfigure}
     \hfill
     \begin{subfigure}[b]{0.49\textwidth}
         \centering
         \includegraphics[width=\textwidth]{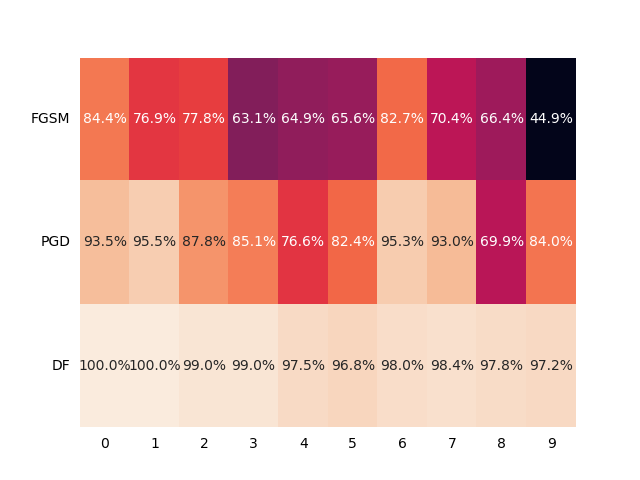}
     \end{subfigure}
     \hfill
     \begin{subfigure}[b]{0.49\textwidth}
         \centering
         \includegraphics[width=\textwidth]{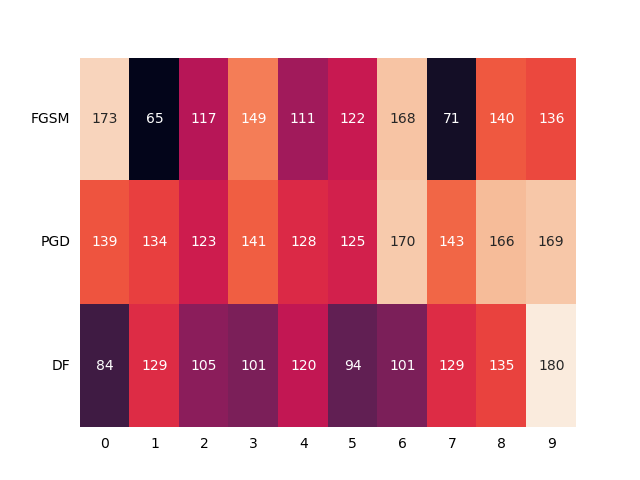}
     \end{subfigure}
     \hfill
     \begin{subfigure}[b]{0.49\textwidth}
         \centering
         \includegraphics[width=\textwidth]{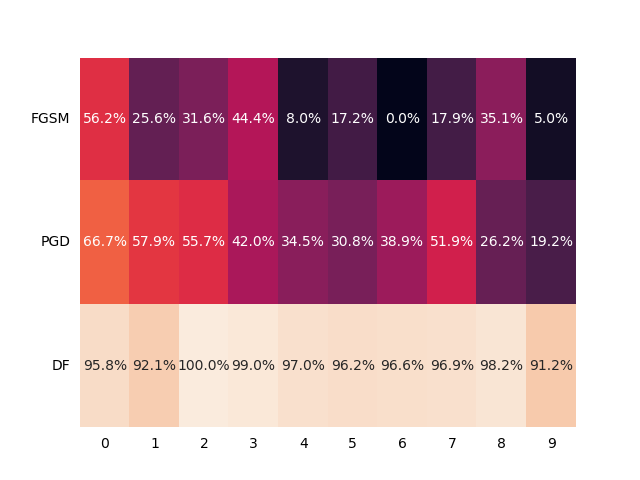}
     \end{subfigure}
     \hfill
     \begin{subfigure}[b]{0.49\textwidth}
         \centering
         \includegraphics[width=\textwidth]{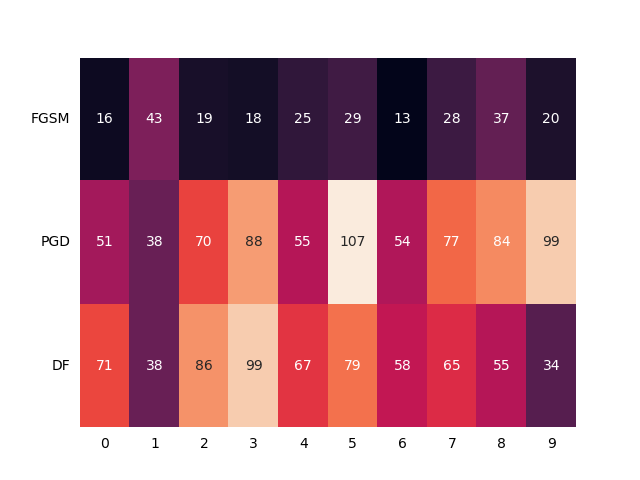}
     \end{subfigure}
        \caption{Experiment 3 (trained jointly - faster initialization). Class/attack accuracy breakdown for the $3$ attack/generator setting. Each row corresponds to one generator. The heatmaps present the accuracy for a given class/attack (left) and the number of samples on which a given generator won (right).}
        \label{fig:repeated-first}
\end{figure}
\clearpage

\begin{figure}[ht]
     \centering
     \begin{subfigure}[b]{0.49\textwidth}
         \centering
         \includegraphics[width=\textwidth]{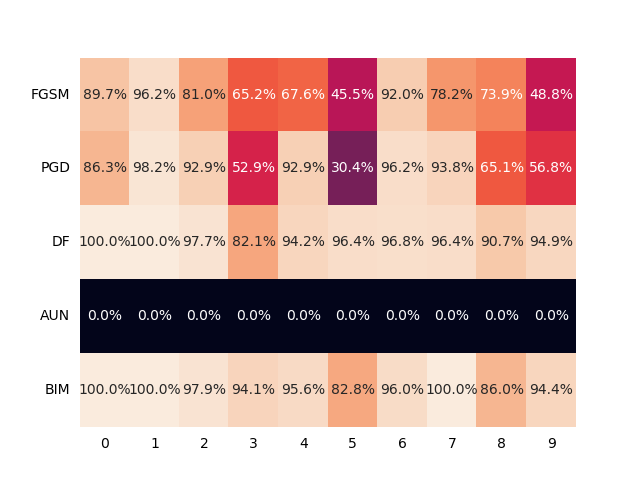}
     \end{subfigure}
     \hfill
     \begin{subfigure}[b]{0.49\textwidth}
         \centering
         \includegraphics[width=\textwidth]{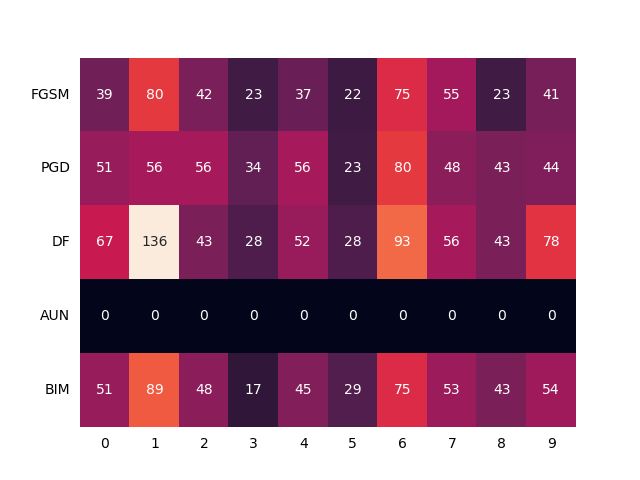}
     \end{subfigure}
     \hfill
     \begin{subfigure}[b]{0.49\textwidth}
         \centering
         \includegraphics[width=\textwidth]{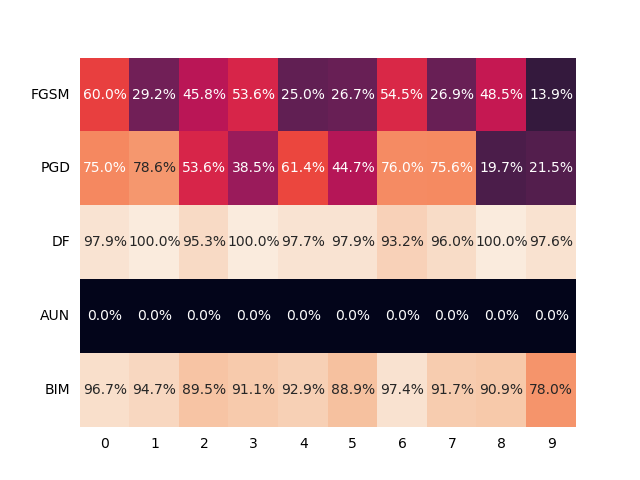}
     \end{subfigure}
     \hfill
     \begin{subfigure}[b]{0.49\textwidth}
         \centering
         \includegraphics[width=\textwidth]{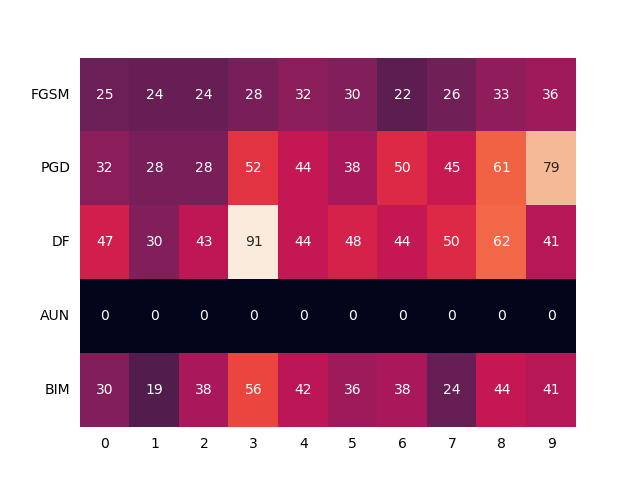}
     \end{subfigure}
     \hfill
     \begin{subfigure}[b]{0.49\textwidth}
         \centering
         \includegraphics[width=\textwidth]{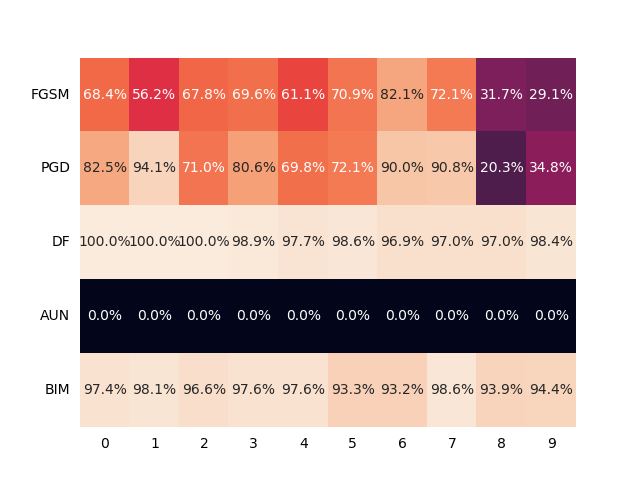}
     \end{subfigure}
     \hfill
     \begin{subfigure}[b]{0.49\textwidth}
         \centering
         \includegraphics[width=\textwidth]{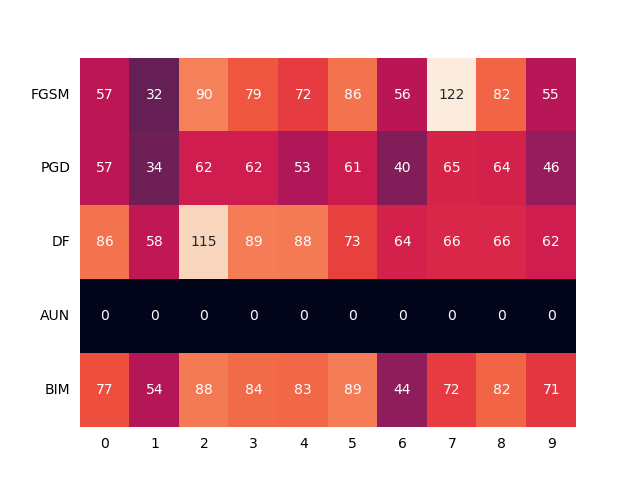}
     \end{subfigure}
        \caption{Experiment 3 (trained jointly - faster initialization). Class/attack accuracy breakdown for the $5$ attack/generator setting, part 1 (generators $1$ to $3$). Each row corresponds to one generator. The heatmaps present the accuracy for a given class/attack (left) and the number of samples on which a given generator won (right).}
\end{figure}
\clearpage

\begin{figure}[ht]
     \centering
     \begin{subfigure}[b]{0.49\textwidth}
         \centering
         \includegraphics[width=\textwidth]{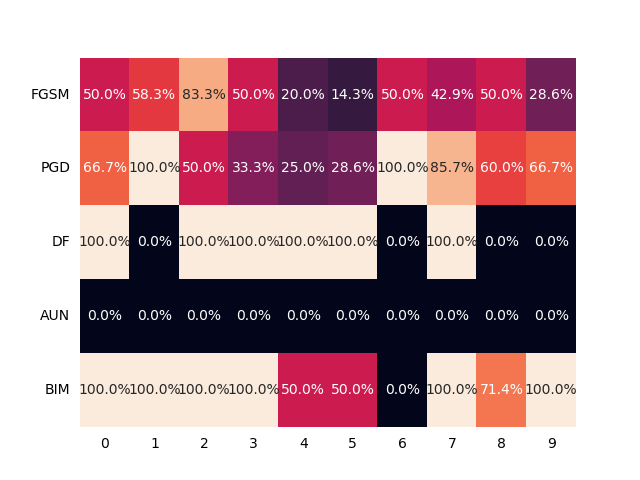}
     \end{subfigure}
     \hfill
     \begin{subfigure}[b]{0.49\textwidth}
         \centering
         \includegraphics[width=\textwidth]{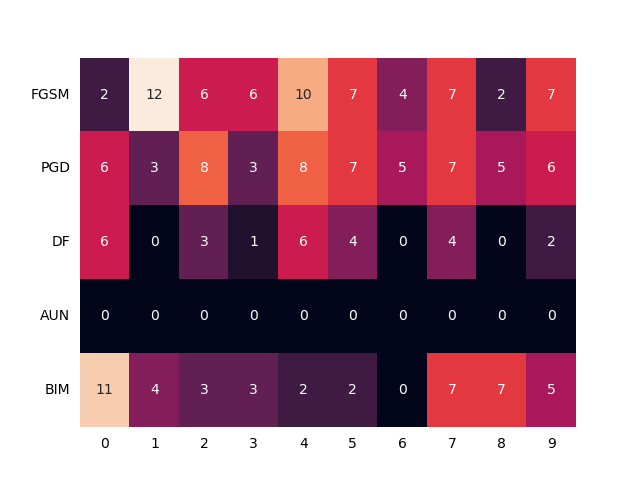}
     \end{subfigure}
     \hfill
     \begin{subfigure}[b]{0.49\textwidth}
         \centering
         \includegraphics[width=\textwidth]{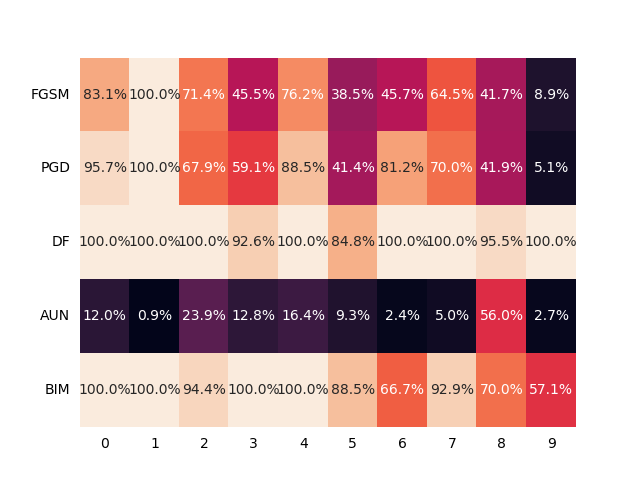}
     \end{subfigure}
     \hfill
     \begin{subfigure}[b]{0.49\textwidth}
         \centering
         \includegraphics[width=\textwidth]{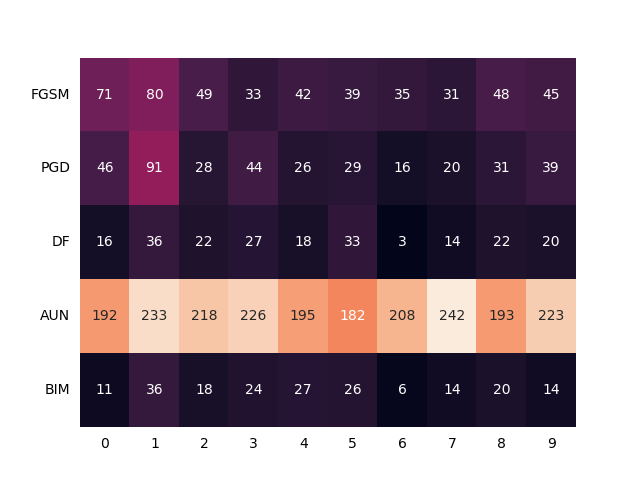}
     \end{subfigure}
        \caption{Experiment 3 (trained jointly - faster initialization). Class/attack accuracy breakdown for the $5$ attack/generator setting, part 2 (generators $4$ and $5$). Each row corresponds to one generator. The heatmaps present the accuracy for a given class/attack (left) and the number of samples on which a given generator won (right).}
\end{figure}
\clearpage

\begin{figure}[ht]
     \centering
     \begin{subfigure}[b]{0.49\textwidth}
         \centering
         \includegraphics[width=\textwidth]{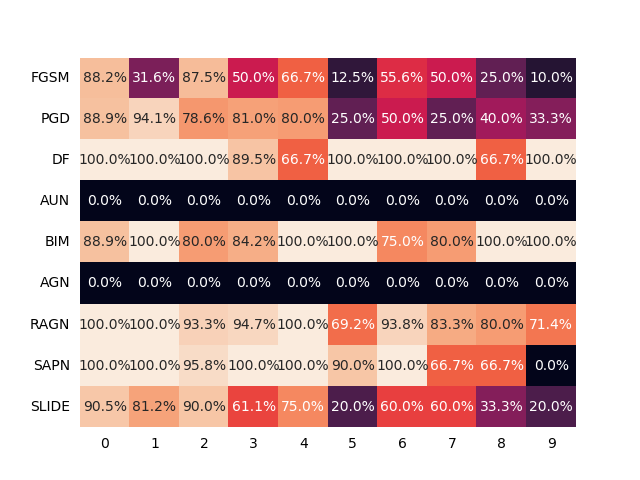}
     \end{subfigure}
     \hfill
     \begin{subfigure}[b]{0.49\textwidth}
         \centering
         \includegraphics[width=\textwidth]{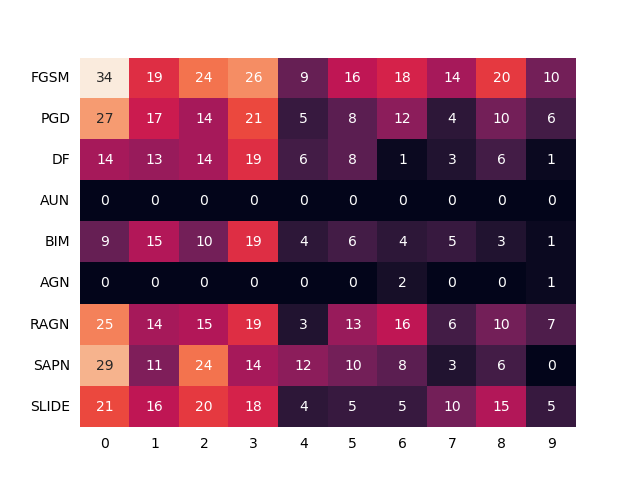}
     \end{subfigure}
     \hfill
     \begin{subfigure}[b]{0.49\textwidth}
         \centering
         \includegraphics[width=\textwidth]{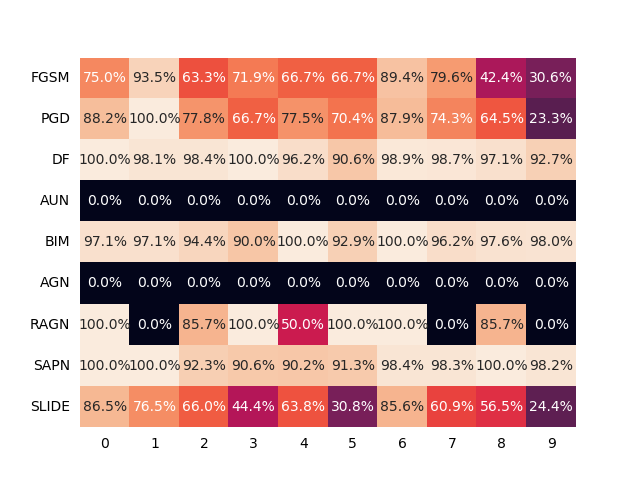}
     \end{subfigure}
     \hfill
     \begin{subfigure}[b]{0.49\textwidth}
         \centering
         \includegraphics[width=\textwidth]{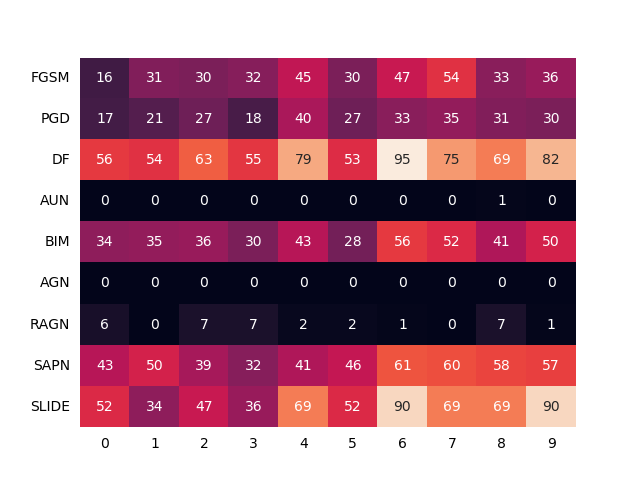}
     \end{subfigure}
     \hfill
     \begin{subfigure}[b]{0.49\textwidth}
         \centering
         \includegraphics[width=\textwidth]{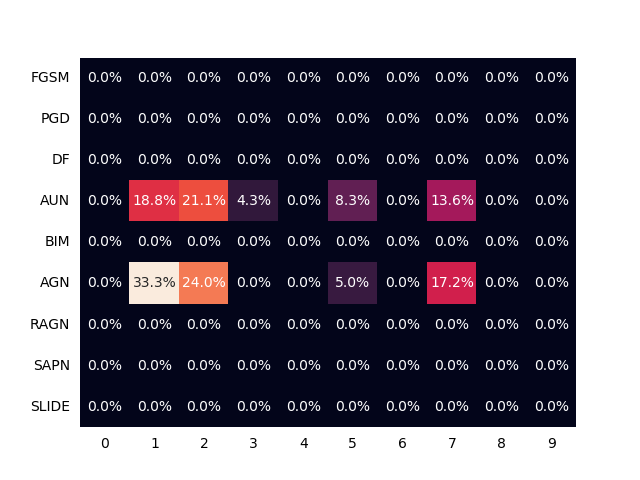}
     \end{subfigure}
     \hfill
     \begin{subfigure}[b]{0.49\textwidth}
         \centering
         \includegraphics[width=\textwidth]{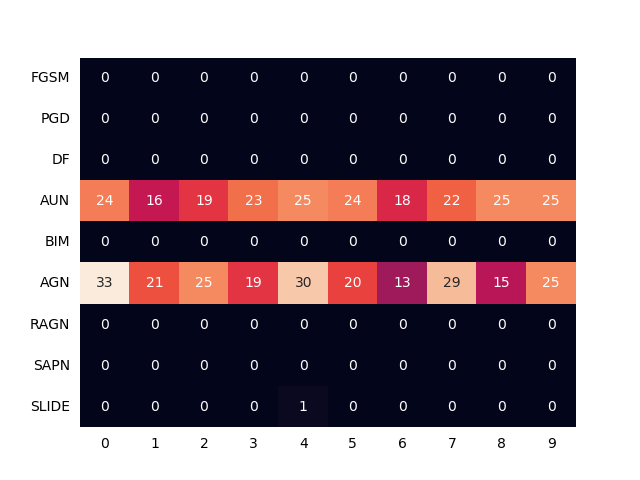}
     \end{subfigure}
        \caption{Experiment 3 (trained jointly - faster initialization). Class/attack accuracy breakdown for the $9$ attack/generator setting, part 1 (generators $1$ to $3$). Each row corresponds to one generator. The heatmaps present the accuracy for a given class/attack (left) and the number of samples on which a given generator won (right).}
\end{figure}
\clearpage

\begin{figure}[ht]
     \centering
     \begin{subfigure}[b]{0.49\textwidth}
         \centering
         \includegraphics[width=\textwidth]{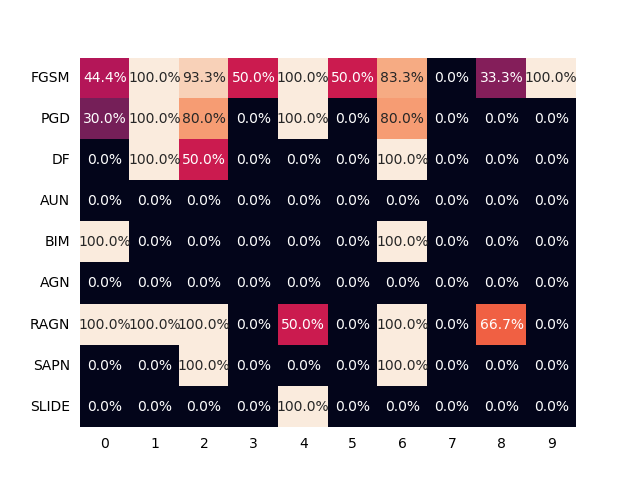}
     \end{subfigure}
     \hfill
     \begin{subfigure}[b]{0.49\textwidth}
         \centering
         \includegraphics[width=\textwidth]{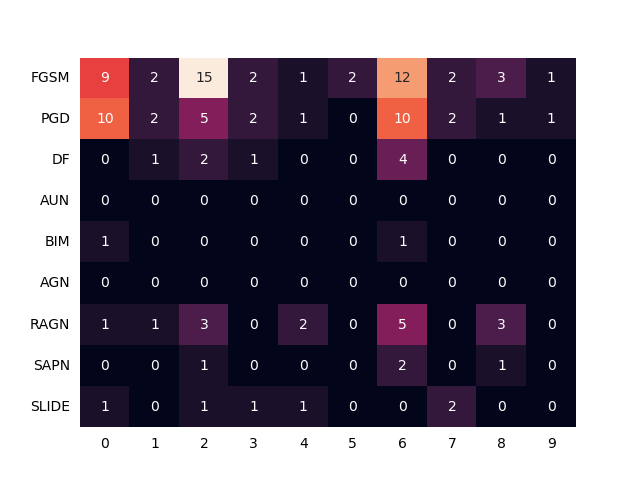}
     \end{subfigure}
     \hfill
     \begin{subfigure}[b]{0.49\textwidth}
         \centering
         \includegraphics[width=\textwidth]{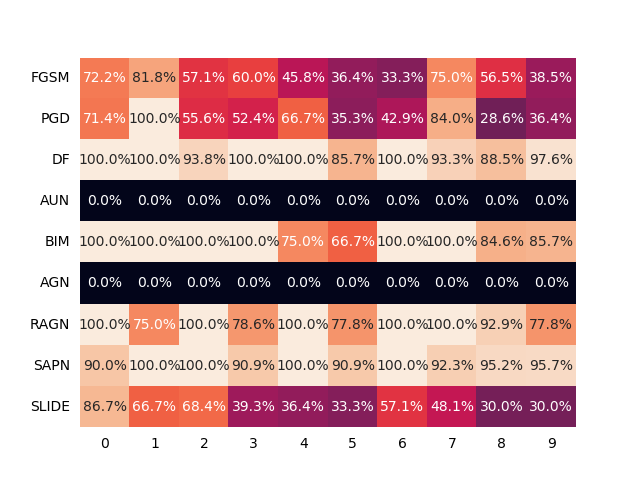}
     \end{subfigure}
     \hfill
     \begin{subfigure}[b]{0.49\textwidth}
         \centering
         \includegraphics[width=\textwidth]{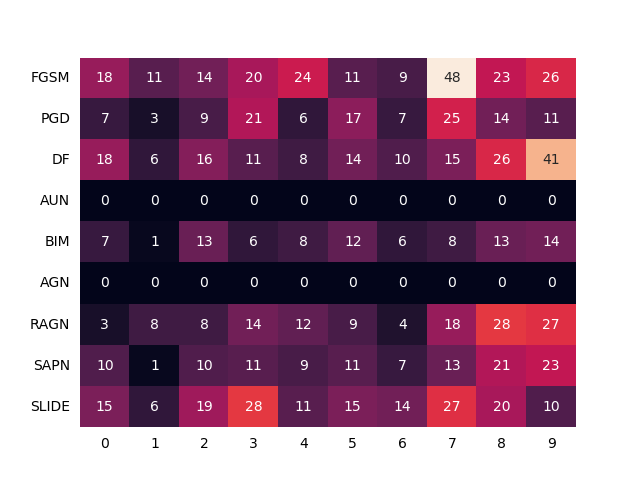}
     \end{subfigure}
     \hfill
     \begin{subfigure}[b]{0.49\textwidth}
         \centering
         \includegraphics[width=\textwidth]{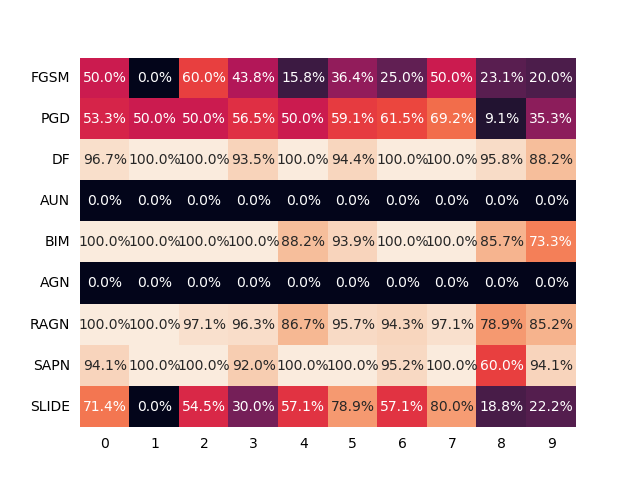}
     \end{subfigure}
     \hfill
     \begin{subfigure}[b]{0.49\textwidth}
         \centering
         \includegraphics[width=\textwidth]{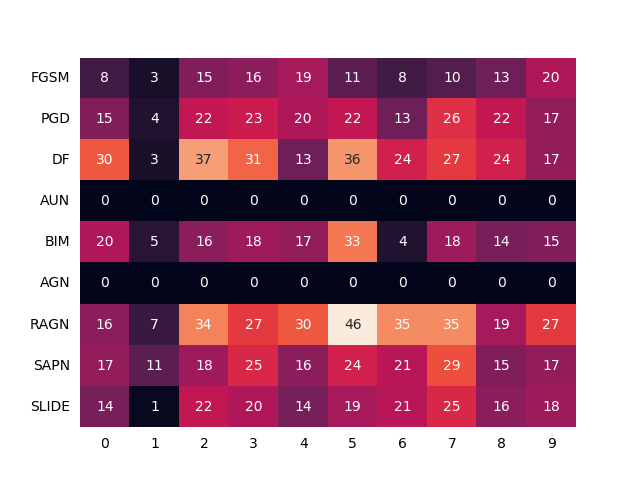}
     \end{subfigure}
        \caption{Experiment 3 (trained jointly - faster initialization). Class/attack accuracy breakdown for the $9$ attack/generator setting, part 2 (generators $4$ to $6$). Each row corresponds to one generator. The heatmaps present the accuracy for a given class/attack (left) and the number of samples on which a given generator won (right).}
\end{figure}
\clearpage

\begin{figure}[ht]
     \centering
     \begin{subfigure}[b]{0.49\textwidth}
         \centering
         \includegraphics[width=\textwidth]{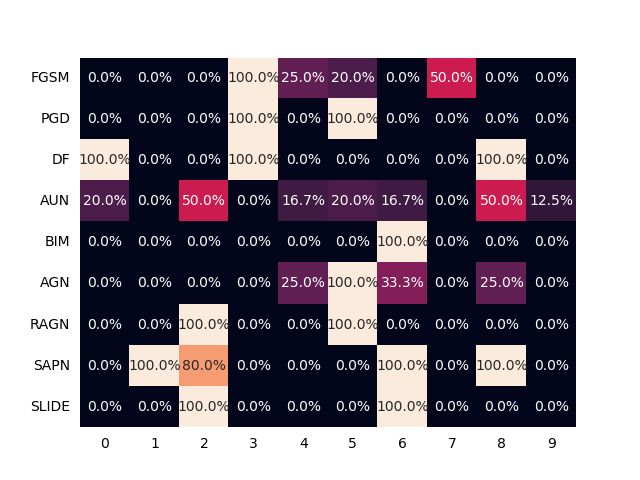}
     \end{subfigure}
     \hfill
     \begin{subfigure}[b]{0.49\textwidth}
         \centering
         \includegraphics[width=\textwidth]{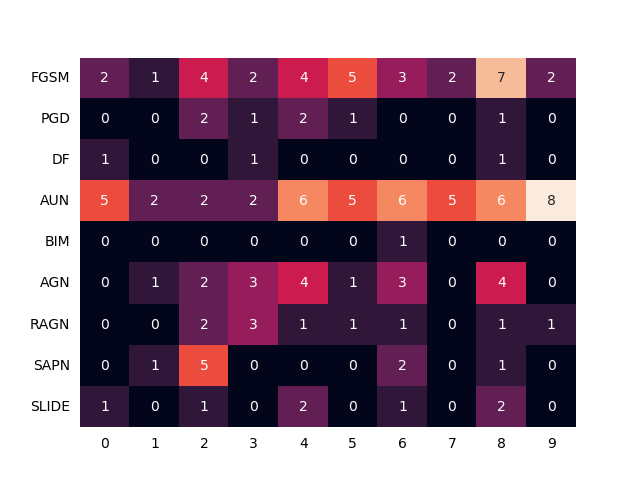}
     \end{subfigure}
     \hfill
     \begin{subfigure}[b]{0.49\textwidth}
         \centering
         \includegraphics[width=\textwidth]{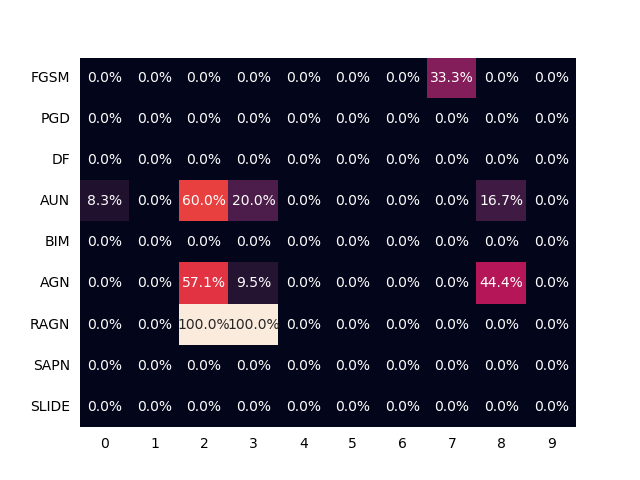}
     \end{subfigure}
     \hfill
     \begin{subfigure}[b]{0.49\textwidth}
         \centering
         \includegraphics[width=\textwidth]{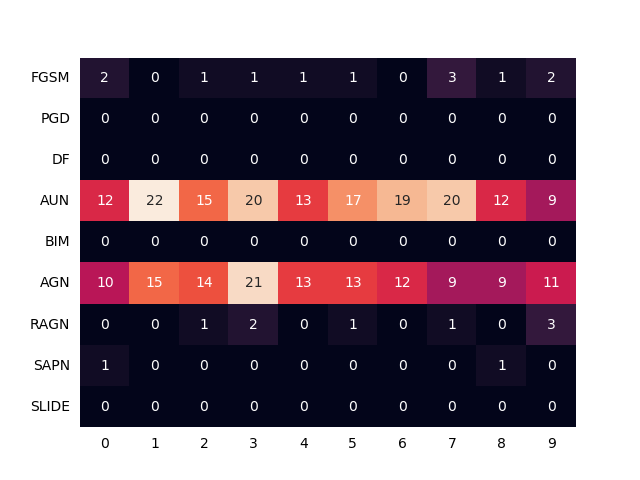}
     \end{subfigure}
     \hfill
     \begin{subfigure}[b]{0.49\textwidth}
         \centering
         \includegraphics[width=\textwidth]{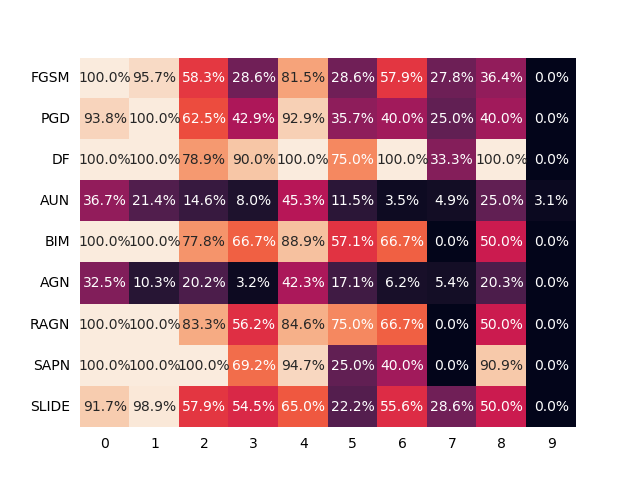}
     \end{subfigure}
     \hfill
     \begin{subfigure}[b]{0.49\textwidth}
         \centering
         \includegraphics[width=\textwidth]{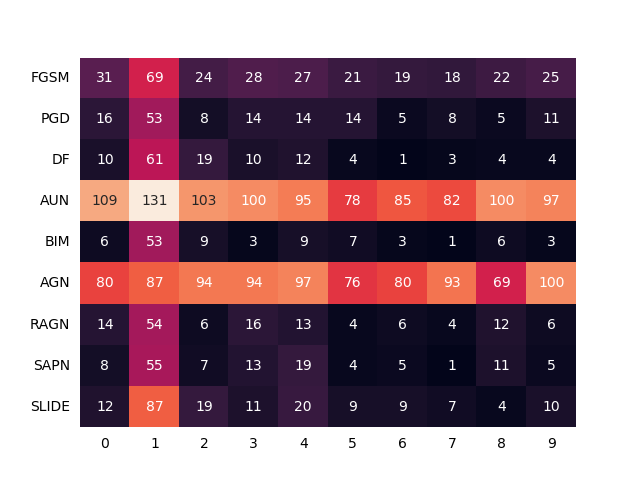}
     \end{subfigure}
        \caption{Experiment 3 (trained jointly - faster initialization). Class/attack accuracy breakdown for the $9$ attack/generator setting, part 3 (generators $7$ to $9$). Each row corresponds to one generator. The heatmaps present the accuracy for a given class/attack (left) and the number of samples on which a given generator won (right).}
\end{figure}
\clearpage

\subsubsection{Experiment 5, trained jointly - faster initialization}

\begin{figure}[ht]
     \centering
     \begin{subfigure}[b]{0.49\textwidth}
         \centering
         \includegraphics[width=\textwidth]{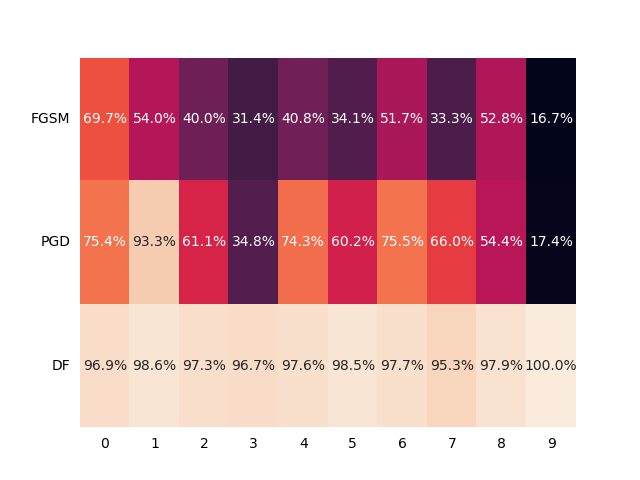}
     \end{subfigure}
     \hfill
     \begin{subfigure}[b]{0.49\textwidth}
         \centering
         \includegraphics[width=\textwidth]{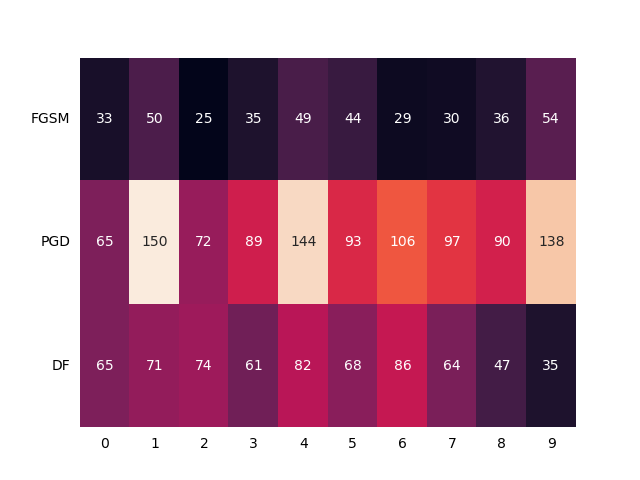}
     \end{subfigure}
     \hfill
     \begin{subfigure}[b]{0.49\textwidth}
         \centering
         \includegraphics[width=\textwidth]{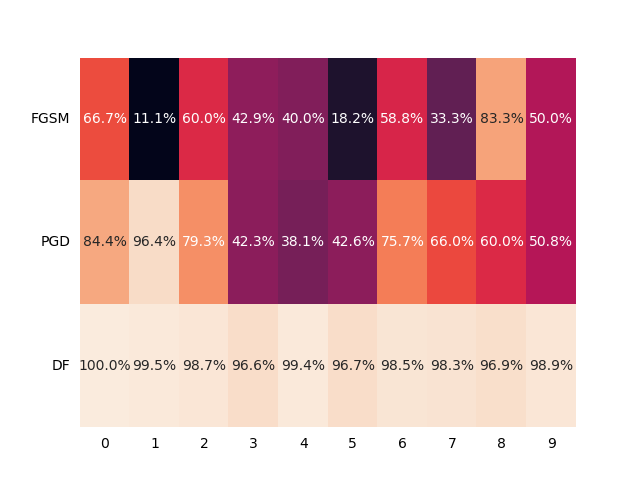}
     \end{subfigure}
     \hfill
     \begin{subfigure}[b]{0.49\textwidth}
         \centering
         \includegraphics[width=\textwidth]{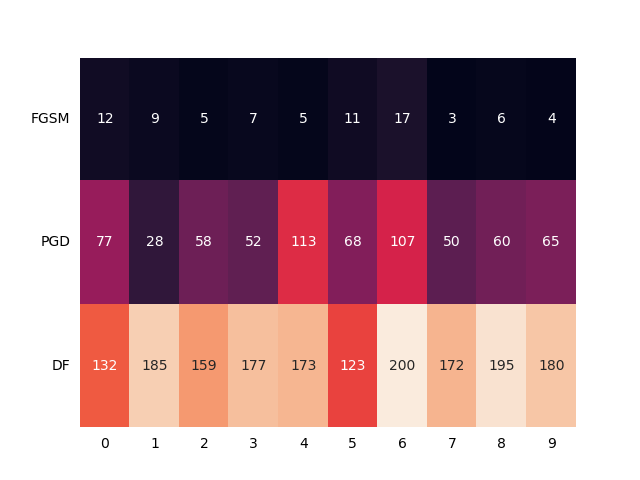}
     \end{subfigure}
     \hfill
     \begin{subfigure}[b]{0.49\textwidth}
         \centering
         \includegraphics[width=\textwidth]{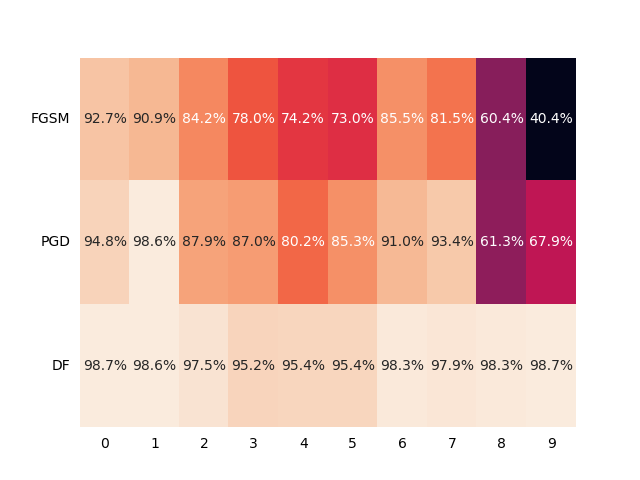}
     \end{subfigure}
     \hfill
     \begin{subfigure}[b]{0.49\textwidth}
         \centering
         \includegraphics[width=\textwidth]{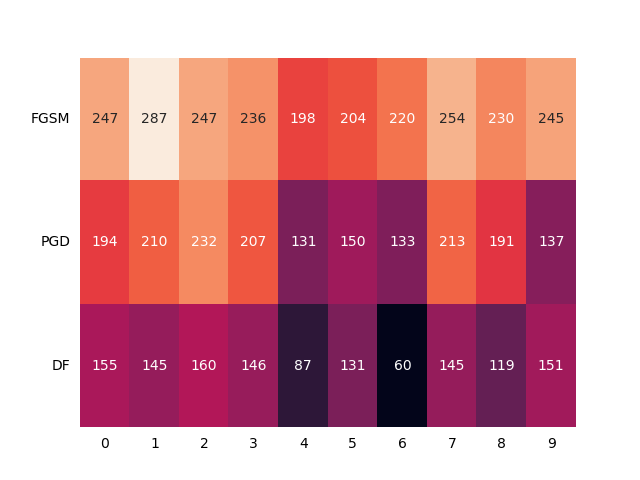}
     \end{subfigure}
        \caption{Experiment 5 (trained jointly - faster initialization). Class/attack accuracy breakdown for the $3$ attack/generator setting. Each row corresponds to one generator. The heatmaps present the accuracy for a given class/attack (left) and the number of samples on which a given generator won (right).}
\end{figure}
\clearpage

\begin{figure}[ht]
     \centering
     \begin{subfigure}[b]{0.49\textwidth}
         \centering
         \includegraphics[width=\textwidth]{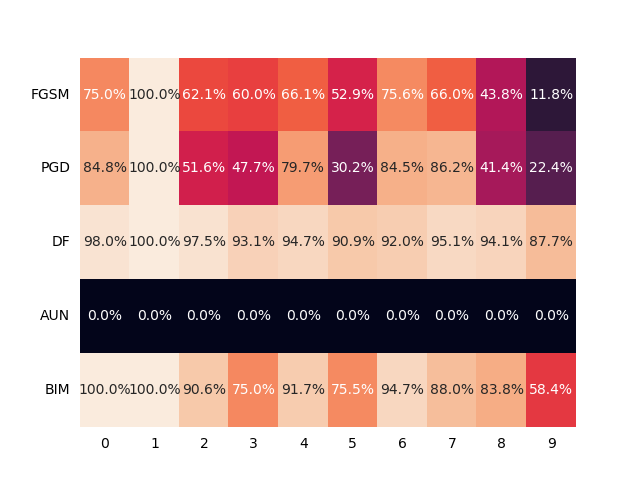}
     \end{subfigure}
     \hfill
     \begin{subfigure}[b]{0.49\textwidth}
         \centering
         \includegraphics[width=\textwidth]{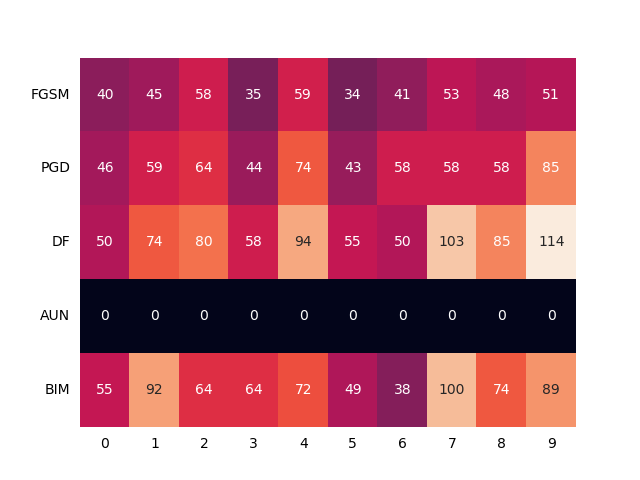}
     \end{subfigure}
     \hfill
     \begin{subfigure}[b]{0.49\textwidth}
         \centering
         \includegraphics[width=\textwidth]{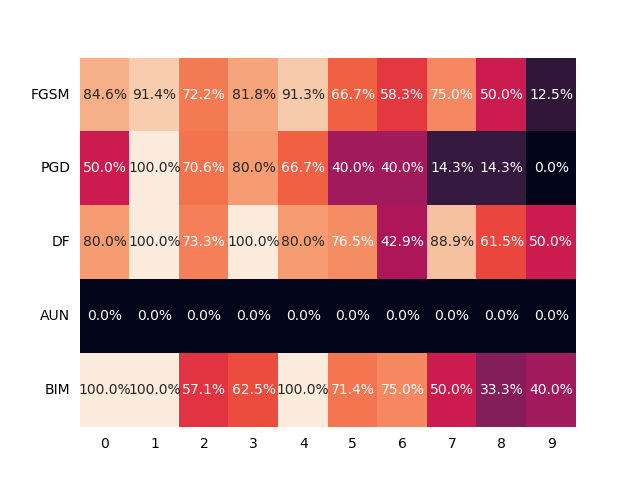}
     \end{subfigure}
     \hfill
     \begin{subfigure}[b]{0.49\textwidth}
         \centering
         \includegraphics[width=\textwidth]{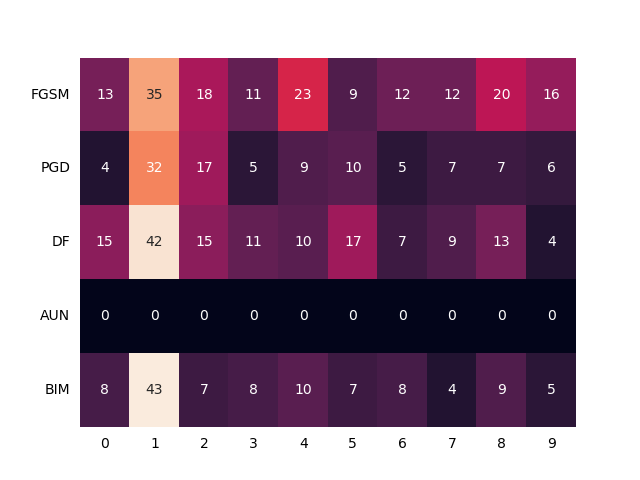}
     \end{subfigure}
     \hfill
     \begin{subfigure}[b]{0.49\textwidth}
         \centering
         \includegraphics[width=\textwidth]{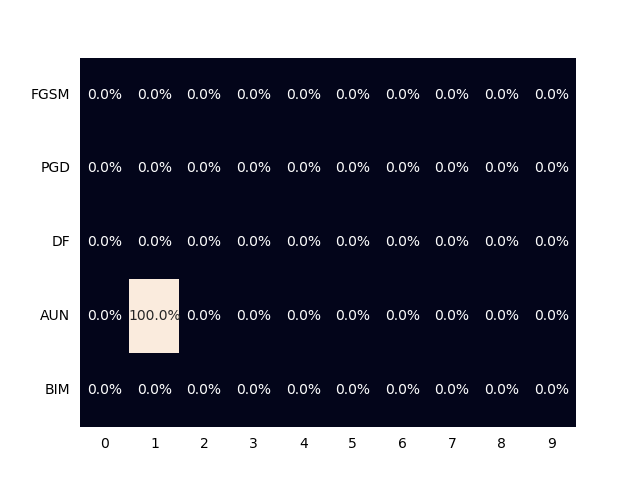}
     \end{subfigure}
     \hfill
     \begin{subfigure}[b]{0.49\textwidth}
         \centering
         \includegraphics[width=\textwidth]{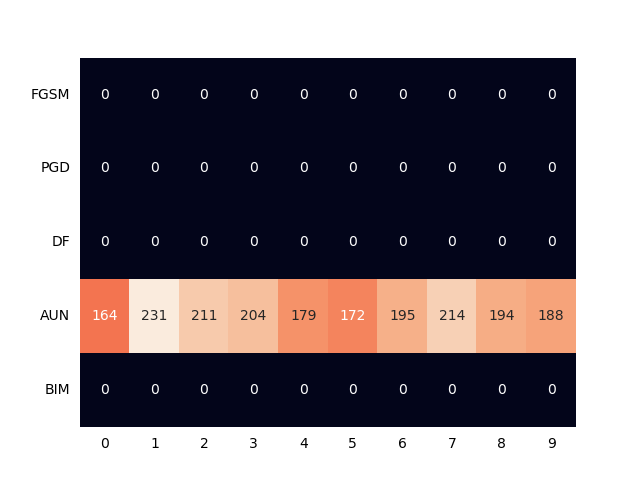}
     \end{subfigure}
        \caption{Experiment 5 (trained jointly - faster initialization). Class/attack accuracy breakdown for the $5$ attack/generator setting, part 1 (generators $1$ to $3$). Each row corresponds to one generator. The heatmaps present the accuracy for a given class/attack (left) and the number of samples on which a given generator won (right).}
\end{figure}
\clearpage

\begin{figure}[ht]
     \centering
     \begin{subfigure}[b]{0.49\textwidth}
         \centering
         \includegraphics[width=\textwidth]{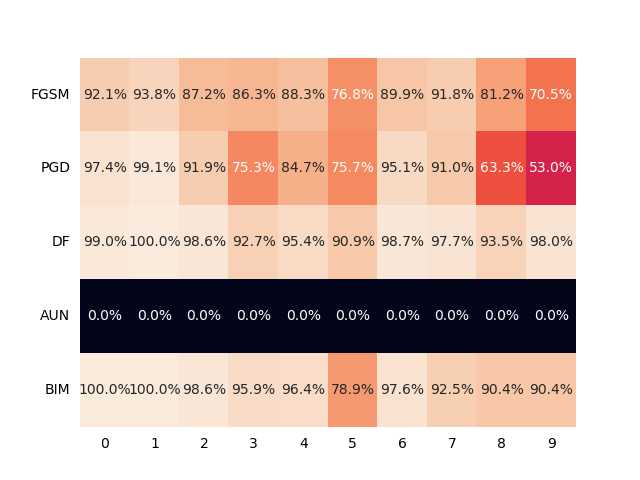}
     \end{subfigure}
     \hfill
     \begin{subfigure}[b]{0.49\textwidth}
         \centering
         \includegraphics[width=\textwidth]{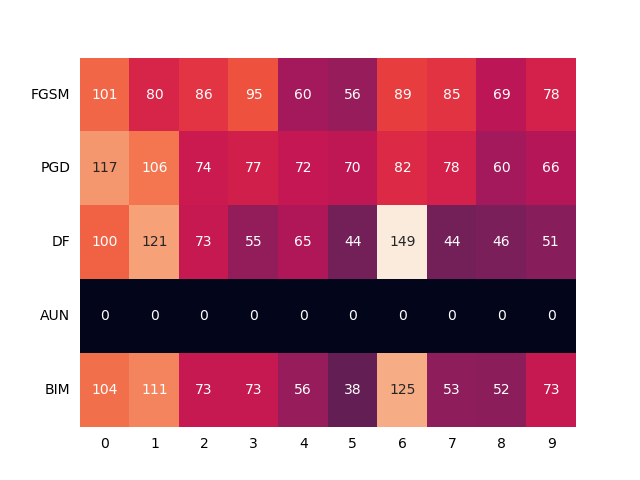}
     \end{subfigure}
     \hfill
     \begin{subfigure}[b]{0.49\textwidth}
         \centering
         \includegraphics[width=\textwidth]{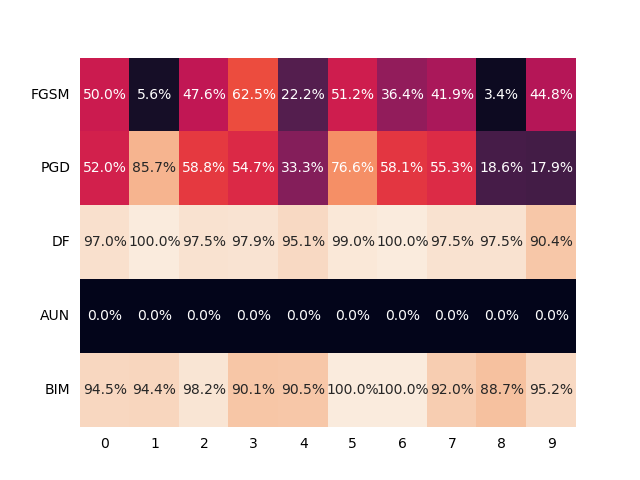}
     \end{subfigure}
     \hfill
     \begin{subfigure}[b]{0.49\textwidth}
         \centering
         \includegraphics[width=\textwidth]{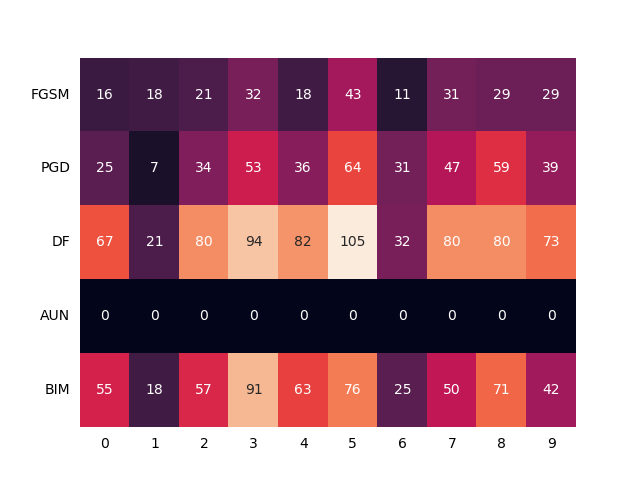}
     \end{subfigure}
        \caption{Experiment 5 (trained jointly - faster initialization). Class/attack accuracy breakdown for the $5$ attack/generator setting, part 2 (generators $4$ and $5$). Each row corresponds to one generator. The heatmaps present the accuracy for a given class/attack (left) and the number of samples on which a given generator won (right).}
\end{figure}
\clearpage

\begin{figure}[ht]
     \centering
     \begin{subfigure}[b]{0.49\textwidth}
         \centering
         \includegraphics[width=\textwidth]{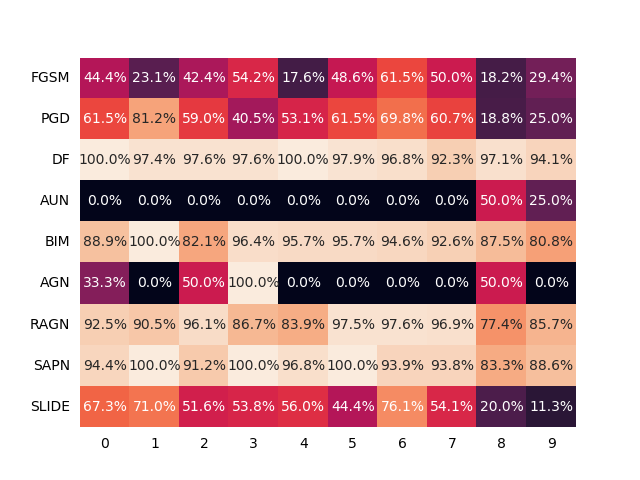}
     \end{subfigure}
     \hfill
     \begin{subfigure}[b]{0.49\textwidth}
         \centering
         \includegraphics[width=\textwidth]{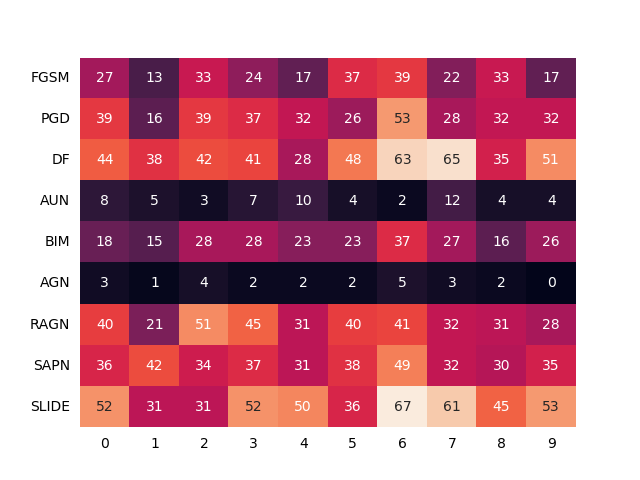}
     \end{subfigure}
     \hfill
     \begin{subfigure}[b]{0.49\textwidth}
         \centering
         \includegraphics[width=\textwidth]{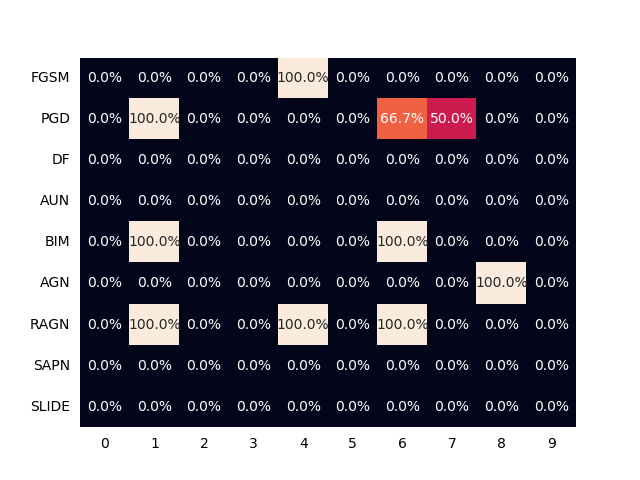}
     \end{subfigure}
     \hfill
     \begin{subfigure}[b]{0.49\textwidth}
         \centering
         \includegraphics[width=\textwidth]{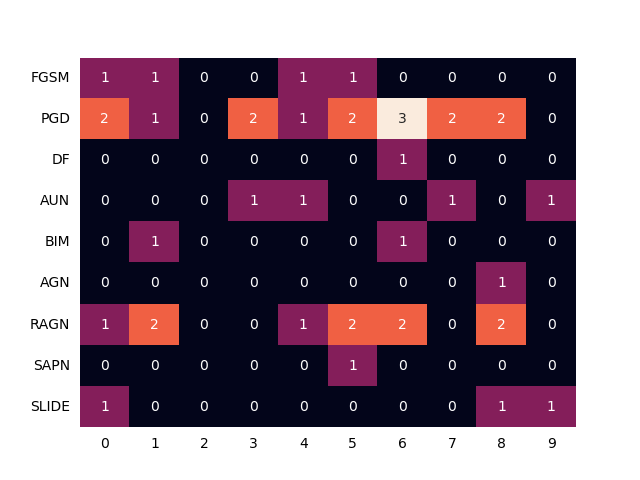}
     \end{subfigure}
     \hfill
     \begin{subfigure}[b]{0.49\textwidth}
         \centering
         \includegraphics[width=\textwidth]{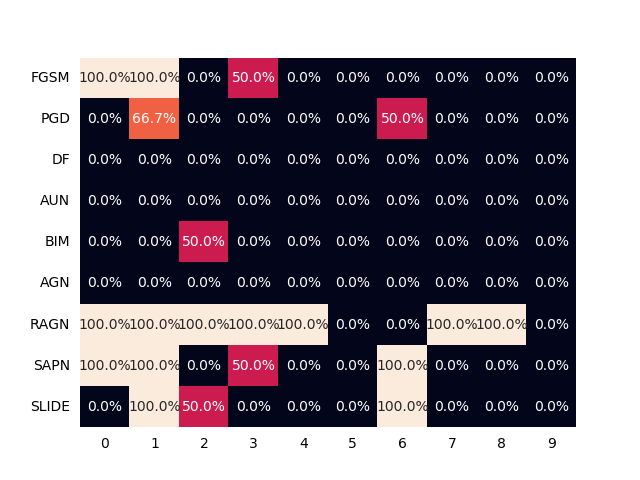}
     \end{subfigure}
     \hfill
     \begin{subfigure}[b]{0.49\textwidth}
         \centering
         \includegraphics[width=\textwidth]{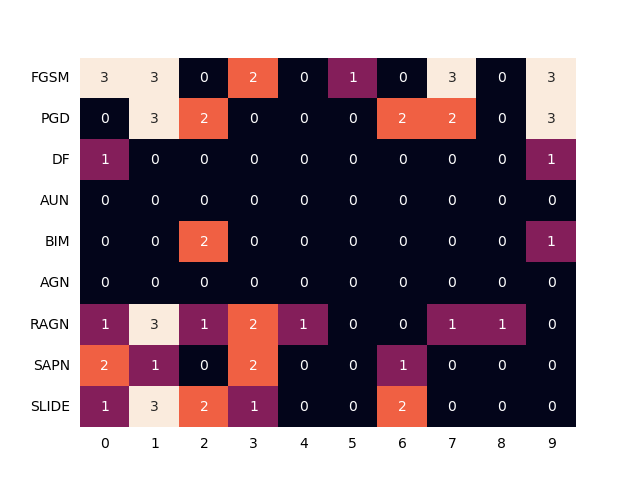}
     \end{subfigure}
        \caption{Experiment 5 (trained jointly - faster initialization). Class/attack accuracy breakdown for the $9$ attack/generator setting, part 1 (generators $1$ to $3$). Each row corresponds to one generator. The heatmaps present the accuracy for a given class/attack (left) and the number of samples on which a given generator won (right).}
\end{figure}
\clearpage

\begin{figure}[ht]
     \centering
     \begin{subfigure}[b]{0.49\textwidth}
         \centering
         \includegraphics[width=\textwidth]{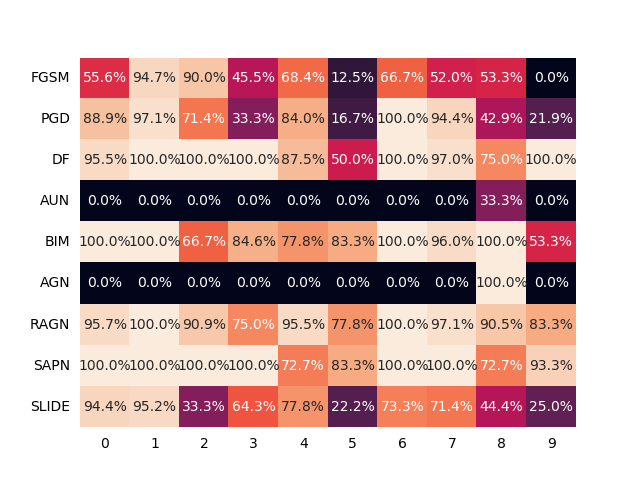}
     \end{subfigure}
     \hfill
     \begin{subfigure}[b]{0.49\textwidth}
         \centering
         \includegraphics[width=\textwidth]{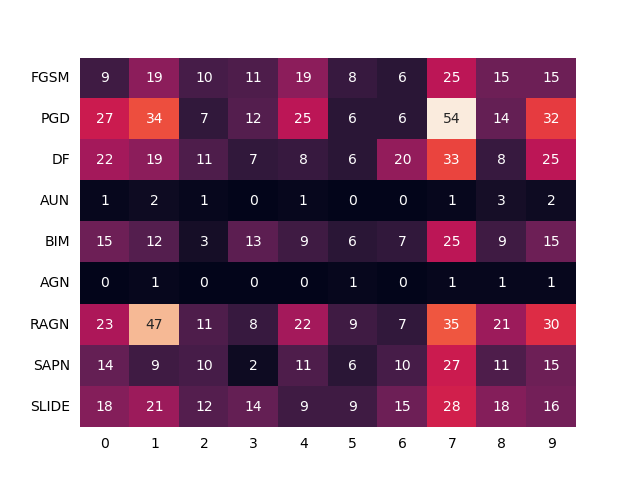}
     \end{subfigure}
     \hfill
     \begin{subfigure}[b]{0.49\textwidth}
         \centering
         \includegraphics[width=\textwidth]{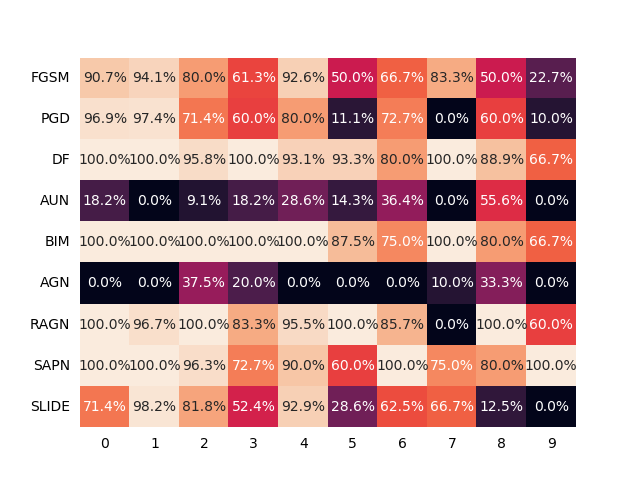}
     \end{subfigure}
     \hfill
     \begin{subfigure}[b]{0.49\textwidth}
         \centering
         \includegraphics[width=\textwidth]{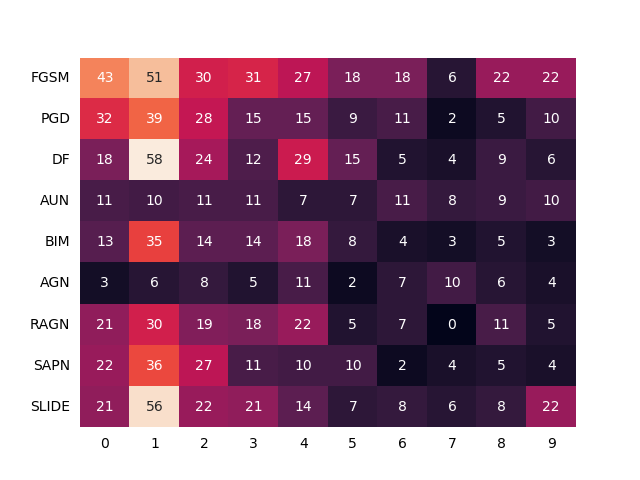}
     \end{subfigure}
     \hfill
     \begin{subfigure}[b]{0.49\textwidth}
         \centering
         \includegraphics[width=\textwidth]{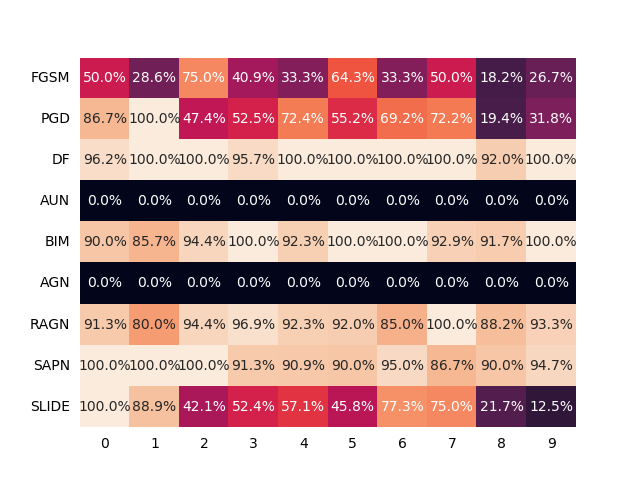}
     \end{subfigure}
     \hfill
     \begin{subfigure}[b]{0.49\textwidth}
         \centering
         \includegraphics[width=\textwidth]{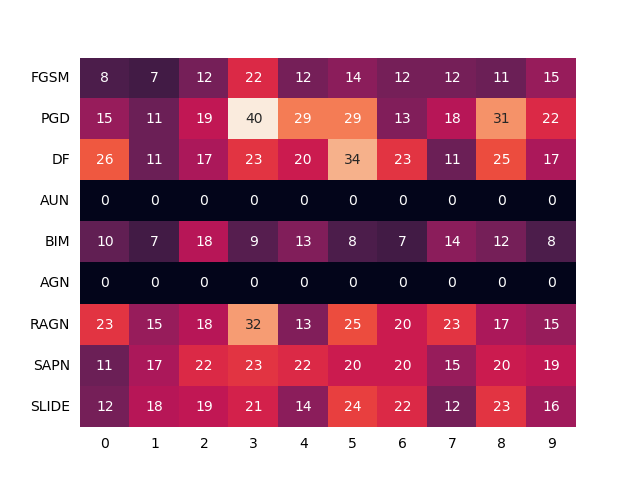}
     \end{subfigure}
        \caption{Experiment 5 (trained jointly - faster initialization). Class/attack accuracy breakdown for the $9$ attack/generator setting, part 2 (generators $4$ to $6$). Each row corresponds to one generator. The heatmaps present the accuracy for a given class/attack (left) and the number of samples on which a given generator won (right).}
\end{figure}
\clearpage

\begin{figure}[ht]
     \centering
     \begin{subfigure}[b]{0.49\textwidth}
         \centering
         \includegraphics[width=\textwidth]{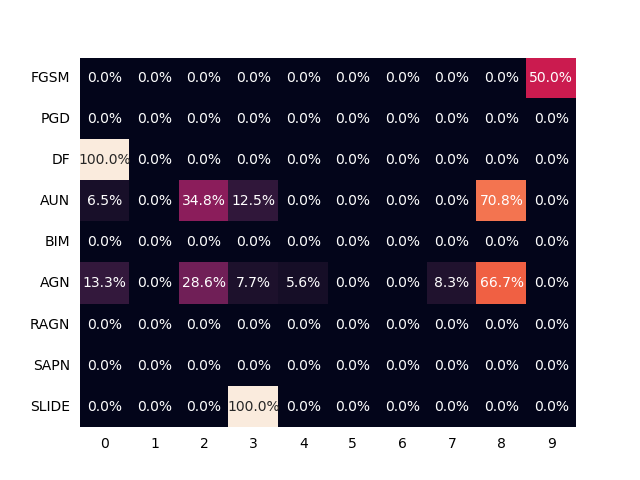}
     \end{subfigure}
     \hfill
     \begin{subfigure}[b]{0.49\textwidth}
         \centering
         \includegraphics[width=\textwidth]{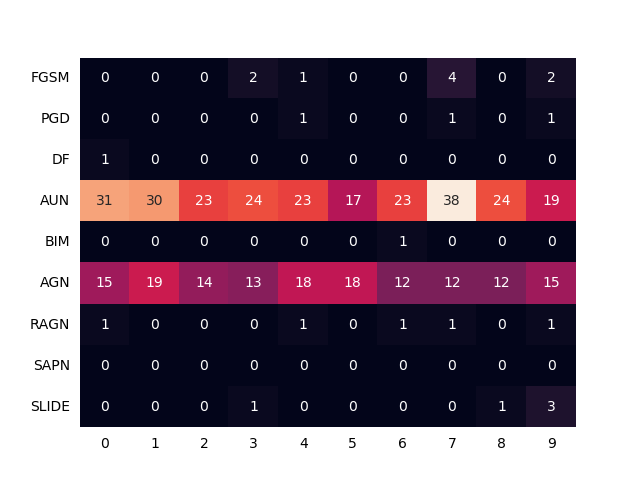}
     \end{subfigure}
     \hfill
     \begin{subfigure}[b]{0.49\textwidth}
         \centering
         \includegraphics[width=\textwidth]{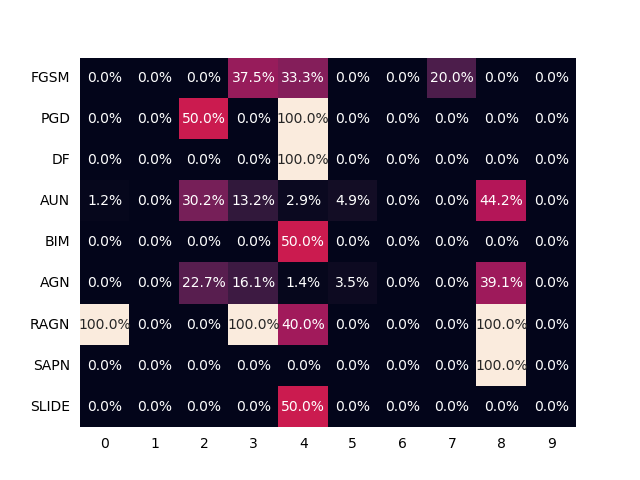}
     \end{subfigure}
     \hfill
     \begin{subfigure}[b]{0.49\textwidth}
         \centering
         \includegraphics[width=\textwidth]{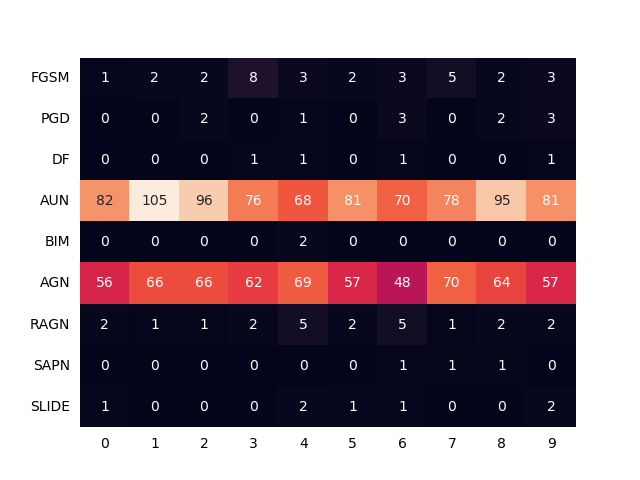}
     \end{subfigure}
     \hfill
     \begin{subfigure}[b]{0.49\textwidth}
         \centering
         \includegraphics[width=\textwidth]{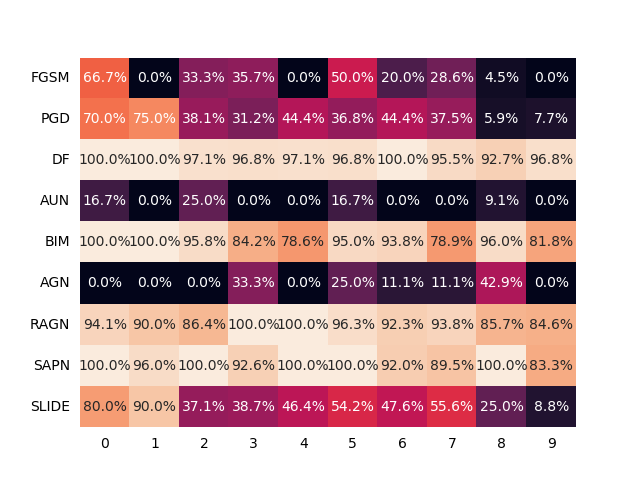}
     \end{subfigure}
     \hfill
     \begin{subfigure}[b]{0.49\textwidth}
         \centering
         \includegraphics[width=\textwidth]{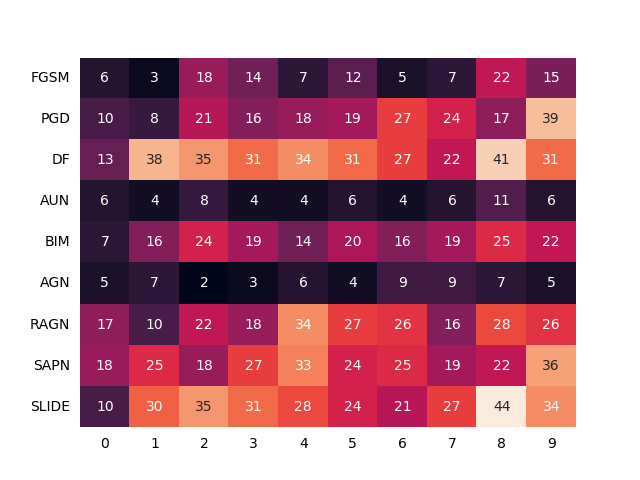}
     \end{subfigure}
        \caption{Experiment 5 (trained jointly - faster initialization). Class/attack accuracy breakdown for the $9$ attack/generator setting, part 3 (generators $7$ to $9$). Each row corresponds to one generator. The heatmaps present the accuracy for a given class/attack (left) and the number of samples on which a given generator won (right).}
\end{figure}
\clearpage

\subsubsection{Experiment 7, trained jointly - faster initialization}

\begin{figure}[ht]
     \centering
     \begin{subfigure}[b]{0.49\textwidth}
         \centering
         \includegraphics[width=\textwidth]{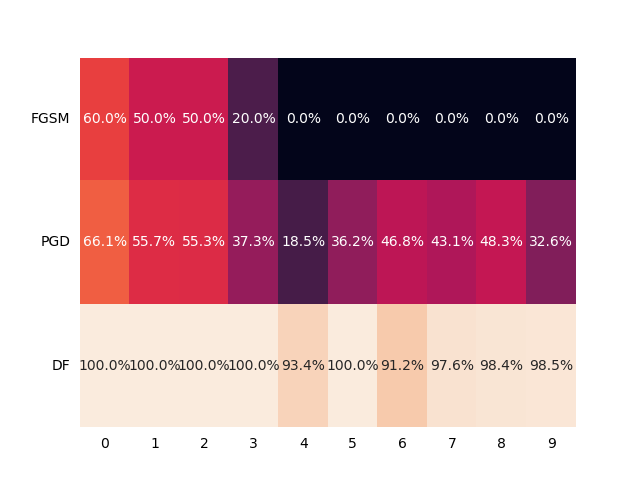}
     \end{subfigure}
     \hfill
     \begin{subfigure}[b]{0.49\textwidth}
         \centering
         \includegraphics[width=\textwidth]{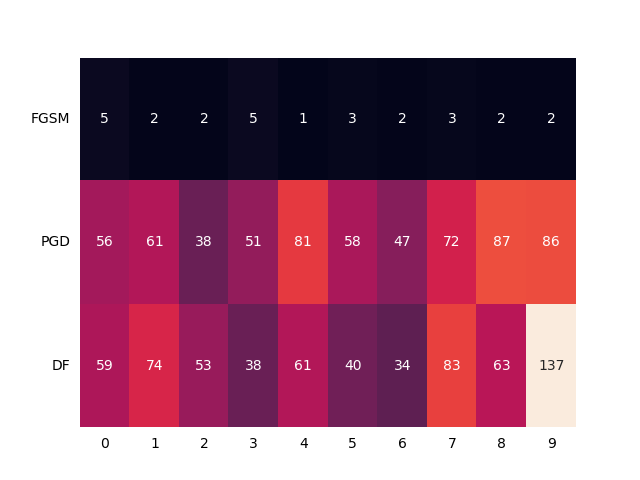}
     \end{subfigure}
     \hfill
     \begin{subfigure}[b]{0.49\textwidth}
         \centering
         \includegraphics[width=\textwidth]{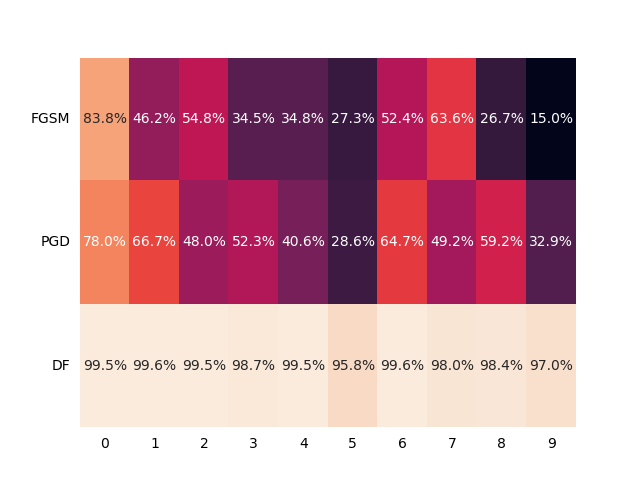}
     \end{subfigure}
     \hfill
     \begin{subfigure}[b]{0.49\textwidth}
         \centering
         \includegraphics[width=\textwidth]{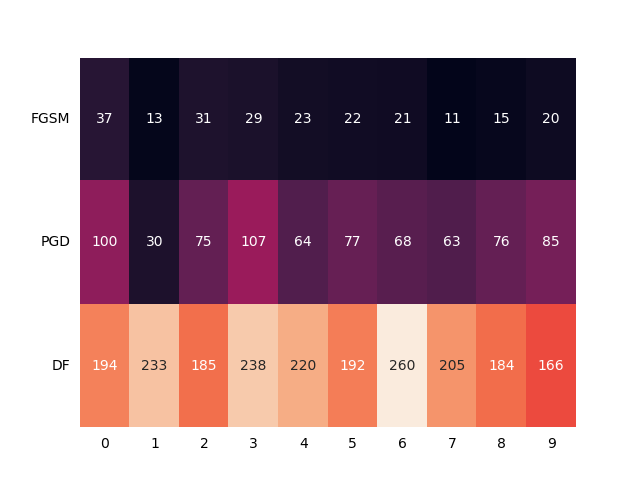}
     \end{subfigure}
     \hfill
     \begin{subfigure}[b]{0.49\textwidth}
         \centering
         \includegraphics[width=\textwidth]{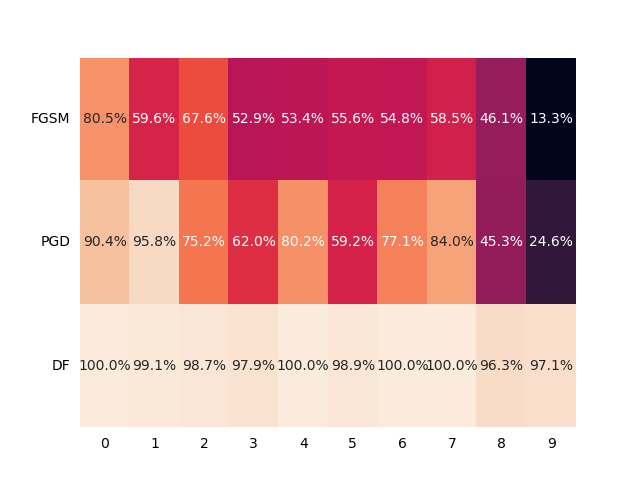}
     \end{subfigure}
     \hfill
     \begin{subfigure}[b]{0.49\textwidth}
         \centering
         \includegraphics[width=\textwidth]{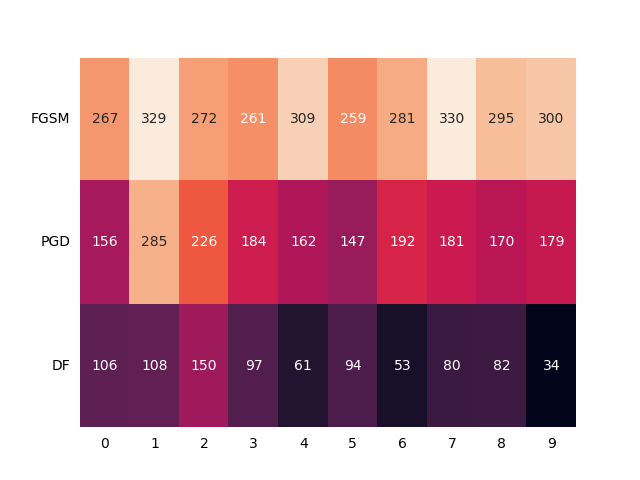}
     \end{subfigure}
        \caption{Experiment 7 (trained jointly - faster initialization). Class/attack accuracy breakdown for the $3$ attack/generator setting. Each row corresponds to one generator. The heatmaps present the accuracy for a given class/attack (left) and the number of samples on which a given generator won (right).}
\end{figure}
\clearpage

\begin{figure}[ht]
     \centering
     \begin{subfigure}[b]{0.49\textwidth}
         \centering
         \includegraphics[width=\textwidth]{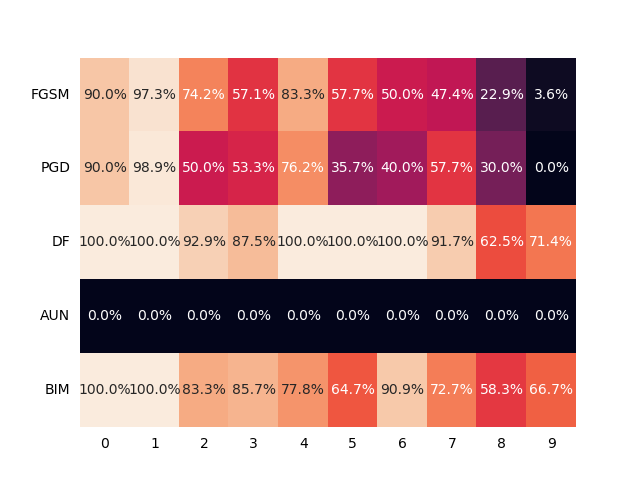}
     \end{subfigure}
     \hfill
     \begin{subfigure}[b]{0.49\textwidth}
         \centering
         \includegraphics[width=\textwidth]{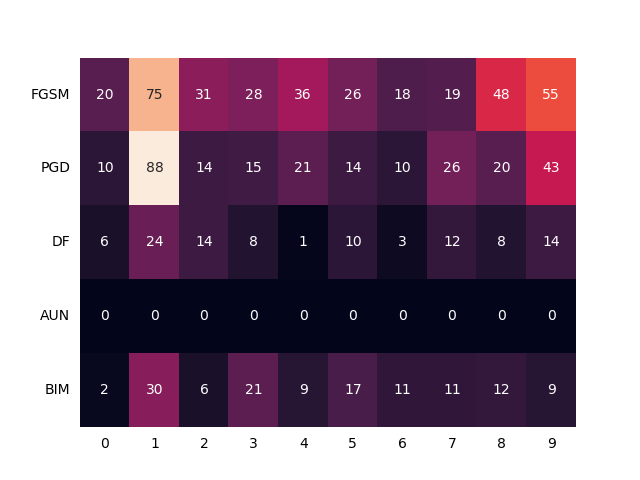}
     \end{subfigure}
     \hfill
     \begin{subfigure}[b]{0.49\textwidth}
         \centering
         \includegraphics[width=\textwidth]{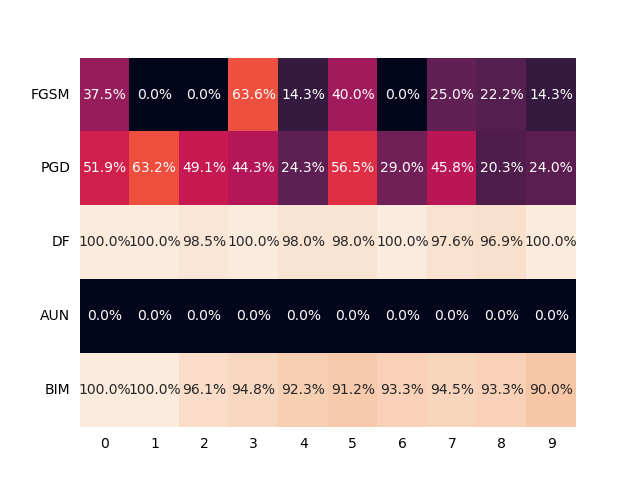}
     \end{subfigure}
     \hfill
     \begin{subfigure}[b]{0.49\textwidth}
         \centering
         \includegraphics[width=\textwidth]{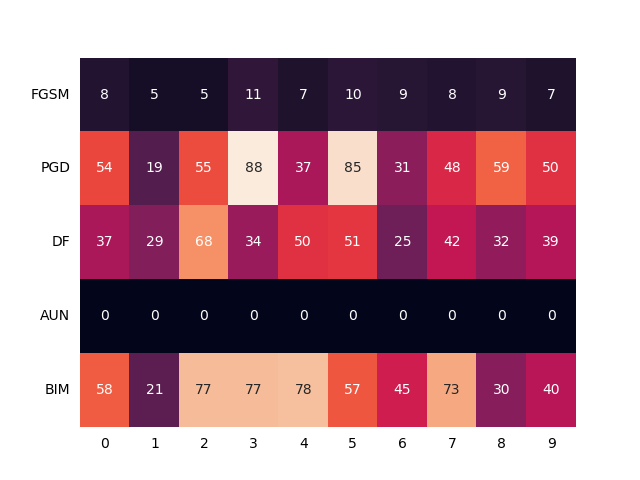}
     \end{subfigure}
     \hfill
     \begin{subfigure}[b]{0.49\textwidth}
         \centering
         \includegraphics[width=\textwidth]{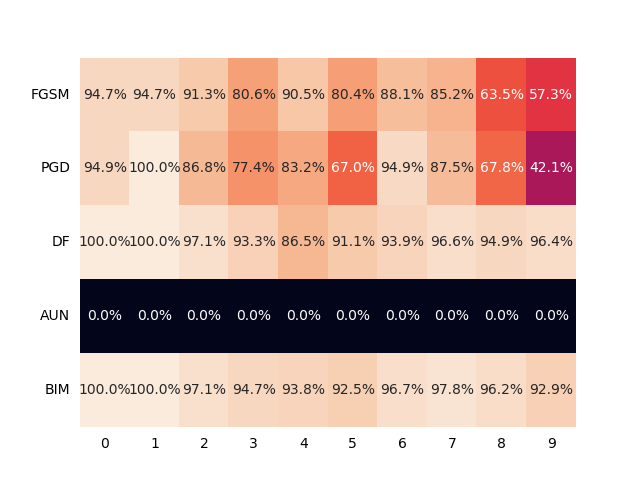}
     \end{subfigure}
     \hfill
     \begin{subfigure}[b]{0.49\textwidth}
         \centering
         \includegraphics[width=\textwidth]{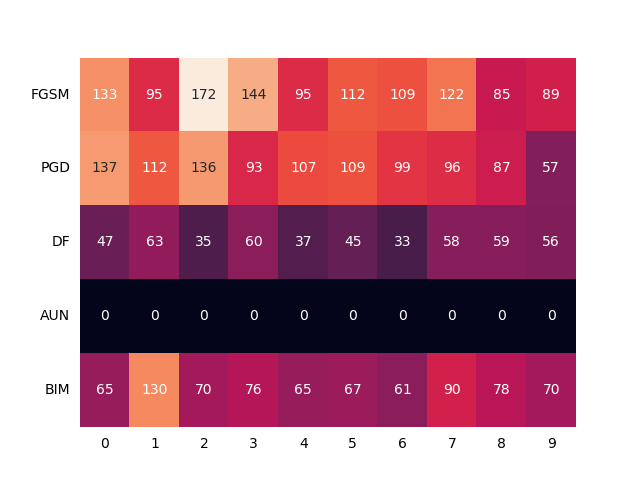}
     \end{subfigure}
        \caption{Experiment 7 (trained jointly - faster initialization). Class/attack accuracy breakdown for the $5$ attack/generator setting, part 1 (generators $1$ to $3$). Each row corresponds to one generator. The heatmaps present the accuracy for a given class/attack (left) and the number of samples on which a given generator won (right).}
\end{figure}
\clearpage

\begin{figure}[ht]
     \centering
     \begin{subfigure}[b]{0.49\textwidth}
         \centering
         \includegraphics[width=\textwidth]{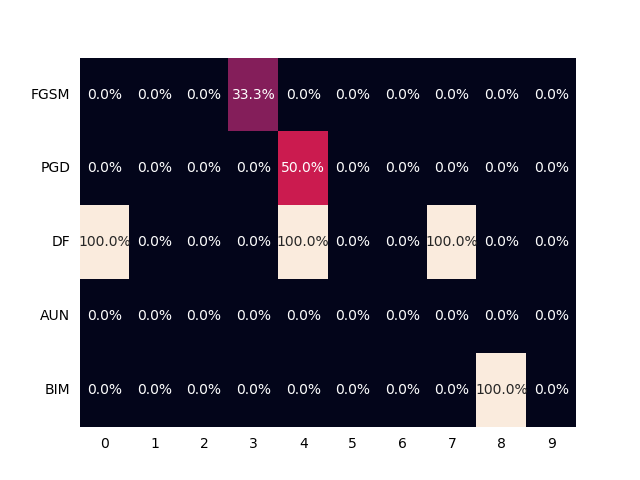}
     \end{subfigure}
     \hfill
     \begin{subfigure}[b]{0.49\textwidth}
         \centering
         \includegraphics[width=\textwidth]{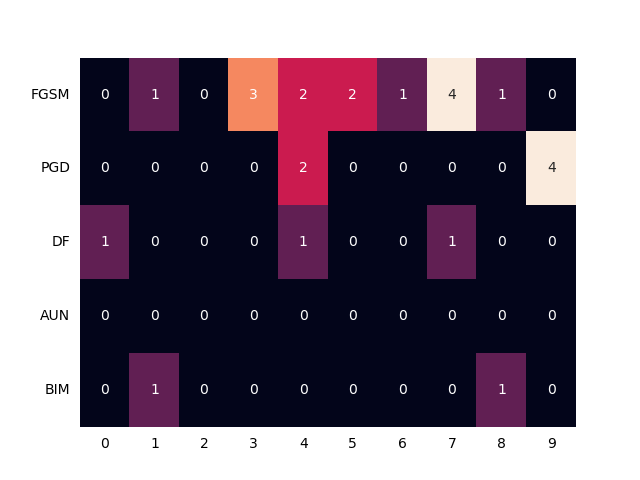}
     \end{subfigure}
     \hfill
     \begin{subfigure}[b]{0.49\textwidth}
         \centering
         \includegraphics[width=\textwidth]{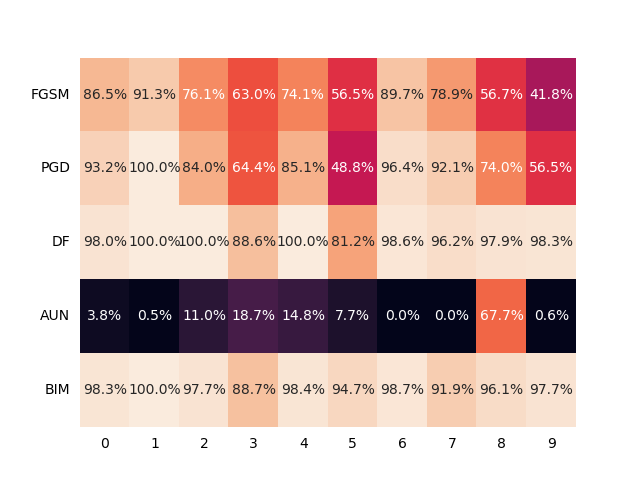}
     \end{subfigure}
     \hfill
     \begin{subfigure}[b]{0.49\textwidth}
         \centering
         \includegraphics[width=\textwidth]{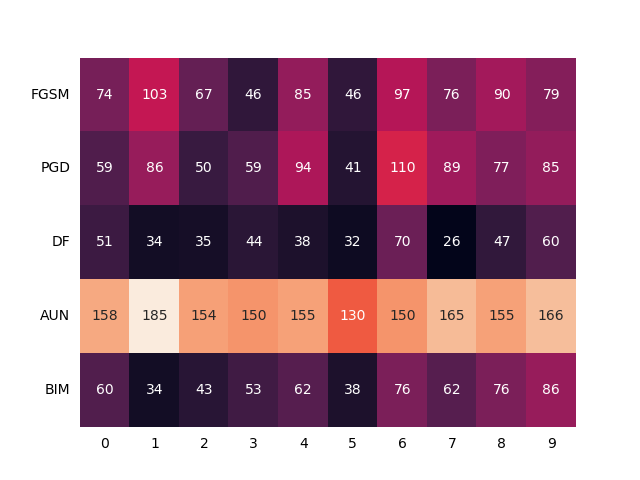}
     \end{subfigure}
        \caption{Experiment 7 (trained jointly - faster initialization). Class/attack accuracy breakdown for the $5$ attack/generator setting, part 2 (generators $4$ and $5$). Each row corresponds to one generator. The heatmaps present the accuracy for a given class/attack (left) and the number of samples on which a given generator won (right).}
\end{figure}
\clearpage

\begin{figure}[ht]
     \centering
     \begin{subfigure}[b]{0.49\textwidth}
         \centering
         \includegraphics[width=\textwidth]{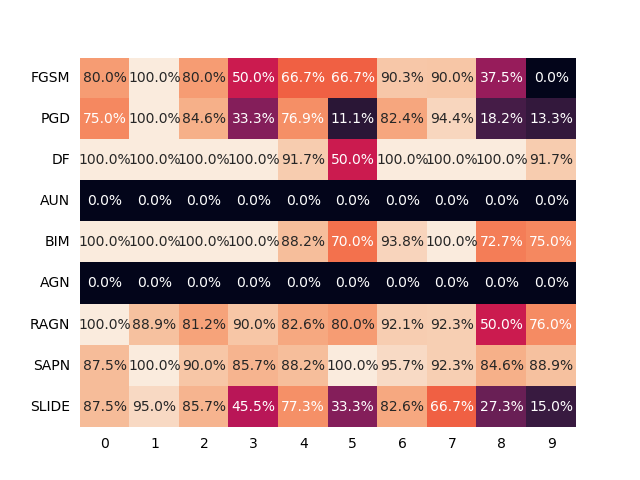}
     \end{subfigure}
     \hfill
     \begin{subfigure}[b]{0.49\textwidth}
         \centering
         \includegraphics[width=\textwidth]{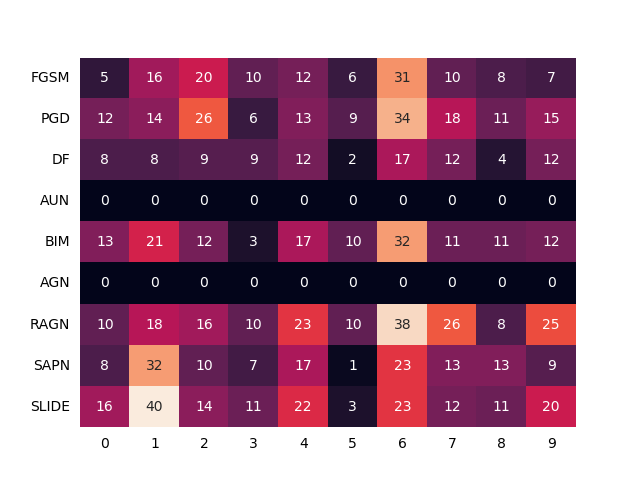}
     \end{subfigure}
     \hfill
     \begin{subfigure}[b]{0.49\textwidth}
         \centering
         \includegraphics[width=\textwidth]{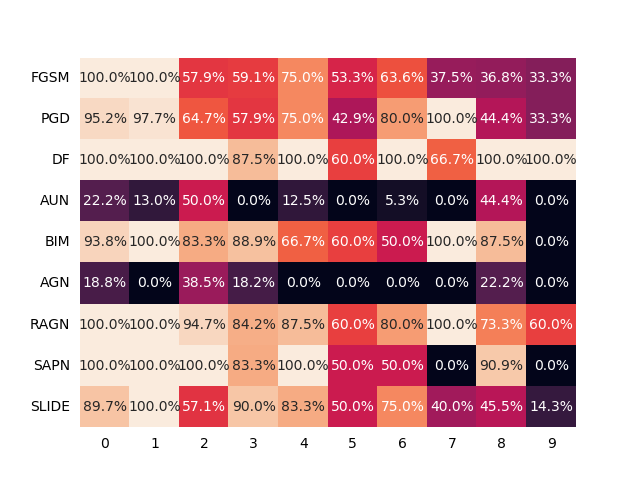}
     \end{subfigure}
     \hfill
     \begin{subfigure}[b]{0.49\textwidth}
         \centering
         \includegraphics[width=\textwidth]{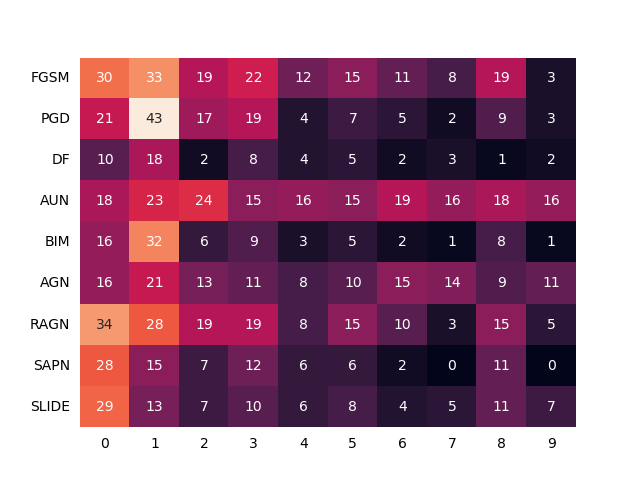}
     \end{subfigure}
     \hfill
     \begin{subfigure}[b]{0.49\textwidth}
         \centering
         \includegraphics[width=\textwidth]{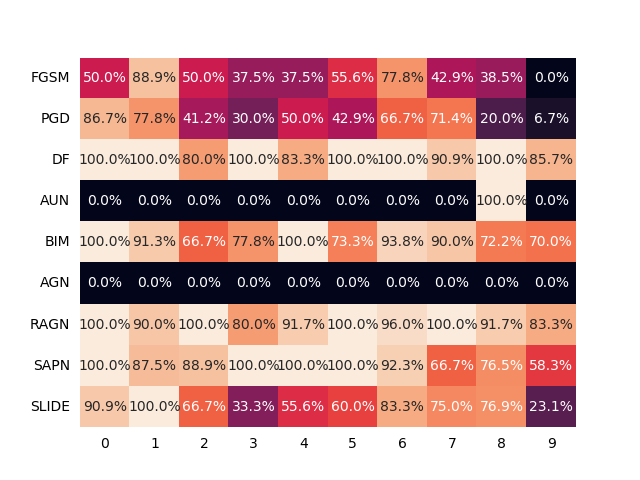}
     \end{subfigure}
     \hfill
     \begin{subfigure}[b]{0.49\textwidth}
         \centering
         \includegraphics[width=\textwidth]{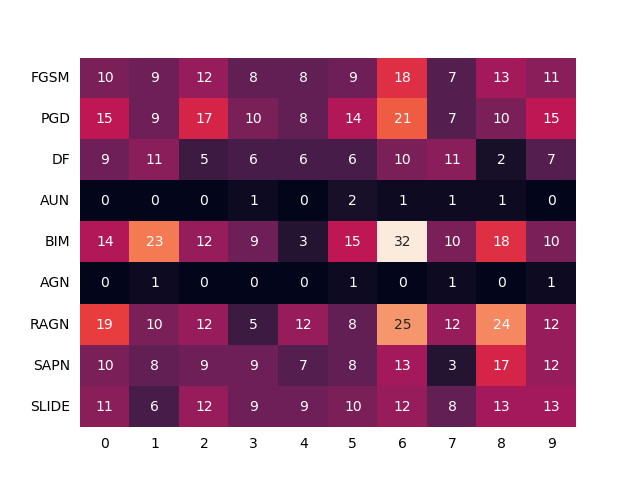}
     \end{subfigure}
        \caption{Experiment 7 (trained jointly - faster initialization). Class/attack accuracy breakdown for the $9$ attack/generator setting, part 1 (generators $1$ to $3$). Each row corresponds to one generator. The heatmaps present the accuracy for a given class/attack (left) and the number of samples on which a given generator won (right).}
\end{figure}
\clearpage

\begin{figure}[ht]
     \centering
     \begin{subfigure}[b]{0.49\textwidth}
         \centering
         \includegraphics[width=\textwidth]{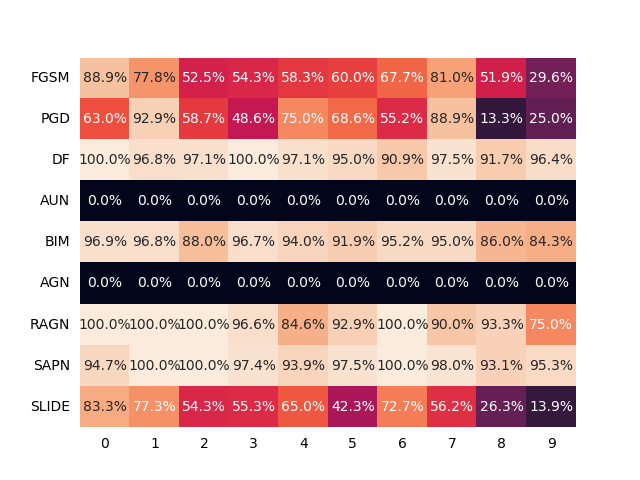}
     \end{subfigure}
     \hfill
     \begin{subfigure}[b]{0.49\textwidth}
         \centering
         \includegraphics[width=\textwidth]{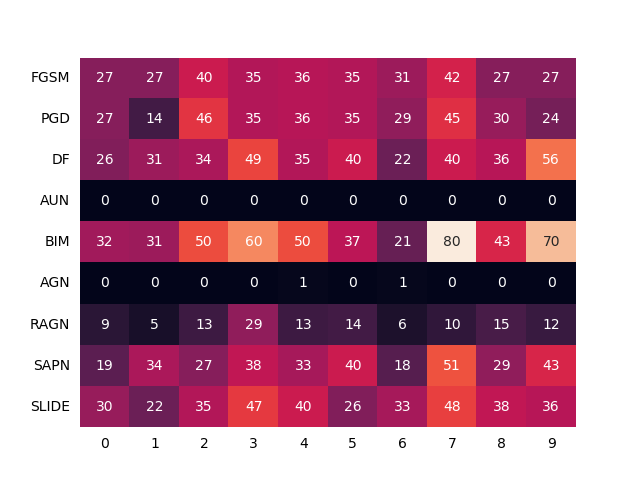}
     \end{subfigure}
     \hfill
     \begin{subfigure}[b]{0.49\textwidth}
         \centering
         \includegraphics[width=\textwidth]{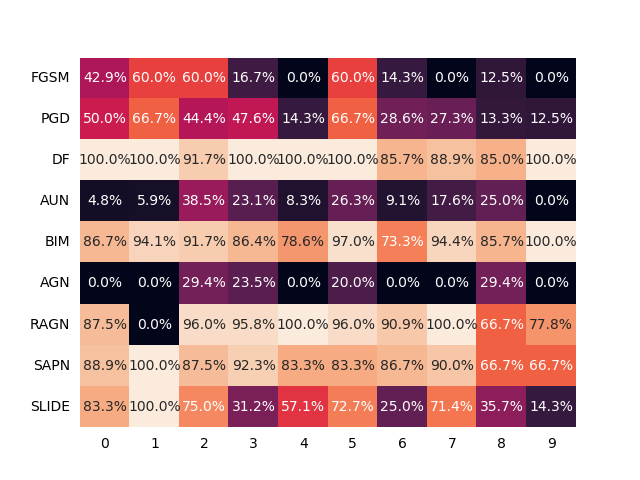}
     \end{subfigure}
     \hfill
     \begin{subfigure}[b]{0.49\textwidth}
         \centering
         \includegraphics[width=\textwidth]{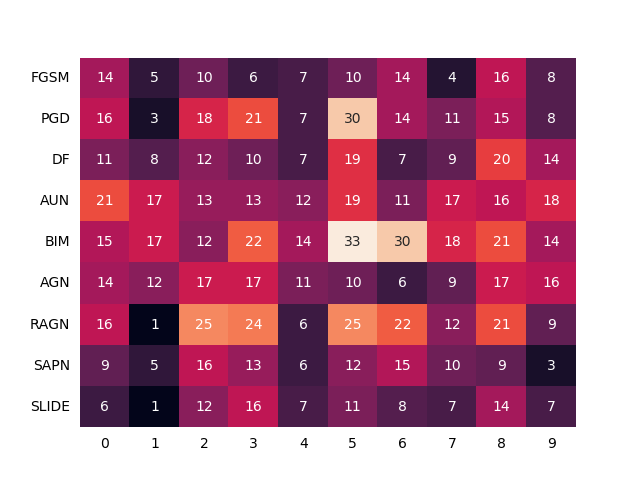}
     \end{subfigure}
     \hfill
     \begin{subfigure}[b]{0.49\textwidth}
         \centering
         \includegraphics[width=\textwidth]{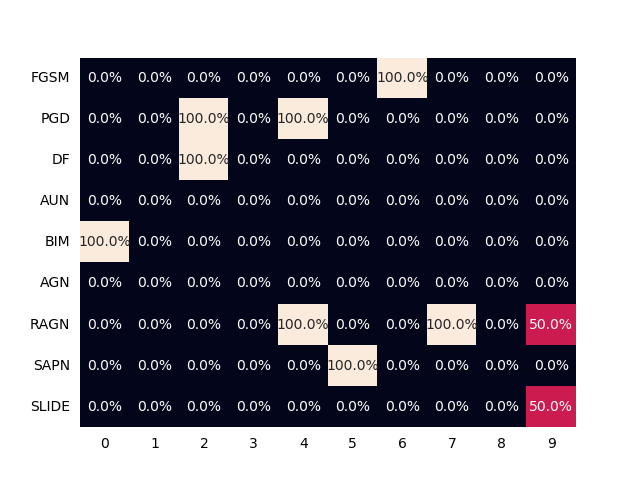}
     \end{subfigure}
     \hfill
     \begin{subfigure}[b]{0.49\textwidth}
         \centering
         \includegraphics[width=\textwidth]{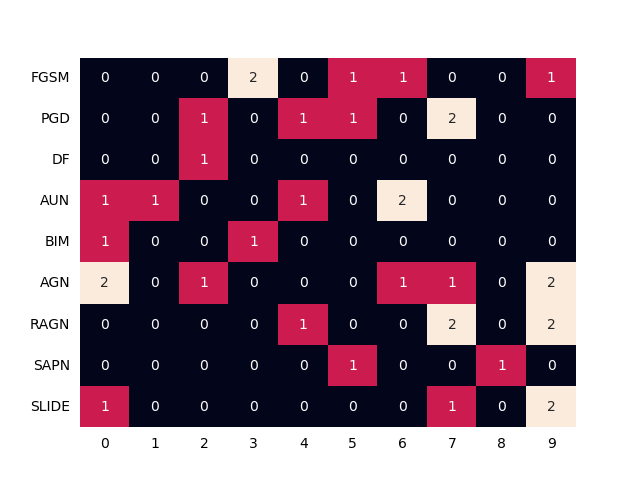}
     \end{subfigure}
        \caption{Experiment 7 (trained jointly - faster initialization). Class/attack accuracy breakdown for the $9$ attack/generator setting, part 2 (generators $4$ to $6$). Each row corresponds to one generator. The heatmaps present the accuracy for a given class/attack (left) and the number of samples on which a given generator won (right).}
\end{figure}
\clearpage

\begin{figure}[ht]
     \centering
     \begin{subfigure}[b]{0.49\textwidth}
         \centering
         \includegraphics[width=\textwidth]{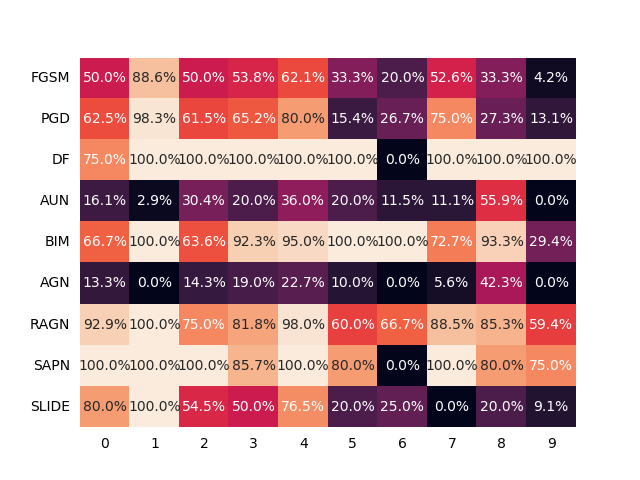}
     \end{subfigure}
     \hfill
     \begin{subfigure}[b]{0.49\textwidth}
         \centering
         \includegraphics[width=\textwidth]{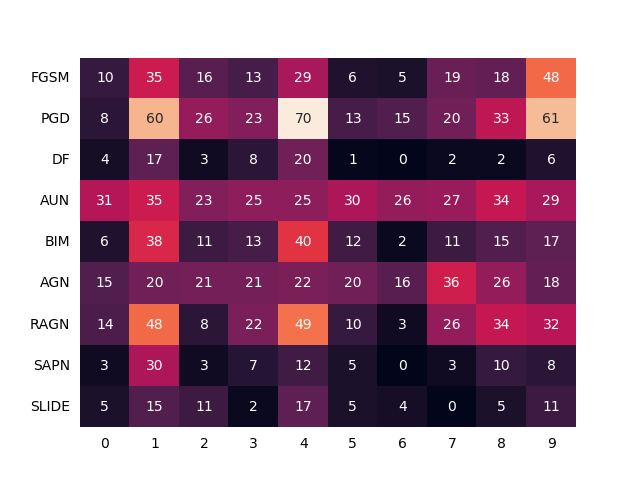}
     \end{subfigure}
     \hfill
     \begin{subfigure}[b]{0.49\textwidth}
         \centering
         \includegraphics[width=\textwidth]{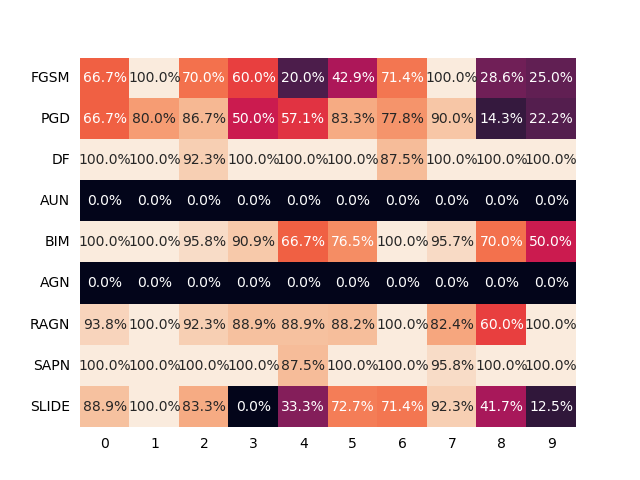}
     \end{subfigure}
     \hfill
     \begin{subfigure}[b]{0.49\textwidth}
         \centering
         \includegraphics[width=\textwidth]{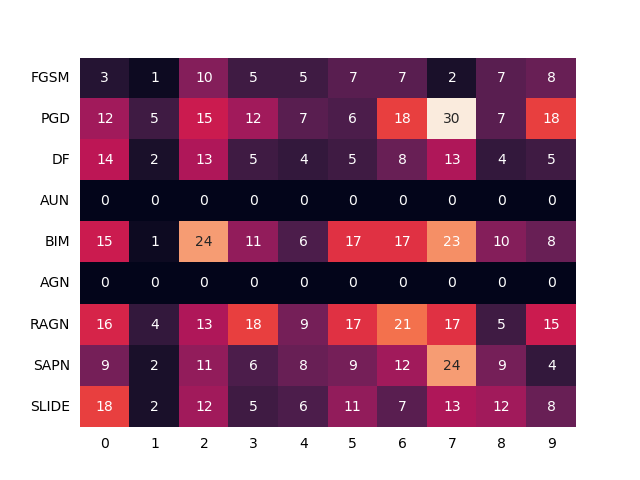}
     \end{subfigure}
     \hfill
     \begin{subfigure}[b]{0.49\textwidth}
         \centering
         \includegraphics[width=\textwidth]{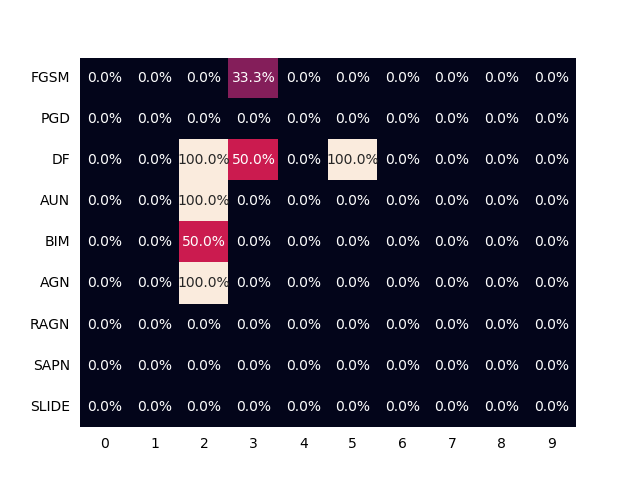}
     \end{subfigure}
     \hfill
     \begin{subfigure}[b]{0.49\textwidth}
         \centering
         \includegraphics[width=\textwidth]{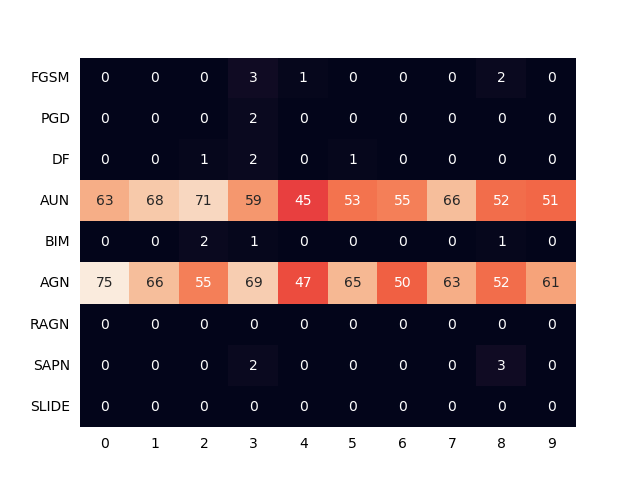}
     \end{subfigure}
        \caption{Experiment 7 (trained jointly - faster initialization). Class/attack accuracy breakdown for the $9$ attack/generator setting, part 3 (generators $7$ to $9$). Each row corresponds to one generator. The heatmaps present the accuracy for a given class/attack (left) and the number of samples on which a given generator won (right).}
        \label{fig:repeated-last}
\end{figure}
\clearpage

\medskip

\end{document}